\documentclass{article}

\PassOptionsToPackage{numbers, compress}{natbib}
\usepackage[final]{neurips_2021}

\usepackage[utf8]{inputenc} % allow utf-8 input
\usepackage[T1]{fontenc}    % use 8-bit T1 fonts
\usepackage{hyperref}       % hyperlinks
\usepackage{url}            % simple URL typesetting
\usepackage{booktabs}       % professional-quality tables
\usepackage{amsfonts}       % blackboard math symbols
\usepackage{nicefrac}       % compact symbols for 1/2, etc.
\usepackage{microtype}      % microtypography
\usepackage{xcolor}         % colors

\usepackage{algorithm}
\usepackage[noend]{algpseudocode}
\usepackage{amsmath}
\usepackage{bbm}
\usepackage{graphicx}
\usepackage{mathtools}
\usepackage{subcaption}

\newcommand{\argmin}{\operatorname*{arg\,min}}
\newcommand{\fracpartial}[2]{\frac{\partial #1}{\partial  #2}}

\newcommand{\psigsmall}{{\small{$\sigma^2_{\hat{y}_{te}}$}}}

\newcommand{\pmusmall}{{\small{$\mu_{\hat{y}_{te}}$}}}
\newcommand{\affinsmall}{{\small{$A^{(k)}$}}}

\newcommand{\eat}[1]{}

\title{Instance-Based Uncertainty Estimation for Gradient-Boosted Regression Trees}

\author{%
  Jonathan Brophy\thanks{\url{https://www.jonathanbrophy.com}} \\
  University of Oregon\\
  \texttt{jbrophy@cs.uoregon.edu} \\
  \And
  Daniel Lowd\\
  University of Oregon\\
  \texttt{lowd@cs.uoregon.edu} \\
}

\begin{document}

\maketitle

\begin{abstract}
Gradient-boosted regression trees~(GBRTs) are hugely popular for solving tabular regression problems, but provide no estimate of uncertainty.
We propose \underline{I}nstance-\underline{B}ased \underline{U}ncertainty estimation for \underline{G}radient-boosted regression trees~(IBUG), a simple method for extending any GBRT point predictor to produce probabilistic predictions. 
IBUG computes a non-parametric distribution around a prediction using the $k$-nearest training instances, where distance is measured with a tree-ensemble kernel.
The runtime of IBUG depends on the number of training examples at each leaf in the ensemble, and can be improved by sampling trees or training instances.
Empirically, we find that IBUG achieves similar or better performance than the previous state-of-the-art across 22 benchmark regression datasets.
We also find that IBUG can achieve improved probabilistic performance by using different base GBRT models, and can more flexibly model the posterior distribution of a prediction than competing methods. We also find that previous methods suffer from poor probabilistic calibration on some datasets, which can be mitigated using a scalar factor tuned on the validation data. Source code is available at~\url{https://github.com/jjbrophy47/ibug}.
\end{abstract}

\section{Introduction}

Despite the impressive success of deep learning models on unstructured data~(e.g., images, audio, text), gradient-boosted trees~\cite{friedman2000additive} remain the preferred choice for \emph{tabular} or \emph{structured} data~\cite{prokhorenkova2018catboost}. In fact, Kaggle CEO Anthony Goldbloom recently described gradient-boosted trees as the most ``glaring difference'' between what is used on Kaggle and what is ``fashionable in academia''~\cite{januschowski2021forecasting}.

Our focus is on tabular data for regression tasks, which vary widely from financial~\cite{abu1996introduction} and retail-product forecasting~\cite{makridakis2022m5} to weather~\cite{gneiting2007strictly,gneiting2014probabilistic} and clinic-mortality prediction~\cite{avati2018improving}. 
Gradient-boosted regression trees (GBRTs) are known to make accurate \emph{point predictions}~\cite{makridakis2021m5a} but provide no estimate of the prediction uncertainty,
%Gradient-boosted regression trees (GBRTs) are known to make accurate predictions for these types of problems~\cite{makridakis2021m5a}. However, GBRTs only produce \emph{point predictions}, with no uncertainty about a given prediction. 
%With the push towards more interpretable models~\cite{sharchilev2018finding,lundberg2018consistent,brophy2020trex}, quantifying the uncertainty of a prediction 
which is desirable for both forecasting practitioners~\cite{bose2017probabilistic,taylor2018forecasting} and the explainable AI~(XAI) community~\cite{ferreira2020people,tjoa2020survey,adadi2018peeking} in general.
%
%Traditional solutions \eat{approaches for probabilistic regression in GBRTs} include computationally expensive Bayesian MCMC methods~\cite{chipman2010bart} or training a separate model for each desired quantile~\cite{koenker2001quantile}.
% in fact, the winning solution for the M5 prediction competition~\cite{makridakis2021m5b} trained 126 separate LightGBM models~\cite{makridakis2021m5a}.
Recently, \eat{however,}~\citet{duan2020ngboost},~\citet{sprangers2021probabilistic}, and~\citet{malinin2021uncertainty} introduced NGBoost, PGBM, and CBU~(CatBoost~\cite{prokhorenkova2018catboost} with uncertainty), respectively, new gradient boosting algorithms that provide state-of-the-art probabilistic predictions. However, NGBoost tends to underperform as a point-predictor, and PGBM and CBU are limited in the types of distributions they can use to model the output. %~(see \S\ref{sec:related_work} for all related work).

We introduce a simple yet effective method for enabling \emph{any} GBRT point-prediction model to produce probabilistic predictions. Our proposed approach, \underline{I}nstance-\underline{B}ased \underline{U}ncertainty estimation for \underline{G}radient-boosted regression trees~(\emph{IBUG}), has two key components:~1)~We leverage the fact that GBRTs accurately model the conditional mean and use this point prediction as the mean in a probabilistic forecast; and~2)~We identify the $k$ training examples with the greatest \emph{affinity} to the target instance and use these examples to estimate the uncertainty of the target prediction. We define the affinity between two instances as the number of times both instances appear in the same leaf throughout the ensemble. Thus, our method acts as a wrapper around any given GBRT model.

In experiments on 21 regression benchmark datasets and one synthetic dataset, we demonstrate the effectiveness of IBUG to deliver on par or improved probabilistic performance as compared to existing state-of-the-art methods while maintaining state-of-the-art point-prediction performance. We also show that probabilistic predictions can be improved by applying IBUG to different GBRT models, something that NGBoost and PGBM cannot do.
Additionally, IBUG can use the training instances closest to the target example to directly model the output distribution using any parametric \emph{or} non-parametric distribution, something NGBoost, PGBM, \emph{and} CBU cannot do.
Finally, we show that sampling trees dramatically improves runtime efficiency for computing training-example affinities without having a significant detrimental impact on the resulting probabilistic predictions, allowing IBUG to scale to larger datasets.

\section{Notation \& Background}
\label{sec:background}

We assume an instance space $\mathcal{X} \subseteq \mathbb{R}^p$ and target space $\mathcal{Y} \subset \mathbb{R}$. Let $\mathcal{D} \coloneqq \{(x_i, y_i)\}_{i=1}^{n}$ be a training dataset in which each instance $x_i \in \mathcal{X}$ is a $p$-dimensional vector {\small $(x_{i}^{j})_{j=1}^{p}$} and $y_i \in \mathcal{Y}$.

\subsection{Gradient-Boosted Regression Trees}
\label{sec:gbrts}

Gradient-boosting~\cite{friedman2000additive} is a powerful machine-learning algorithm that iteratively adds weak learners to construct a model~$f: \mathcal{X} \rightarrow \mathbb{R}$ that minimizes some empirical risk~$\mathcal{L}: \mathbb{R} \times \mathbb{R} \rightarrow \mathbb{R}$.
The model is defined by a recursive relationship:~$f_{0}(x) = \gamma$, $\dots$, $f_{t}(x) = f_{t-1}(x) + \eta\ m_{t}(x)$
in which~$f_0$ is the base learner,~$\gamma$ is an initial estimate,~$f_t$ is the model at iteration~$t$,~$m_t$ is the weak learner added during iteration~$t$ to improve the model, and~$\eta$ is the learning rate. 

Gradient-boosted regression trees~(GBRTs) typically choose~$\ell$ to be the mean squared error~(MSE),~$\gamma$ as~$\frac{1}{n}\sum_{i=1}^n y_i$~(mean output of the training instances), and regression trees as weak learners. Each weak learner is typically chosen to approximate the negative gradient~\cite{malinin2021uncertainty}:~$m_{t} = \argmin_{m}\ \frac{1}{n} \sum_{i=1}^{n} (-g_t^i - m(x_i))^2$
% \begin{align}
%     m_{t} &= \argmin_{m}\ \frac{1}{n} \sum_{i=1}^{n} (-g_t^i - m(x_i))^2,
% \end{align}
in which~{\small $g_t^i = \fracpartial{\ell(y_i, \hat{y}_i)}{\hat{y}_i}$} is the functional gradient of the $i$th training instance at iteration~$t$ with respect to~$\hat{y}_i = f_{t-1}(x_i)$.

The weak learner at iteration~$t$ recursively partitions the instance space into $M_t$ disjoint regions {\small $\{{r_{t}^{j}}\}_{j=1}^{M_t}$}. Each region is called a leaf, and the parameter value for leaf~$j$ at tree\footnote{We use the terms \emph{tree} and \emph{iteration} interchangeably.}~$t$ is typically determined~(given a fixed structure) using a one-step Newton-estimation method~\citep{ke2017lightgbm}:~{\small $\theta_{t}^{j} = -\sum_{i \in I_{t}^{j}} g_{t}^i / (\sum_{i \in I_{t}^{j}} h_{t}^i + \lambda$)}
% \begin{align}
% \theta_{t}^{j} &= -\frac{\sum_{i \in I_{t}^{j}} w_i g_{t}^i}{\sum_{i \in I_{t}^{j}} w_i h_{t}^i + \lambda}
% \end{align}
in which {\small $I_{t}^{j} = \{{(x_i, y_i) \mid x_i \in r_{t}^{j}}\}_{i=1}^{n}$} is the instance set of leaf~$j$ for tree~$t$, {\small $h_t^i$} is the second derivative of the $i$th training instance w.r.t.~$\hat{y}_i$,
and $\lambda$ is a regularization constant. Thus, the output of $m_t$ can be written as follows:~$m_{t}(x) = \sum_{j=1}^{M_t} \theta_{t}^{j} \ \mathbbm{1}[x \in r_{t}^{j}]$
% \begin{align}
% m_{t}(x) = \sum_{j=1}^{M_t} \theta_{t}^{j} \ \mathbbm{1}[x \in r_{t}^{j}],
% \end{align}
in which~$\mathbbm{1}$ is the indicator function.
The final GBRT model generates a prediction for a target example $x_{te}$ by summing the values of the leaves $x_{te}$ traverses to across all~$T$ iterations: $\hat{y}_{te} = \sum_{t=1}^T m_t(x_{te})$.

\subsection{Probabilistic Regression}
\label{sec:prob_reg}

Our focus is on probabilistic regression---estimating the conditional probability distribution~$P(y|x)$ for some target variable~$y$ 
%\in \mathcal{Y}$  % DL: technically, a random variable is a mapping from Omega (the outcome space) to some domain (in this case, the real numbers or a subset of that space). A random variable is not a real number, it's a function.
given some input vector~$x \in \mathcal{X}$. Unfortunately, traditional GBRT models only output scalar values. Under a squared-error loss function, these scalar values can be interpreted as the conditional mean in a Gaussian distribution with some~(unknown) constant variance. However, homoscedasticity is a strong assumption and unknown constant variance has little value in a probabilistic prediction; thus, in order to allow heteroscedasticity, the predicted distribution needs at least two parameters to convey both the magnitude and uncertainty of the prediction~\cite{duan2020ngboost}.

Natural Gradient Boosting~(NGBoost) is a recent method by~\citet{duan2020ngboost} that tackles the aforementioned problems by estimating the parameters of a desired distribution using a multi-parameter boosting approach that trains a separate ensemble for each parameter of the distribution. NGBoost employs the natural gradient to be invariant to parameterization, but requires the inversion of many small matrices (each the size of the number of parameters) to do so. Empirically, NGBoost generates state-of-the-art probabilistic predictions, but tends to underperform as a point predictor.

More recently,~\citet{sprangers2021probabilistic} introduced Probabilistic Gradient Boosting Machines~(PGBM), a single model that optimizes for point performance, but can also generate accurate probabilistic predictions. PGBM treats leaf values as stochastic random variables, using sample statistics to model the mean and variance of each leaf value. PGBM estimates the output mean and variance of a target example using the estimated parameters of each leaf it is assigned to. The predicted mean and variance are then used as parameters in a specified distribution to generate a probabilistic prediction. PGBM has been shown to produce state-of-the-art probabilistic predictions; however, computing the necessary leaf statistics during training can be computationally expensive, especially as the number of leaves in the ensemble increases~(see~\S\ref{subsec:sampling_trees}). Also, since only the mean and variance are predicted for a given target example, PGBM is limited to distributions using only location and scale to model the output.

Finally,~\citet{malinin2021uncertainty} introduce CatBoost with uncertainty~(CBU), a method that estimates uncertainty using ensembles of GBRT models. Similar to NGBoost, multiple ensembles are learned to output the mean and variance. However, CBU also constructs a \emph{virtual ensemble}---a set of overlapping partitions of the learned GBRT trees---to estimate the uncertainty of a prediction by taking the mean of the variances output from the virtual ensemble. Their approach uses a recently proposed stochastic gradient Langevin boosting algorithm~\cite{ustimenko2021sglb} to sample from the true posterior via the virtual ensmble~(in the limit); however, their formulation of uncertainty is limited only to the first and second moments, similar to PGBM.

To address the shortcomings of existing approaches, we introduce a simple method that performs well on both point and probabilistic performance, \emph{and} can more flexibly model the output than previous approaches. Our method can also be applied to~\emph{any} GBRT model, adding additional flexibility.

\section{Instance-Based Uncertainty}
\label{sec:IBUG}

Instance-based methods such as $k$-nearest neighbors have been around for decades and have been useful for many different machine learning tasks~\cite{peterson2009k}. However, defining neighbors based on a fixed metric like Euclidean distance may lead to suboptimal performance, especially as the dimensionality of the dataset increases. More recently, it has been shown that random forests can be used as an adaptive nearest neighbors method~\cite{davies2014random,lin2006random} which identifies the most similar examples to a given instance using the learned model structure. This \emph{supervised tree kernel} can more effectively measure the similarity between examples, and has been used for clustering~\cite{moosmann2006fast} and local linear modeling~\cite{bloniarz2016supervised} as well as instance-\cite{brophy2020trex} and feature-based attribution explanations~\cite{plumb2018model}, for example.

In this work, we apply the idea of a supervised tree kernel to help model the \emph{uncertainty} of a given GBRT prediction. Our approach, \emph{\textbf{\underline{I}}nstance-\textbf{\underline{B}}ased \textbf{\underline{U}}ncertainty estimation for \textbf{\underline{G}}radient-boosted regression trees}~(IBUG), identifies the neighborhood of similar training examples to a target example using the structure of the GBRT, and then uses those instances to generate a probabilistic prediction. IBUG works for \emph{any} GBRT, and can more flexibly model the output than competing methods.

\subsection{Identification of High-Affinity Neighbors}
\label{sec:pred_uncertainty}

At its core, IBUG uses the~$k$ training examples with the largest \emph{affinity} to the target example to model the conditional output distribution. Given a GBRT model $f$, we define the affinity between two examples simply as the number of times each instance appears in the same leaf across all trees in~$f$. Thus, the affinity of the $i$th training example~$x_i$ to a target example~$x_{te}$ can be written as:
\begin{align}
\label{eq:affinity}
A(x_i, x_{te}) = \sum_{i=1}^{T} \mathbbm{1}[R_t(x_i) = R_t(x_{te})],
\end{align}
in which~$R_t(x_i)$ is the leaf~$x_i$ is assigned to for tree~$t$. Alg.~\ref{alg:affinity} summarizes the procedure for computing affinity scores for all training examples. This metric clusters similar examples together based on the learned model representation~(i.e., the tree structures). Intuitively, if two examples appear in the same leaf in every tree throughout the ensemble, then both examples are predicted in an identical manner. One may also view Eq.~\eqref{eq:affinity} as an indication of which training examples most often affect the leaf values~$x_{te}$ is assigned to and thus implicitly which examples are likely to have a big effect on the prediction~$\hat{y}_{te}$. This similarity metric is similar to the random forest kernel~\cite{davies2014random}, however, unlike random forests, GBRTs are typically constructed to a shallower depth, resulting in more training examples assigned to the same leaf~(see~\S\ref{app_sec:leaf_density} for additional details about leaf density in GBRTs).

\begin{minipage}[t]{0.46\textwidth}
\begin{algorithm}[H]
\caption{IBUG affinity computation.}
\label{alg:affinity}
\footnotesize
\begin{algorithmic}[1]
\Require Input instance $x \in \mathcal{X}$, GBRT model~$f$.
\Procedure{ComputeAffinities}{$x$, $f$}
\State $A \gets \Vec{0}$
\Comment{Init. train affinities}
\For{$t=1 \ldots T$}
    \Comment{Visit each tree}
    \State Get instance set~$I_t^l$ for leaf~$l = R_{t}(x)$
    \For{$i \in I_{t}^{l}$}
        \Comment{Increment affinities}
        \State $A_{i} \gets A_{i} + 1$
    \EndFor
\EndFor \\
\Return $A$
\EndProcedure
\end{algorithmic}
\end{algorithm}
\end{minipage}
\hfill
\begin{minipage}[t]{0.5\textwidth}
\begin{algorithm}[H]
\caption{IBUG probabilistic prediction.}
\label{alg:predict}
\footnotesize
\begin{algorithmic}[1]
\Require Input $x \in \mathcal{X}$, GBRT model~$f$, $k$ highest-affinity neighbors~\affinsmall, min. variance~$\rho$, variance calibration parameters $\gamma$ and $\delta$, target distribution~$D$.
\Procedure{ProbPredict}{$x$, $f$, \affinsmall, $\rho$, $\gamma$, $\delta$, $D$}
\State $\mu_{\hat{y}} \gets f(x)$
\Comment{GBRT scalar output}
\State $\sigma^2_{\hat{y}} \gets \max (\sigma^2($\affinsmall$),\ \rho)$
\Comment{Ensure $\sigma^2 > 0$}
\State $\sigma^2_{\hat{y}} \gets \gamma \sigma^2_{\hat{y}} + \delta$
\Comment{Var. calibration, Eq.~\eqref{eq:calibrate}} \\
\Return $D(A^{(k)} | \mu_{\hat{y}}, \sigma^2_{\hat{y}})$
\Comment{Eq.~\eqref{eq:pred_any}}
\EndProcedure
\end{algorithmic}
\end{algorithm}
\end{minipage}

\subsection{Modeling the Output Distribution}
\label{sec:posterior}

IBUG has a multitude of choices when modeling the conditional output distribution.
The simplest and most common approach is to model the output assuming a Gaussian distribution~\cite{duan2020ngboost,sprangers2021probabilistic}. We use the scalar output of~$f$:~$\mu_{\hat{y}_{te}} = f(x_{te})$ to model the conditional mean since GBRTs already produce accurate point predictions. Then, we use the~$k$ training instances with the largest affinity to~$x_{te}$---we denote this set~\affinsmall---to compute the variance~\psigsmall.

\paragraph{Calibrating prediction variance.}

The $k$-nearest neighbors generally do a good job of determining the relative uncertainty of different predictions, but on some datasets, the resulting variance is systematically too large or too small. To correct for this, we apply an additional affine transformation before making the prediction:
\begin{align}
\label{eq:calibrate}
\sigma^2_{\hat{y}_{te}} \leftarrow \gamma \sigma^2_{\hat{y}_{te}} + \delta, 
\end{align}
where $\gamma$ and $\delta$ are tuned on validation data after $k$ has been selected. Instead of exhaustively searching over all values of $\gamma$ or $\delta$, we use either the multiplicative factor (tuning $\gamma$ with $\delta=0$) or the additive factor (tuning $\delta$ with $\gamma=1$), and choose between them using their performance on validation data.

We find this simple calibration step consistently improves probabilistic performance for not only IBUG, but competing methods as well, and at a relatively small cost compared to training the model.

\paragraph{Flexible posterior modeling.}

In general, we can generate a probabilistic prediction
using~\pmusmall~and~\psigsmall~for any distribution that uses location and scale (note PGBM and CBU can \emph{only} model these types of distributions).
However, IBUG can additionally use~\affinsmall~to directly fit any continuous distribution~$D$, including those with high-order moments:
\begin{align}
\label{eq:pred_any}
\hat{D}_{te} = D(A^{(k)} | \mu_{\hat{y}_{te}}, \sigma^2_{\hat{y}_{te}}).
\end{align}
Eq.~\eqref{eq:pred_any} is defined such that~$D$ can be fit directly with~\affinsmall~using MLE~(maximum likelihood estimation)~\cite{myung2003tutorial}, or may be fit using~\pmusmall~or~\psigsmall~as fixed parameter values with~\affinsmall~fitting any other parameters of the distribution. Overall, directly fitting all or some additional parameters in~$D$---for example, the shape parameter in a Weibull distribution---is a benefit over PGBM and CBU, which can only optimize for a \emph{global} shape value using a gridsearch-like approach with extra validation data.

Note that NGBoost can model any parameterized distribution, but must specify this choice before training; in contrast, IBUG can optimize this choice \emph{after} training. Additionally, IBUG may choose~$D$ to be a \emph{non-parametric density estimator} such as KDE~(kernel density estimation)~\cite{sheather1991reliable}, which PGBM, CBU, and NGBoost cannot do.

\subsection{Summary}

In summary, Alg.~\ref{alg:predict} provides pseudocode for generating a probabilistic prediction with IBUG. Note Algs.~\ref{alg:affinity}~and~\ref{alg:predict} work for \emph{any} GBRT model, allowing practitioners to employ IBUG to adapt multiple different point predictors into probabilistic estimators and select the model with the best performance. Empirically, we show using different base models for IBUG can result in improved probabilistic performance than using just one~(\S\ref{subsec:base_models}).

IBUG is a nearest neighbors approach and thus seems well-suited to estimating aleatoric uncertainty---remaining uncertainty due to irreducible error or the inherent stochasticity in the system~\cite{hullermeier2021aleatoric}---since it can quantify the range of outcomes to be expected given the observed features. However, we use predictions on held-out data to tune the number of nearest neighbors and the variance calibration hyperparameters; thus, we effectively optimize prediction uncertainty encompassing both aleatoric uncertainty and epistemic uncertainty---error due to the imperfections of the model and the training data~\cite{d2021tale,malinin2021uncertainty}. The evaluation measures in our experiments thus also focus on predictive uncertainty.

\section{Computational Efficiency}

% We now discuss the runtime of IBUG and methods for increasing its efficiency.

\paragraph{Training efficiency.} Since IBUG works with standard GBRT models, it inherits the training efficiency of modern GBRT implementations such as  XGBoost~\citep{chen2016xgboost}, LightGBM~\citep{ke2017lightgbm}, and CatBoost~\citep{prokhorenkova2018catboost}.
% It also benefits from any future developments in training efficiency, with no need to update the IBUG algorithm.

\paragraph{Prediction efficiency.}
If there are $T$ trees in the ensemble and each leaf has at most $n_l$ training instances assigned to it, then IBUG's prediction time is $O(T n_l)$, since it considers each instance in each leaf. Note training instances that do not appear in a leaf with the target instance do not increase prediction time; what matters most is thus the number of instances at each leaf.
% Therefore, what matters most is not the total size of the dataset, but the number of instances at each leaf.
We find LightGBM often induces regression trees with large leaves---in some cases, over half the dataset is assigned to a single leaf~(see~\S\ref{app_sec:leaf_density} for details). Thus, prediction time still grows with the size of the dataset, as is typical for instance-based methods. This higher prediction time is the price IBUG pays for greater flexibility. 

Prediction efficiency can be increased at training time by using deeper GBRTs with fewer instances in each leaf, after training by subsampling the instances considered for predictions, or at prediction time by sampling the trees used to compute affinities. We explore this last option in the next subsection.

% NOTE: Empirically, this doesn't seem to be true...
% Although leaf density will have an impact on efficiency, runtimes may vary considerably depending on~$x_{te}$ and the type of GBRT model~(e.g., layer-wise~\cite{chen2016xgboost,prokhorenkova2018catboost} vs. leaf-wise models~\cite{ke2017lightgbm,scikit-learn}).

\subsection{Sampling Trees}
\label{sec:sampling_trees}

The most expensive operation when generating a probabilistic prediction with IBUG is computing the affinity vector~(Eq.~\ref{eq:affinity}). In order to increase prediction efficiency, we can instead work with a subset of the trees~$\tau < T$ in the ensemble. We can build this subset by sampling trees uniformly at random, taking the first trees learned (representing the largest gradient steps), or the last trees learned (representing the fine-tuning steps).

By sampling trees, the runtime complexity reduces to~$O(\tau n_l)$, which provides significant speedups when~$\tau \ll T$.
% Generally, selecting~$\tau < T$ introduces a tradeoff between computational complexity and predictive accuracy; however,
In our empirical evaluation, we find that taking a subset of the first trees learned generally works best, significantly increasing prediction efficiency while maintaining accurate probabilistic predictions~(\S\ref{subsec:sampling_trees}).

\subsection{Accelerated~$k$ Tuning}
\label{app_sec:accel_k_tuning}

Choosing an appropriate value of~$k$ is critical for generating accurate probabilistic predictions in IBUG. Thus, we aim to tune~$k$ using a held-out validation dataset~$\mathcal{D}_{val} \subset \mathcal{D}$ and an appropriate probabilistic scoring metric such as negative log likelihood~(NLL). Unfortunately, typical tuning procedures would result in the same affinity vectors being computed---an expensive operation---for each candidate value of~$k$. To mitigate this issue, we perform a custom tuning procedure that reuses computed affinity vectors for all values of~$k$. More specifically, IBUG computes an affinity vector~$A$ for a given validation example~$x_{val}$, and then sorts~$A$ in descending order~(i.e., largest affinity first). Then, IBUG takes the top~$k$ training instances, and generates and scores the resulting probabilistic prediction. For each subsequent value of~$k$, the same sorted affinity list can be used, avoiding duplicate computation. We summarize this procedure in~Alg.~\ref{alg:train}.

Once~$k$ is chosen, we may encounter a new unseen target instance in which the variance of the~$k$-highest affinity training examples for that target example is zero or extremely small. In this case, we set the predicted target variance to~$\rho$, which is set during tuning to the minimum~(nonzero) variance computed over all predictions in the validation set for the chosen~$k$. In practice, we find instances of abnormally low variance to be rare with appropriately chosen values of~$k$.

\begin{algorithm}[h]
\caption{IBUG accelerated tuning of~$k$.}
\label{alg:train}
\begin{algorithmic}[1]
\Require Validation dataset $\mathcal{D}_{val} \subset \mathcal{D}$, GBRT model~$f$, list of candidates~$K$, target distribution~$D$, probabilistic scoring metric~$V$, minimum variance~$\rho=1\mathrm{e}{-15}$.
\Procedure{FastTuneK}{$\mathcal{D}_{val}$, $f$, $K$, $D$, $V$, $\rho$}
\For{$(x_j, y_j) \in \mathcal{D}_{val}$}
    \State $A \gets$ \text{\sc ComputeAffinities($x_j,f$)}
    \Comment{Algorithm~\ref{alg:affinity}}
    \State $A \gets$ \text{Argsort $A$ in descending order}
    \For{$k \in K$}
    \Comment{Use same ordering for each~$k$}
        \State $A^{(k)} \gets$ \text{Take first~$k$ training instances($A, k$)}
        \State $\hat{D}_{y_j}^k \gets$ \text{\sc ProbPredict($x_j, f, A^{(k)}, \rho, 1, 0, D$)}
        \Comment{Algorithm~\ref{alg:predict}}
        \State $S_j^k \gets V(y_j, \hat{D}_{y_j}^k)$
        \Comment{Save validation score}
    \EndFor
\EndFor
\State $k \gets$ \text{Select best $k$ from $S$}
\State $\rho \gets$ \text{Select minimum $\sigma^2$ from $\hat{D}^k$} \\
\Return $k$, $\rho$
\EndProcedure
\end{algorithmic}
\end{algorithm}

\section{Experiments}
\label{sec:experiments}

In this section, we demonstrate IBUG's ability to produce competitive probabilistic and point predictions as compared to current state-of-the-art methods on a large set of regression datasets~(\S\ref{sec:methodology},~\S\ref{sec:performance_results}). Then, we show that IBUG can use different base models to improve probabilistic performance~(\S\ref{subsec:base_models}), flexibly model the posterior distribution~(\S\ref{subsec:posterior}), and use approximations to speed up probabilistic predictions while maintaining competitive performance~(\S\ref{subsec:sampling_trees}).

\paragraph{Implementation and Reproducibility.}

We implement IBUG in Python, using Cython---a Python package allowing the development of C extensions---to store a unified representation of the model structure. IBUG supports all modern gradient boosting frameworks including XGBoost~\citep{chen2016xgboost}, LightGBM~\citep{ke2017lightgbm}, and CatBoost~\citep{prokhorenkova2018catboost}. Experiments are run on publicly available datasets using an Intel(R) Xeon(R) CPU E5-2690 v4 @ 2.6GHz with 60GB of RAM @ 2.4GHz. Links to all data sources as well as the code for IBUG and all experiments is available at~\url{https://github.com/jjbrophy47/ibug}.

\subsection{Methodology}
\label{sec:methodology}

We now compare IBUG's probabilistic and point predictions to NGBoost~\cite{duan2020ngboost}, PGBM~\cite{sprangers2021probabilistic}, and CBU~\cite{malinin2021uncertainty} on 21 benchmark regression datasets and one synthetic dataset. Additional dataset details are in~\S\ref{app_sec:datasets}.

\paragraph{Metrics.}

We compute the average continuous ranked probability score~(CRPS~$\downarrow$) and negative log likelihood~(NLL~$\downarrow$)~\cite{gneiting2007strictly,zamo2018estimation} over the test set to evaluate probabilistic performance. To evaluate point performance, we use root mean squared error~(RMSE~$\downarrow$). For all metrics, lower is better. See~\S\ref{app_sec:reproducibility} for detailed descriptions.

\paragraph{Protocol.}

We follow a similar protocol to~\citet{sprangers2021probabilistic} and~\citet{duan2020ngboost}. We use 10-fold cross-validation to create 10 90/10 train/test folds for each dataset. For each fold, the 90\% training set is randomly split into an 80/20 train/validation set to tune any hyperparameters. Once the hyperparameters are tuned, 
% with the lowest validation score are selected,
the model is retrained using the entire 90\% training set. For probabilistic predictions, a normal distribution is used to model the output.

\paragraph{Significance Testing.}

We report counts of the number of datasets in which a given method performed better (``Win''), worse (``Loss''), or not statistically different~(``Tie'') relative to a comparator using a two-sided paired t-test over the 10 random folds with a significance level of 0.05.

\paragraph{Hyperparameters.}

We tune NGBoost the same way as in~\citet{duan2020ngboost}. Since PGBM, CBU, and IBUG optimize a point prediction metric, we tune their hyperparameters similarly.
% For KNN, we tune two different~$k$ values:~one that optimizes point performance, and another that optimizes probabilistic performance.
We also tune variance calibration parameters $\gamma$ and $\delta$ for each method~(\S\ref{sec:posterior}).
Exact hyperparameter values evaluated and selected are in~\S\ref{app_sec:hyperparameters}. Unless specified otherwise, we use CatBoost~\cite{prokhorenkova2018catboost} as the base model for IBUG.

\begin{table}[t]
\caption{Probabilistic~(CRPS) performance for each method on each dataset. Lower is better. Normal distributions are used for all probabilistic predictions. Results are averaged over 10 folds, and standard errors are shown in subscripted parentheses. The best method for each dataset is bolded, as well as those with standard errors that overlap the best method. \emph{Bottom row}: Head-to-head comparison between IBUG/IBUG+CBU and each method showing the number of wins, ties, and losses~(W-T-L) across all datasets. On average, IBUG+CBU provides the most accurate probabilistic predictions.}
\label{tab:results_crps}
\centering
\vspace{0.5em}
\begin{tabular}{lccccc}
\toprule
Dataset & NGBoost & PGBM & CBU & IBUG & IBUG+CBU \\
\midrule
Ames & 38346$_{(547)}$ & 10872$_{(355)}$ & 11008$_{(330)}$ & {\bfseries 10434}$_{(367)}$ & {\bfseries 10194}$_{(368)}$ \\
Bike & 12.4$_{(0.955)}$ & 1.183$_{(0.041)}$ & 0.833$_{(0.036)}$ & 0.974$_{(0.048)}$ & {\bfseries 0.766}$_{(0.032)}$ \\
California & 1e11$_{(1e11)}$ & 0.222$_{(0.001)}$ & 0.217$_{(0.001)}$ & 0.213$_{(0.001)}$ & {\bfseries 0.207}$_{(0.001)}$ \\
Communities & 0.068$_{(0.002)}$ & 0.068$_{(0.002)}$ & 0.067$_{(0.002)}$ & {\bfseries 0.065}$_{(0.002)}$ & {\bfseries 0.065}$_{(0.002)}$ \\
Concrete & 3.410$_{(0.182)}$ & 1.927$_{(0.086)}$ & {\bfseries 1.788}$_{(0.077)}$ & 1.849$_{(0.098)}$ & {\bfseries 1.741}$_{(0.082)}$ \\
Energy & 0.519$_{(0.043)}$ & {\bfseries 0.147}$_{(0.006)}$ & 0.196$_{(0.009)}$ & {\bfseries 0.143}$_{(0.009)}$ & 0.157$_{(0.008)}$ \\
Facebook & 4.022$_{(0.099)}$ & 3.554$_{(0.095)}$ & 3.211$_{(0.059)}$ & 3.073$_{(0.066)}$ & {\bfseries 2.977}$_{(0.070)}$ \\
Kin8nm & 0.095$_{(0.001)}$ & 0.061$_{(0.001)}$ & 0.057$_{(0.001)}$ & {\bfseries 0.051}$_{(0.001)}$ & {\bfseries 0.051}$_{(0.001)}$ \\
Life & 2.897$_{(1.465)}$ & 0.815$_{(0.027)}$ & 0.772$_{(0.024)}$ & 0.794$_{(0.023)}$ & {\bfseries 0.731}$_{(0.022)}$ \\
MEPS & {\bfseries 5.527}$_{(0.196)}$ & 6.448$_{(0.092)}$ & 6.050$_{(0.109)}$ & 6.150$_{(0.114)}$ & 6.016$_{(0.113)}$ \\
MSD & 4.524$_{(0.005)}$ & 4.576$_{(0.005)}$ & 4.363$_{(0.004)}$ & 4.410$_{(0.005)}$ & {\bfseries 4.347}$_{(0.004)}$ \\
Naval & 0.003$_{(0.000)}$ & 0.000$_{(0.000)}$ & 0.000$_{(0.000)}$ & 0.000$_{(0.000)}$ & {\bfseries 0.000}$_{(0.000)}$ \\
News & {\bfseries 2191}$_{(47.5)}$ & 2361$_{(52.6)}$ & 2346$_{(52.6)}$ & 2545$_{(41.0)}$ & 2380$_{(52.1)}$ \\
Obesity & 3.208$_{(0.028)}$ & 1.860$_{(0.022)}$ & {\bfseries 1.740}$_{(0.017)}$ & 1.866$_{(0.021)}$ & 1.771$_{(0.019)}$ \\
Power & 2.105$_{(0.023)}$ & 1.531$_{(0.019)}$ & {\bfseries 1.473}$_{(0.022)}$ & 1.542$_{(0.020)}$ & {\bfseries 1.471}$_{(0.021)}$ \\
Protein & 5427$_{(5409)}$ & 1.823$_{(0.011)}$ & 1.788$_{(0.009)}$ & 1.784$_{(0.008)}$ & {\bfseries 1.742}$_{(0.009)}$ \\
STAR & 132$_{(1.589)}$ & {\bfseries 131}$_{(1.380)}$ & {\bfseries 130}$_{(1.283)}$ & {\bfseries 130}$_{(1.214)}$ & {\bfseries 129}$_{(1.198)}$ \\
Superconductor & 2.405$_{(0.028)}$ & {\bfseries 0.126}$_{(0.004)}$ & 0.150$_{(0.004)}$ & 0.153$_{(0.006)}$ & {\bfseries 0.128}$_{(0.004)}$ \\
Synthetic & 5.779$_{(0.042)}$ & {\bfseries 5.737}$_{(0.039)}$ & {\bfseries 5.739}$_{(0.040)}$ & {\bfseries 5.731}$_{(0.040)}$ & {\bfseries 5.730}$_{(0.040)}$ \\
Wave & 571020$_{(883)}$ & 3891$_{(73.9)}$ & 2349$_{(10.3)}$ & 2679$_{(16.0)}$ & {\bfseries 2026}$_{(9.538)}$ \\
Wine & 0.385$_{(0.005)}$ & {\bfseries 0.323}$_{(0.005)}$ & 0.337$_{(0.006)}$ & {\bfseries 0.322}$_{(0.006)}$ & {\bfseries 0.321}$_{(0.006)}$ \\
Yacht & 1.177$_{(0.158)}$ & {\bfseries 0.292}$_{(0.042)}$ & {\bfseries 0.281}$_{(0.048)}$ & {\bfseries 0.276}$_{(0.048)}$ & {\bfseries 0.255}$_{(0.046)}$ \\
\midrule
% NGBoost W-T-L & - & 3-4-15 & 2-4-16 & 2-3-17 & 2-3-17 \\
% PGBM W-T-L & 15-4-3 & - & 3-7-12 & 2-9-11 & 1-6-15 \\
% CBU W-T-L & 16-4-2 & 12-7-3 & - & 8-5-9 & 2-2-18 \\
IBUG W-T-L & 17-3-2 & 11-9-2 & 9-5-8 & - & 1-6-15 \\
IBUG+CBU W-T-L & 17-3-2 & 15-6-1 & 18-2-2 & 15-6-1 & - \\
\bottomrule
\end{tabular}
\end{table}

\subsection{Probabilistic and Point Performance}
\label{sec:performance_results}

We first compare IBUG's probabilistic and point predictions to each baseline on each dataset. See Table~\ref{tab:results_crps} for detailed CRPS results; due to space constraints, results for additional probabilistic metrics~(e.g., NLL) as well as point performance results are in~\S\ref{app_sec:additional_metrics}. Our main findings are as follows:
\begin{itemize}
    \item On probabilistic performance, IBUG performs equally well or better than NGBoost and PGBM, winning on 17 and 11~(out of 22) datasets respectively, while losing on only 2 and 2~(respectively). Since CBU and IBUG performance is similar, we combine the two approaches, averaging their outputs; we denote this simple ensemble \emph{IBUG+CBU}. Surprisingly, IBUG+CBU works very well, losing on only a maximum of 2 datasets when faced head-to-head against any other method; these results suggest IBUG and CBU are complimentary approaches.
    
    \item On point performance, PGBM, CBU, and IBUG performed significantly better than NGBoost; this is consistent with previous work and is perhaps unsurprising since NGBoost is optimized for probabilistic performance, not point performance. However, IBUG generally performed better than PGBM, winning on 13 datasets and losing on only 1 dataset; and performed slightly better than CBU, winning on 6 datasets with no losses.
\end{itemize}

We also compare IBUG with two additional baselines---$k$NN and BART~\cite{chipman2010bart}---shown in \S\ref{app_sec:knn}--\ref{app_sec:bart} due to space constraints. We find IBUG generally outperforms these methods in both probabilistic and point performance. Overall, the results in this section suggest IBUG generates both competitive probabilistic and point predictions compared to existing methods.

\subsection{Different Base Models}
\label{subsec:base_models}

\begin{table}[t]
    \caption{Probabilistic~(CRPS, NLL) performance on the test set for IBUG using different base models. Results are averaged over 10 folds, and standard errors are shown in subscripted parentheses; lower is better. On 6 and 5 datasets, respectively, either IBUG-LightGBM or IBUG-XGBoost significantly outperforms IBUG-CatBoost on the validation set and subsequently on the test set, demonstrating the potential for improved probabilistic performance by using IBUG with different base models.}
    \label{tab:base_model}
    \vspace{0.5em}
    \centering
    \scriptsize
    \begin{tabular}{@{}lc|cc@{}}
        \toprule
        & \multicolumn{3}{c}{Test CRPS~($\downarrow$)} \\
        \cmidrule(lr){2-4}
        Dataset & CatBoost & LightGBM & XGBoost \\
        \midrule
% Ames & {\bfseries 10434}$_{(367)}$ & 10809$_{(303)}$ & {\bfseries 10830}$_{(462)}$ \\
Bike & 0.974$_{(0.048)}$ & {\bfseries 0.819}$_{(0.024)}$ & {\bfseries 0.849}$_{(0.012)}$ \\
% California & {\bfseries 0.213}$_{(0.001)}$ & {\bfseries 0.213}$_{(0.001)}$ & 0.218$_{(0.001)}$ \\
% Communities & {\bfseries 0.065}$_{(0.002)}$ & {\bfseries 0.067}$_{(0.002)}$ & {\bfseries 0.066}$_{(0.002)}$ \\
% Concrete & {\bfseries 1.849}$_{(0.098)}$ & 1.964$_{(0.079)}$ & 1.969$_{(0.086)}$ \\
% Energy & {\bfseries 0.143}$_{(0.009)}$ & {\bfseries 0.140}$_{(0.008)}$ & 0.151$_{(0.008)}$ \\
% Facebook & 3.073$_{(0.066)}$ & {\bfseries 2.993}$_{(0.064)}$ & 3.067$_{(0.072)}$ \\
% Kin8nm & {\bfseries 0.051}$_{(0.001)}$ & 0.059$_{(0.001)}$ & 0.065$_{(0.001)}$ \\
% Life & {\bfseries 0.794}$_{(0.023)}$ & {\bfseries 0.772}$_{(0.027)}$ & {\bfseries 0.782}$_{(0.023)}$ \\
% MEPS & {\bfseries 6.150}$_{(0.114)}$ & {\bfseries 6.282}$_{(0.159)}$ & {\bfseries 6.148}$_{(0.117)}$ \\
MSD & 4.410$_{(0.005)}$ & {\bfseries 4.372}$_{(0.005)}$ & 4.418$_{(0.005)}$ \\
% Naval & {\bfseries 0.000}$_{(0.000)}$ & 0.000$_{(0.000)}$ & 0.000$_{(0.000)}$ \\
News & 2545$_{(41.0)}$ & {\bfseries 2436}$_{(50.8)}$ & 2551$_{(56.0)}$ \\
% Obesity & {\bfseries 1.866}$_{(0.021)}$ & {\bfseries 1.849}$_{(0.024)}$ & 1.897$_{(0.018)}$ \\
Power & 1.542$_{(0.020)}$ & 1.536$_{(0.022)}$ & {\bfseries 1.518}$_{(0.018)}$ \\
Protein & 1.784$_{(0.008)}$ & {\bfseries 1.683}$_{(0.009)}$ & 1.788$_{(0.008)}$ \\
% STAR & {\bfseries 130}$_{(1.214)}$ & 131$_{(1.508)}$ & {\bfseries 130}$_{(1.538)}$ \\
Supercon. & 0.153$_{(0.006)}$ & {\bfseries0.090}$_{(0.005)}$ & {\bfseries 0.010}$_{(0.003)}$ \\
% Synthetic & {\bfseries 5.731}$_{(0.040)}$ & {\bfseries 5.733}$_{(0.039)}$ & {\bfseries 5.739}$_{(0.041)}$ \\
% Wave & {\bfseries 2679}$_{(16.0)}$ & 4291$_{(16.0)}$ & 3514$_{(16.1)}$ \\
% Wine & {\bfseries 0.322}$_{(0.006)}$ & {\bfseries 0.321}$_{(0.006)}$ & {\bfseries 0.324}$_{(0.005)}$ \\
% Yacht & 0.276$_{(0.048)}$ & 0.490$_{(0.085)}$ & {\bfseries 0.227}$_{(0.031)}$ \\
% IBUG-CB W-T-L & - & 4-12-6 & 6-13-3 \\
% IBUG-LGB W-T-L & 6-12-4 & - & 7-12-3 \\
% IBUG-XGB W-T-L & 3-13-6 & 3-12-7 & - \\
        \bottomrule
    \end{tabular}
\quad
    \centering
    \begin{tabular}{@{}lc|cc@{}}
        \toprule
        & \multicolumn{3}{c}{Test NLL~($\downarrow$)} \\
        \cmidrule(lr){2-4}
        Dataset & CatBoost & LightGBM & XGBoost \\
        \midrule
% Ames & {\bfseries 11.2}$_{(0.030)}$ & {\bfseries 11.2}$_{(0.028)}$ & {\bfseries 11.2}$_{(0.036)}$ \\
Bike & 1.886$_{(0.056)}$ & {\bfseries 1.292}$_{(0.048)}$ & {\bfseries 1.662}$_{(0.024)}$ \\
% California & 0.477$_{(0.010)}$ & {\bfseries 0.454}$_{(0.012)}$ & 0.476$_{(0.009)}$ \\
% Communities & {\bfseries -0.639}$_{(0.135)}$ & {\bfseries -0.633}$_{(0.105)}$ & {\bfseries -0.601}$_{(0.153)}$ \\
% Concrete & {\bfseries 2.980}$_{(0.146)}$ & 3.418$_{(0.255)}$ & {\bfseries 3.105}$_{(0.192)}$ \\
% Energy & {\bfseries 1.644}$_{(0.514)}$ & {\bfseries 1.483}$_{(0.613)}$ & {\bfseries 1.453}$_{(0.450)}$ \\
% Facebook & 2.175$_{(0.067)}$ & {\bfseries 2.063}$_{(0.070)}$ & {\bfseries 2.061}$_{(0.038)}$ \\
% Kin8nm & {\bfseries -0.841}$_{(0.008)}$ & -0.762$_{(0.008)}$ & -0.691$_{(0.007)}$ \\
% Life & 1.858$_{(0.033)}$ & {\bfseries 1.824}$_{(0.049)}$ & {\bfseries 1.917}$_{(0.103)}$ \\
% MEPS & {\bfseries 3.793}$_{(0.052)}$ & 3.871$_{(0.043)}$ & {\bfseries 3.775}$_{(0.039)}$ \\
MSD & 3.415$_{(0.002)}$ & {\bfseries 3.409}$_{(0.002)}$ & {\bfseries 3.402}$_{(0.002)}$ \\
Naval & -6.208$_{(0.010)}$ & {\bfseries -6.281}$_{(0.007)}$ & -5.853$_{(0.014)}$ \\
% News & {\bfseries 10.6}$_{(0.208)}$ & {\bfseries 10.6}$_{(0.171)}$ & {\bfseries 10.6}$_{(0.218)}$ \\
Obesity & 2.646$_{(0.009)}$ & {\bfseries 2.593}$_{(0.016)}$ & {\bfseries 2.624}$_{(0.010)}$ \\
% Power & {\bfseries 2.575}$_{(0.036)}$ & {\bfseries 2.554}$_{(0.029)}$ & {\bfseries 2.543}$_{(0.032)}$ \\
% Protein & {\bfseries 2.653}$_{(0.054)}$ & {\bfseries 2.608}$_{(0.047)}$ & {\bfseries 2.609}$_{(0.022)}$ \\
% STAR & {\bfseries 6.853}$_{(0.008)}$ & {\bfseries 6.860}$_{(0.011)}$ & {\bfseries 6.856}$_{(0.011)}$ \\
Supercon. & 0.783$_{(0.181)}$ & {\bfseries -0.496}$_{(0.169)}$ & 20.4$_{(23.2)}$ \\
% Synthetic & {\bfseries 3.738}$_{(0.007)}$ & {\bfseries 3.739}$_{(0.007)}$ & {\bfseries 3.740}$_{(0.007)}$ \\
% Wave & {\bfseries 10.5}$_{(0.030)}$ & 10.7$_{(0.007)}$ & 10.5$_{(0.010)}$ \\
% Wine & {\bfseries 0.910}$_{(0.016)}$ & {\bfseries 0.910}$_{(0.016)}$ & {\bfseries 0.917}$_{(0.014)}$ \\
% Yacht & {\bfseries 1.799}$_{(1.307)}$ & {\bfseries 0.939}$_{(0.290)}$ & {\bfseries 1.726}$_{(1.350)}$ \\
% IBUG-CB W-T-L & - & 3-12-7 & 2-17-3 \\
% IBUG-LGB W-T-L & 7-12-3 & - & 4-16-2 \\
% IBUG-XGB W-T-L & 3-17-2 & 2-16-4 & - \\
        \bottomrule
    \end{tabular}
\end{table}

Here we experiment using different base models for IBUG besides CatBoost~\cite{prokhorenkova2018catboost}; specifically, we use LightGBM~\cite{ke2017lightgbm} and XGBoost~\cite{chen2016xgboost}, two popular gradient boosting frameworks. Table~\ref{tab:base_model} shows that using a different base model can result in improved probabilistic performance. This highlights IBUG's agnosticism to GBRT type, enabling practitioners to apply IBUG to future models with improved point prediction performance.

\begin{figure*}[t]
\centering
\includegraphics[width=\textwidth]{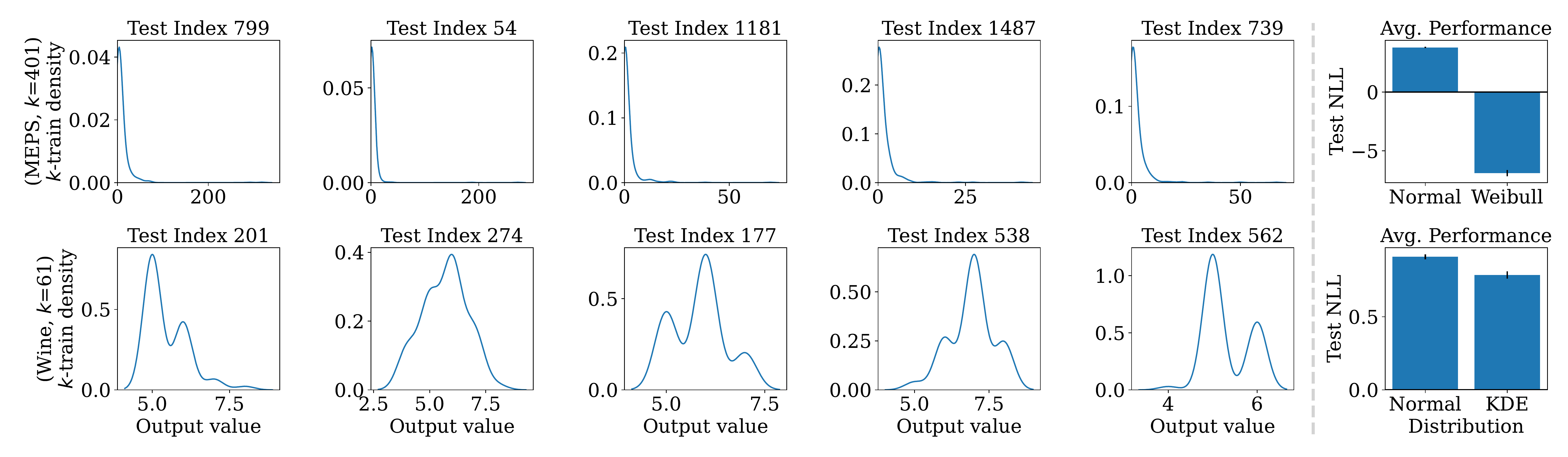}
\caption{\emph{Left}: Distribution of the~$k$-nearest training examples for 5 randomly-selected test instances from the MEPS~(top) and Wine~(bottom) datasets. \emph{Right}: Test NLL~(with standard error) when modeling the posterior using two different distributions~(lower is better). IBUG can model parametric \emph{and} non-parametric distributions that better fit the underlying data than assuming normality.}
\label{fig:posterior}
\end{figure*}

\subsection{Posterior Modeling}
\label{subsec:posterior}

One of the unique benefits of IBUG is the ability to directly model the output using empirical samples~(Figure~\ref{fig:posterior}), giving practitioners a better sense of the output distribution for specific predictions. IBUG can optimize a distribution \emph{after} training, and has more flexibility in the types of distributions it can model---from distributions using just location and scale to those with high-order moments as well as non-parametric density estimators. To test this flexibility, we model each probabilistic prediction using the following distributions:~normal, skewnormal, lognormal, Laplace, student t, logistic, Gumbel, Weibull, and KDE; we then select the distribution with the best average NLL on the validation set, and evaluate its probabilistic performance on the test set.

Figure~\ref{fig:posterior} demonstrates that the selected distributions for the MEPS and Wine datasets achieve better probabilistic performance than assuming normality. Qualitatively, the empirical densities of~\affinsmall~for a randomly sampled set of test instances reaffirms the selected distributions.
As an additional comparison, we report CBU achieves a test NLL of~$3.699_{\pm 0.038}$ and~$1.025_{\pm 0.028}$ for the MEPS and wine datasets~(respectively) using a normal distribution, while IBUG achieves $-6.887_{\pm 0.260}$ and $0.785_{\pm 0.025}$ using Weibull and KDE estimation~(respectively).
For the MEPS dataset, the selected Weibull distribution takes a shape parameter, which IBUG estimates directly on a \emph{per prediction} basis using~\affinsmall and MLE. In contrast, PGBM or CBU would need to optimize a global shape value using a validation set, which is likely to be suboptimal for individual predictions.

\subsection{Variance Calibration}

Table~\ref{tab:delta} shows probabilistic performance comparisons of each method against itself with and without variance calibration. In all cases, variance calibration~(\S\ref{sec:posterior}) either maintains or improves performance for all methods, especially CBU.
Overall, these results suggest that variance calibration should be a standard procedure for probabilistic prediction, unless using a method that has particularly well-calibrated predictions to begin with. We therefore use variance calibration in all of our results.

Additionally,~\S\ref{app_sec:no_calibration} shows performance results for all methods \emph{without} variance calibration. Overall, we observe similar relative performance trends as when applying calibration~(Table~\ref{tab:results_crps}).

\begin{table}[tb]
    \caption{Probabilistic performance comparison of each method with vs. without variance calibration. In all cases, calibration maintains or improves performance; it is especially helpful for CBU.}
    \vspace{0.5em}
    \label{tab:delta}
    \centering
    \small % FIXME FIXME FIXME
    \begin{tabular}{@{}lccc|ccc@{}}
        \toprule
        & \multicolumn{3}{c}{CRPS} & \multicolumn{3}{c}{NLL} \\
        \cmidrule(lr){2-4}\cmidrule(lr){5-7}
        Method & Wins & ties & Losses & Wins & Ties & Losses \\
        \midrule
        % KNN & 4 & 18 & 0
        %     & 8 & 14  & 0 \\
        NGBoost & 9  & 13 & 0
                & 1 & 21 & 0 \\
        PGBM & 13 & 9  & 0
             & 11 & 11 & 0 \\
        CBU  & 17 & 5 & 0
             & 11 & 11 & 0 \\
        IBUG & 13 & 9 & 0
             & 5 & 17 & 0 \\
        % IBUG+CBU & 10 & 10 & 2
        %      & 6 & 12 & 4 \\
        \bottomrule
    \end{tabular}
\end{table}

\begin{figure*}[t]
\centering
    \includegraphics[width=1.0\linewidth]{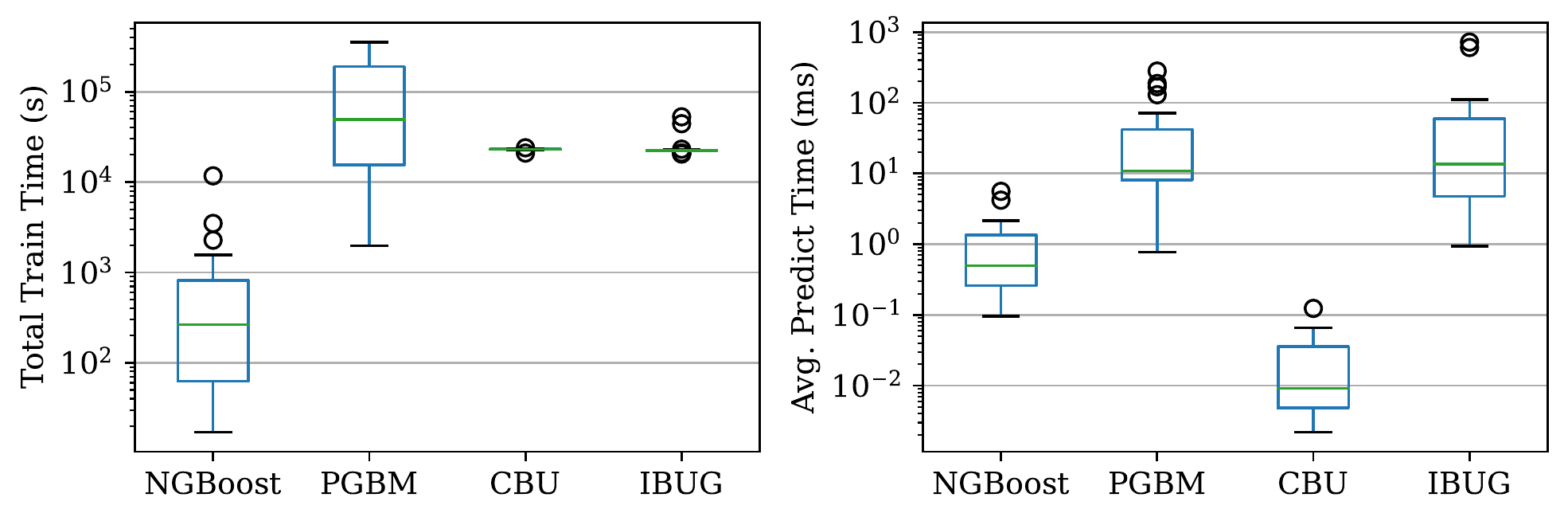}
    \caption{Runtime comparison. \emph{Left}: Total train time (including tuning). \emph{Right}: Average prediction time per test example. Results are shown for all datasets, averaged over 10 folds~(exact values are in~\S\ref{app_sec:runtime}, Tables~\ref{app_tab:runtime_train} and~\ref{app_tab:runtime_prediction}). On average, IBUG has comparable training times to PGBM and CBU, but is relatively slow for prediction.}
\label{fig:runtime}
\end{figure*}

\begin{figure*}[ht]
\centering
\includegraphics[width=\textwidth]{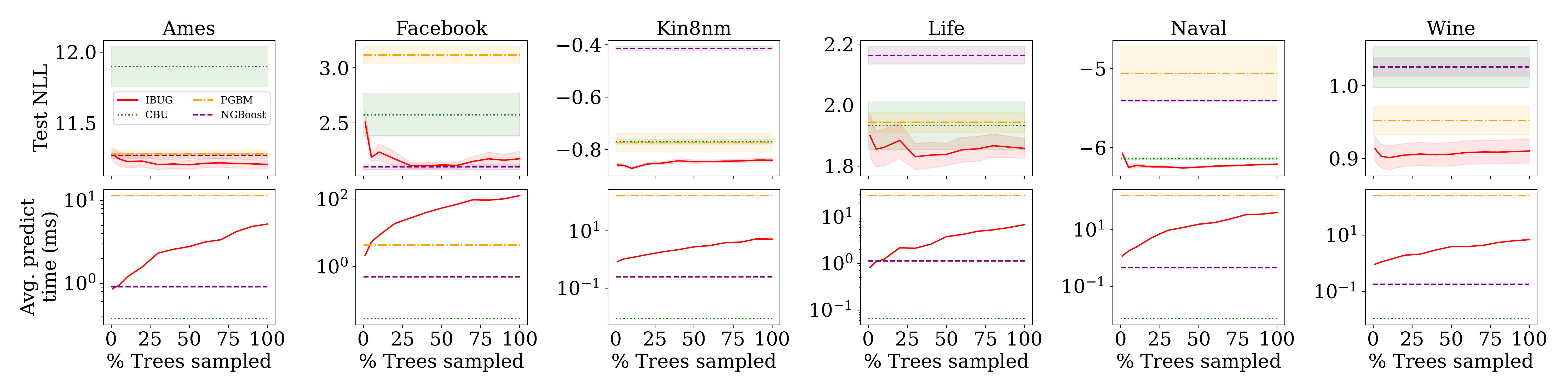}
\caption{Change in probabilistic~(NLL) performance~(top) and average prediction time~(in seconds) per test example~(bottom) as a function of~$\tau$ for six datasets with trees sampled \emph{first-to-last}; lower is better. NGBoost, PGBM, and CBU are added for additional context. The shaded regions represent the standard error. Overall, average prediction time decreases significantly as~$\tau$ decreases while test NLL often remains relatively stable, enabling IBUG to generate probabilistic predictions with significant increased efficiency.
}
\label{fig:tree_frac}
\end{figure*}

\subsection{Sampling Trees}
\label{subsec:sampling_trees}

Figure~\ref{fig:runtime} shows the runtime for each method broken down into total training time (including tuning) and prediction time per test example. On average, IBUG has similar training times to PGBM and CBU, but on some datasets, IBUG is roughly an order of magnitude faster than PGBM. For predictions, IBUG is similar to PGBM but relatively slow compared to NGBoost and CBU.

However, by sampling~$\tau < T$ trees when computing the affinity vector, IBUG can significantly reduce prediction time. Figure~\ref{fig:tree_frac} shows results when sampling trees first-to-last, which typically works best over all tree-sampling strategies~(alternate sampling strategies are evaluated in~\S\ref{app_sec:tree_sampling}).
As~$\tau$ decreases, we observe average prediction time decreases roughly 1-2 orders of magnitude while probabilistic performance remains relatively stable until~$\tau/T$ reaches roughly 1--5\%, at which point probabilistic performance sometimes starts to decrease more rapidly. Note for the Ames and Life datasets, IBUG can reach the same average prediction time as NGBoost while maintaining the same or better probabilistic performance than NGBoost, PGBM, and CBU. These results demonstrate that if speed is a concern, IBUG can approximate the affinity computation to speed up prediction times while maintaining competitive probabilistic performance.

\section{Additonal Related Work}\label{sec:related_work}

Traditional approaches to probabilistic regression include generalized additive models for location, scale, and shape~(GAMLSS), which allow for a flexible choice of distribution for the target variable but are restricted to pre-specified model forms~\cite{rigby2005generalized}. Prophet~\cite{taylor2018forecasting} also produces probabilistic estimates for generalized additive models, but has been shown to underperform as compared to more recent approaches~\cite{sen2019think,alexandrov2020gluonts}. Bayesian methods~\cite{neal1996bayesian,graves2011practical} naturally generate uncertainty estimates by integrating over the posterior; however, exact solutions are limited to simple models, and more complex models such as Bayesian Additive Regression Trees~(BART)~\cite{chipman2010bart,luo2021bast} require computationally expensive sampling techniques~(e.g., MCMC~\cite{andrieu2003introduction}) to provide approximate solutions.

Other approaches to probabilistic regression tasks include conformal predictions~\cite{shafer2008tutorial,taieb2015probabilistic,angelopoulos2021gentle} which produce confidence intervals via empirical errors obtained in the past, and quantile regression~\cite{hasson2021probabilistic,koenker2001quantile,meinshausen2006quantile,romano2019conformalized}.
% which aim to predict the quantile of a given test instance.
Similar to PGBM, distributional forests~(DFs)~\cite{schlosser2019distributional} estimate distributional parameters in each leaf, and average these estimates over all trees in the forest.
% However, DFs are variants of random forests~(RFs), and GBRTs are known to regularly outperform RFs on regression problems, making this approach less suitable for high-performance tasks.
% Recent interest in
Deep learning approaches for probabilistic regression~\cite{rangapuram2018deep,wang2019deep,alexandrov2020gluonts} have increased recently, with notable approaches such as DeepAR~\cite{salinas2020deepar} and methods based on transformer architectures~\cite{li2019enhancing,lim2021temporal}.
% However, despite the success of deep learning models on perceptual tasks~(e.g., audio, image, text), gradient-boosted trees are typically still the preferred model of choice when used in practice, especially when the data is structured~(tabular) or limited~\cite{makridakis2021m5a,makridakis2022m5}.

\section{Conclusion}\label{sec:conclusion}

IBUG uses ideas from instance-based learning to enable probabilistic predictions for \emph{any} GBRT point predictor. IBUG generates probabilistic predictions by using the~$k$-nearest training instances to the test instance found using the structure of the trees in the ensemble. Our results on 22 regression datasets demonstrate this simple wrapper produces competitive probabilistic and point predictions to current state-of-the-art methods, most notably NGBoost~\cite{duan2020ngboost}, PGBM~\cite{sprangers2021probabilistic}, and CBU~\cite{malinin2021uncertainty}. We also show that IBUG can more flexibly model the posterior distribution of a prediction using any parametric \emph{or} non-parametric density estimator.
IBUG's one limitation is relatively slow prediction time. However, we show that approximations in the search for the~$k$-nearest training instances can significantly speed up prediction time; predictions are also easily parallelizable in IBUG. For future work, we plan to investigate other approximations to the affinity vector such as subsampling or even reweighting training instances, which may lead to significant speed ups of IBUG.

\begin{ack}
We would like to thank Zayd Hammoudeh for useful discussions and feedback and the reviewers for their constructive comments that improved this paper.
This work was supported by a grant from the Air Force Research Laboratory and the Defense Advanced Research Projects Agency (DARPA)---agreement number FA8750-16-C-0166, subcontract K001892-00-S05, as well as a second grant from DARPA, agreement number HR00112090135. This work benefited from access to the University of Oregon high-performance computer, Talapas.
\end{ack}

\setcitestyle{numbers}
\bibliographystyle{plainnat}
\bibliography{main}

\begin{thebibliography}{76}
\providecommand{\natexlab}[1]{#1}
\providecommand{\url}[1]{\texttt{#1}}
\expandafter\ifx\csname urlstyle\endcsname\relax
  \providecommand{\doi}[1]{doi: #1}\else
  \providecommand{\doi}{doi: \begingroup \urlstyle{rm}\Url}\fi

\bibitem[Abu-Mostafa and Atiya(1996)]{abu1996introduction}
Yaser~S Abu-Mostafa and Amir~F Atiya.
\newblock Introduction to financial forecasting.
\newblock \emph{Applied Intelligence}, 6\penalty0 (3):\penalty0 205--213, 1996.

\bibitem[Adadi and Berrada(2018)]{adadi2018peeking}
Amina Adadi and Mohammed Berrada.
\newblock Peeking inside the black-box: A survey on explainable artificial
  intelligence ({XAI}).
\newblock \emph{IEEE Access}, 6:\penalty0 52138--52160, 2018.

\bibitem[Alexandrov et~al.(2020)Alexandrov, Benidis, Bohlke-Schneider,
  Flunkert, Gasthaus, Januschowski, Maddix, Rangapuram, Salinas, Schulz,
  et~al.]{alexandrov2020gluonts}
Alexander Alexandrov, Konstantinos Benidis, Michael Bohlke-Schneider, Valentin
  Flunkert, Jan Gasthaus, Tim Januschowski, Danielle~C Maddix, Syama~Sundar
  Rangapuram, David Salinas, Jasper Schulz, et~al.
\newblock Gluon{TS}: Probabilistic and neural time series modeling in python.
\newblock \emph{Journal of Machine Learning Research}, 21\penalty0
  (116):\penalty0 1--6, 2020.

\bibitem[Andrieu et~al.(2003)Andrieu, De~Freitas, Doucet, and
  Jordan]{andrieu2003introduction}
Christophe Andrieu, Nando De~Freitas, Arnaud Doucet, and Michael~I Jordan.
\newblock An introduction to {MCMC} for machine learning.
\newblock \emph{Machine Learning}, 50\penalty0 (1):\penalty0 5--43, 2003.

\bibitem[Angelopoulos and Bates(2021)]{angelopoulos2021gentle}
Anastasios~N Angelopoulos and Stephen Bates.
\newblock A gentle introduction to conformal prediction and distribution-free
  uncertainty quantification.
\newblock \emph{arXiv preprint arXiv:2107.07511}, 2021.

\bibitem[Avati et~al.(2018)Avati, Jung, Harman, Downing, Ng, and
  Shah]{avati2018improving}
Anand Avati, Kenneth Jung, Stephanie Harman, Lance Downing, Andrew Ng, and
  Nigam~H Shah.
\newblock Improving palliative care with deep learning.
\newblock \emph{BMC Medical Informatics and Decision Making}, 18\penalty0
  (4):\penalty0 55--64, 2018.

\bibitem[Bertin-Mahieux et~al.(2011)Bertin-Mahieux, Ellis, Whitman, and
  Lamere]{Bertin-Mahieux2011}
Thierry Bertin-Mahieux, Daniel~P.W. Ellis, Brian Whitman, and Paul Lamere.
\newblock The million song dataset.
\newblock In \emph{{Proceedings of the 12th International Conference on Music
  Information Retrieval ({ISMIR} 2011)}}, 2011.

\bibitem[Bloniarz et~al.(2016)Bloniarz, Talwalkar, Yu, and
  Wu]{bloniarz2016supervised}
Adam Bloniarz, Ameet Talwalkar, Bin Yu, and Christopher Wu.
\newblock Supervised neighborhoods for distributed nonparametric regression.
\newblock In \emph{Proceedings of the 18th International Conference on
  Artificial Intelligence and Statistics}, pages 1450--1459, 2016.

\bibitem[B{\"o}se et~al.(2017)B{\"o}se, Flunkert, Gasthaus, Januschowski,
  Lange, Salinas, Schelter, Seeger, and Wang]{bose2017probabilistic}
Joos-Hendrik B{\"o}se, Valentin Flunkert, Jan Gasthaus, Tim Januschowski,
  Dustin Lange, David Salinas, Sebastian Schelter, Matthias Seeger, and Yuyang
  Wang.
\newblock Probabilistic demand forecasting at scale.
\newblock In \emph{Proceedings of the VLDB Endowment}, volume~10, pages
  1694--1705. VLDB Endowment, 2017.

\bibitem[Breiman(1996)]{breiman1996bagging}
Leo Breiman.
\newblock Bagging predictors.
\newblock \emph{Machine Learning}, 24\penalty0 (2):\penalty0 123--140, 1996.

\bibitem[Brophy and Lowd(2020)]{brophy2020trex}
Jonathan Brophy and Daniel Lowd.
\newblock Trex: Tree-ensemble representer-point explanations.
\newblock In \emph{ICML Workshop on Extending Explainable {AI}}, 2020.

\bibitem[Chen and Guestrin(2016)]{chen2016xgboost}
Tianqi Chen and Carlos Guestrin.
\newblock {XGBoost}: A scalable tree boosting system.
\newblock In \emph{Proceedings of the 22nd ACM SIGKDD International Conference
  on Knowledge Discovery and Data Mining}, 2016.

\bibitem[Chipman et~al.(2010)Chipman, George, and McCulloch]{chipman2010bart}
Hugh~A Chipman, Edward~I George, and Robert~E McCulloch.
\newblock {BART}: Bayesian additive regression trees.
\newblock \emph{The Annals of Applied Statistics}, 4\penalty0 (1):\penalty0
  266--298, 2010.

\bibitem[Chung et~al.(2021)Chung, Char, Guo, Schneider, and
  Neiswanger]{chung2021uncertainty}
Youngseog Chung, Ian Char, Han Guo, Jeff Schneider, and Willie Neiswanger.
\newblock Uncertainty toolbox: An open-source library for assessing,
  visualizing, and improving uncertainty quantification.
\newblock \emph{arXiv preprint arXiv:2109.10254}, 2021.

\bibitem[Cohen et~al.(2009)Cohen, Cohen, and Banthin]{cohen2009medical}
Joel~W Cohen, Steven~B Cohen, and Jessica~S Banthin.
\newblock The medical expenditure panel survey: A national information resource
  to support healthcare cost research and inform policy and practice.
\newblock \emph{Medical Care}, pages S44--S50, 2009.

\bibitem[Coraddu et~al.(2016)Coraddu, Oneto, Ghio, Savio, Anguita, and
  Figari]{coraddu2016machine}
Andrea Coraddu, Luca Oneto, Aessandro Ghio, Stefano Savio, Davide Anguita, and
  Massimo Figari.
\newblock Machine learning approaches for improving condition-based maintenance
  of naval propulsion plants.
\newblock \emph{Journal of Engineering for the Maritime Environment, Part M},
  230\penalty0 (1):\penalty0 136--153, 2016.

\bibitem[Cortez et~al.(2009)Cortez, Cerdeira, Almeida, Matos, and
  Reis]{cortez2009modeling}
Paulo Cortez, Ant{\'o}nio Cerdeira, Fernando Almeida, Telmo Matos, and Jos{\'e}
  Reis.
\newblock Modeling wine preferences by data mining from physicochemical
  properties.
\newblock \emph{Decision Support Systems}, 47\penalty0 (4):\penalty0 547--553,
  2009.

\bibitem[Davies and Ghahramani(2014)]{davies2014random}
Alex Davies and Zoubin Ghahramani.
\newblock The random forest kernel and other kernels for big data from random
  partitions.
\newblock \emph{arXiv preprint arXiv:1402.4293}, 2014.

\bibitem[De~Cock(2011)]{de2011ames}
Dean De~Cock.
\newblock Ames, {I}owa: Alternative to the {B}oston housing data as an end of
  semester regression project.
\newblock \emph{Journal of Statistics Education}, 19\penalty0 (3), 2011.

\bibitem[D'souza et~al.(2021)D'souza, Nussbaum, Agarwal, and Hooker]{d2021tale}
Daniel D'souza, Zach Nussbaum, Chirag Agarwal, and Sara Hooker.
\newblock A tale of two long tails.
\newblock \emph{arXiv preprint arXiv:2107.13098}, 2021.

\bibitem[Dua and Graff(2019)]{Dua:2019}
Dheeru Dua and Casey Graff.
\newblock {UCI} machine learning repository.
\newblock \url{http://archive.ics.uci.edu/ml}, 2019.
\newblock [Online; accessed 12-September-2021].

\bibitem[Duan et~al.(2020)Duan, Anand, Ding, Thai, Basu, Ng, and
  Schuler]{duan2020ngboost}
Tony Duan, Avati Anand, Daisy~Yi Ding, Khanh~K Thai, Sanjay Basu, Andrew Ng,
  and Alejandro Schuler.
\newblock Ngboost: Natural gradient boosting for probabilistic prediction.
\newblock In \emph{Proceedings of the 37th International Conference on Machine
  Learning}, pages 2690--2700. PMLR, 2020.

\bibitem[Fanaee-T and Gama(2013)]{bike_sharing}
Hadi Fanaee-T and Joao Gama.
\newblock Event labeling combining ensemble detectors and background knowledge.
\newblock \emph{Progress in Artificial Intelligence}, pages 1--15, 2013.

\bibitem[Fernandes et~al.(2015)Fernandes, Vinagre, and
  Cortez]{fernandes2015proactive}
Kelwin Fernandes, Pedro Vinagre, and Paulo Cortez.
\newblock A proactive intelligent decision support system for predicting the
  popularity of online news.
\newblock In \emph{Proceedings of the 17th Portuguese Conference on Artificial
  Intelligence}, pages 535--546. Springer, 2015.

\bibitem[Ferreira and Monteiro(2020)]{ferreira2020people}
Juliana~J Ferreira and Mateus~S Monteiro.
\newblock What are people doing about {XAI} user experience? {A} survey on {AI}
  explainability research and practice.
\newblock In \emph{Proceedings of the 22nd International Conference on
  Human-Computer Interaction}, pages 56--73. Springer, 2020.

\bibitem[Friedman et~al.(2000)Friedman, Hastie, and
  Tibshirani]{friedman2000additive}
Jerome Friedman, Trevor Hastie, and Robert Tibshirani.
\newblock Additive logistic regression: A statistical view of boosting.
\newblock \emph{The Annals of Statistics}, 28\penalty0 (2):\penalty0 337--407,
  2000.

\bibitem[Friedman(1991)]{friedman1991multivariate}
Jerome~H Friedman.
\newblock Multivariate adaptive regression splines.
\newblock \emph{The Annals of Statistics}, pages 1--67, 1991.

\bibitem[Gneiting and Katzfuss(2014)]{gneiting2014probabilistic}
Tilmann Gneiting and Matthias Katzfuss.
\newblock Probabilistic forecasting.
\newblock \emph{Annual Review of Statistics and Its Application}, 1:\penalty0
  125--151, 2014.

\bibitem[Gneiting and Raftery(2007)]{gneiting2007strictly}
Tilmann Gneiting and Adrian~E Raftery.
\newblock Strictly proper scoring rules, prediction, and estimation.
\newblock \emph{Journal of the American Statistical Association}, 102\penalty0
  (477):\penalty0 359--378, 2007.

\bibitem[Graves(2011)]{graves2011practical}
Alex Graves.
\newblock Practical variational inference for neural networks.
\newblock In \emph{Proceedings of the 25th International Conference on Neural
  Information Processing Systems}, volume~24, 2011.

\bibitem[Hamidieh(2018)]{hamidieh2018data}
Kam Hamidieh.
\newblock A data-driven statistical model for predicting the critical
  temperature of a superconductor.
\newblock \emph{Computational Materials Science}, 154:\penalty0 346--354, 2018.

\bibitem[Hasson et~al.(2021)Hasson, Wang, Januschowski, and
  Gasthaus]{hasson2021probabilistic}
Hilaf Hasson, Bernie Wang, Tim Januschowski, and Jan Gasthaus.
\newblock Probabilistic forecasting: A level-set approach.
\newblock In \emph{Proceedings of the 35th International Conference on Neural
  Information Processing Systems}, volume~34, 2021.

\bibitem[H{\"u}llermeier and Waegeman(2021)]{hullermeier2021aleatoric}
Eyke H{\"u}llermeier and Willem Waegeman.
\newblock Aleatoric and epistemic uncertainty in machine learning: An
  introduction to concepts and methods.
\newblock \emph{Machine Learning}, 110\penalty0 (3):\penalty0 457--506, 2021.

\bibitem[Januschowski et~al.(2021)Januschowski, Wang, Torkkola, Erkkil{\"a},
  Hasson, and Gasthaus]{januschowski2021forecasting}
Tim Januschowski, Yuyang Wang, Kari Torkkola, Timo Erkkil{\"a}, Hilaf Hasson,
  and Jan Gasthaus.
\newblock Forecasting with trees.
\newblock \emph{International Journal of Forecasting}, 2021.

\bibitem[Kaya et~al.(2012)Kaya, T{\"u}fekci, and G{\"u}rgen]{kaya2012local}
Heysem Kaya, Pmar T{\"u}fekci, and Fikret~S G{\"u}rgen.
\newblock Local and global learning methods for predicting power of a combined
  gas \& steam turbine.
\newblock In \emph{Proceedings of the 2nd International Conference on Emerging
  Trends in Computer and Electronics Engineering ({ICETCEE})}, pages 13--18,
  2012.

\bibitem[Ke et~al.(2017)Ke, Meng, et~al.]{ke2017lightgbm}
Guolin Ke, Qi~Meng, et~al.
\newblock {LightGBM}: A highly efficient gradient boosting decision tree.
\newblock In \emph{Proceedings of the 31st International Conference on Neural
  Information Processing Systems}, 2017.

\bibitem[Koenker and Hallock(2001)]{koenker2001quantile}
Roger Koenker and Kevin~F Hallock.
\newblock Quantile regression.
\newblock \emph{Journal of Economic Perspectives}, 15\penalty0 (4):\penalty0
  143--156, 2001.

\bibitem[Li et~al.(2019)Li, Jin, Xuan, Zhou, Chen, Wang, and
  Yan]{li2019enhancing}
Shiyang Li, Xiaoyong Jin, Yao Xuan, Xiyou Zhou, Wenhu Chen, Yu-Xiang Wang, and
  Xifeng Yan.
\newblock Enhancing the locality and breaking the memory bottleneck of
  transformer on time series forecasting.
\newblock In \emph{Proceedings of the 33rd International Conference on Neural
  Information Processing Systems}, volume~32, pages 5243--5253, 2019.

\bibitem[Lim et~al.(2021)Lim, Ar{\i}k, Loeff, and Pfister]{lim2021temporal}
Bryan Lim, Sercan~{\"O} Ar{\i}k, Nicolas Loeff, and Tomas Pfister.
\newblock Temporal fusion transformers for interpretable multi-horizon time
  series forecasting.
\newblock \emph{International Journal of Forecasting}, 2021.

\bibitem[Lin and Jeon(2006)]{lin2006random}
Yi~Lin and Yongho Jeon.
\newblock Random forests and adaptive nearest neighbors.
\newblock \emph{Journal of the American Statistical Association}, 101\penalty0
  (474):\penalty0 578--590, 2006.

\bibitem[Luo et~al.(2021)Luo, Sang, and Mallick]{luo2021bast}
Zhao~Tang Luo, Huiyan Sang, and Bani Mallick.
\newblock {BAST}: Bayesian additive regression spanning trees for complex
  constrained domain.
\newblock In \emph{Proceedings of the 35th International Conference on Neural
  Information Processing Systems}, volume~34, 2021.

\bibitem[Makridakis et~al.(2021)Makridakis, Spiliotis, Assimakopoulos, Chen,
  Gaba, Tsetlin, and Winkler]{makridakis2021m5a}
Spyros Makridakis, Evangelos Spiliotis, Vassilios Assimakopoulos, Zhi Chen,
  Anil Gaba, Ilia Tsetlin, and Robert~L Winkler.
\newblock The {M}5 uncertainty competition: Results, findings and conclusions.
\newblock \emph{International Journal of Forecasting}, 2021.

\bibitem[Makridakis et~al.(2022)Makridakis, Spiliotis, and
  Assimakopoulos]{makridakis2022m5}
Spyros Makridakis, Evangelos Spiliotis, and Vassilios Assimakopoulos.
\newblock M5 accuracy competition: Results, findings, and conclusions.
\newblock \emph{International Journal of Forecasting}, 2022.

\bibitem[Malinin et~al.(2021)Malinin, Prokhorenkova, and
  Ustimenko]{malinin2021uncertainty}
Andrey Malinin, Liudmila Prokhorenkova, and Aleksei Ustimenko.
\newblock Uncertainty in gradient boosting via ensembles.
\newblock In \emph{Proceedings of the 9th International Conference on Learning
  Representations}, 2021.

\bibitem[Meinshausen and Ridgeway(2006)]{meinshausen2006quantile}
Nicolai Meinshausen and Greg Ridgeway.
\newblock Quantile regression forests.
\newblock \emph{Journal of Machine Learning Research}, 7\penalty0 (6), 2006.

\bibitem[Moosmann et~al.(2006)Moosmann, Triggs, and Jurie]{moosmann2006fast}
Frank Moosmann, Bill Triggs, and Frederic Jurie.
\newblock Fast discriminative visual codebooks using randomized clustering
  forests.
\newblock In \emph{Proceedings of the 20th International Conference on Neural
  Information Processing Systems}, pages 985--992, 2006.

\bibitem[Myung(2003)]{myung2003tutorial}
In~Jae Myung.
\newblock Tutorial on maximum likelihood estimation.
\newblock \emph{Journal of Mathematical Psychology}, 47\penalty0 (1):\penalty0
  90--100, 2003.

\bibitem[Neal(1996)]{neal1996bayesian}
Radford~M Neal.
\newblock \emph{Bayesian Learning for Neural Networks}.
\newblock Springer-Verlag, 1996.

\bibitem[Pace and Barry(1997)]{pace1997sparse}
R~Kelley Pace and Ronald Barry.
\newblock Sparse spatial autoregressions.
\newblock \emph{Statistics \& Probability Letters}, 33\penalty0 (3):\penalty0
  291--297, 1997.

\bibitem[Peterson(2009)]{peterson2009k}
Leif~E Peterson.
\newblock K-nearest neighbor.
\newblock \emph{Scholarpedia}, 4\penalty0 (2):\penalty0 1883, 2009.

\bibitem[Plumb et~al.(2018)Plumb, Molitor, and Talwalkar]{plumb2018model}
Gregory Plumb, Denali Molitor, and Ameet~S Talwalkar.
\newblock Model agnostic supervised local explanations.
\newblock In \emph{Proceedings of the 32nd International Conference on Neural
  Information Processing Systems}, pages 2515--2524, 2018.

\bibitem[Prokhorenkova et~al.(2018)Prokhorenkova, Gusev,
  et~al.]{prokhorenkova2018catboost}
Liudmila Prokhorenkova, Gleb Gusev, et~al.
\newblock {CatBoost}: Unbiased boosting with categorical features.
\newblock In \emph{Proceedings of the 32nd International Conference on Neural
  Information Processing Systems}, 2018.

\bibitem[Rajarshi(2017)]{life}
Kumar Rajarshi.
\newblock Life expectancy ({WHO}).
\newblock
  \url{https://www.kaggle.com/kumarajarshi/life-expectancy-who?ref=hackernoon.com&select=Life+Expectancy+Data.csv},
  2017.
\newblock [Online; accessed 12-September-2021].

\bibitem[Rangapuram et~al.(2018)Rangapuram, Seeger, Gasthaus, Stella, Wang, and
  Januschowski]{rangapuram2018deep}
Syama~Sundar Rangapuram, Matthias~W Seeger, Jan Gasthaus, Lorenzo Stella,
  Yuyang Wang, and Tim Januschowski.
\newblock Deep state space models for time series forecasting.
\newblock In \emph{Proceedings of the 32nd International Conference on Neural
  Information Processing Systems}, volume~31, pages 7785--7794, 2018.

\bibitem[Redmond and Baveja(2002)]{redmond2002data}
Michael Redmond and Alok Baveja.
\newblock A data-driven software tool for enabling cooperative information
  sharing among police departments.
\newblock \emph{European Journal of Operational Research}, 141\penalty0
  (3):\penalty0 660--678, 2002.

\bibitem[Rigby and Stasinopoulos(2005)]{rigby2005generalized}
Robert~A Rigby and D~Mikis Stasinopoulos.
\newblock Generalized additive models for location, scale and shape.
\newblock \emph{Journal of the Royal Statistical Society: Series C (Applied
  Statistics)}, 54\penalty0 (3):\penalty0 507--554, 2005.

\bibitem[Romano et~al.(2019)Romano, Patterson, and
  Candes]{romano2019conformalized}
Yaniv Romano, Evan Patterson, and Emmanuel Candes.
\newblock Conformalized quantile regression.
\newblock In \emph{Proceedings of the 33rd International Conference on Neural
  Information Processing Systems}, volume~32, pages 3543--3553, 2019.

\bibitem[Salinas et~al.(2020)Salinas, Flunkert, Gasthaus, and
  Januschowski]{salinas2020deepar}
David Salinas, Valentin Flunkert, Jan Gasthaus, and Tim Januschowski.
\newblock Deepar: Probabilistic forecasting with autoregressive recurrent
  networks.
\newblock \emph{International Journal of Forecasting}, 36\penalty0
  (3):\penalty0 1181--1191, 2020.

\bibitem[Schlosser et~al.(2019)Schlosser, Hothorn, Stauffer, and
  Zeileis]{schlosser2019distributional}
Lisa Schlosser, Torsten Hothorn, Reto Stauffer, and Achim Zeileis.
\newblock Distributional regression forests for probabilistic precipitation
  forecasting in complex terrain.
\newblock \emph{The Annals of Applied Statistics}, 13\penalty0 (3):\penalty0
  1564--1589, 2019.

\bibitem[Sen et~al.(2019)Sen, Yu, and Dhillon]{sen2019think}
Rajat Sen, Hsiang-Fu Yu, and Inderjit Dhillon.
\newblock Think globally, act locally: A deep neural network approach to
  high-dimensional time series forecasting.
\newblock In \emph{Proceedings of the 33rd International Conference on Neural
  Information Processing Systems}, pages 4837--4846, 2019.

\bibitem[Shafer and Vovk(2008)]{shafer2008tutorial}
Glenn Shafer and Vladimir Vovk.
\newblock A tutorial on conformal prediction.
\newblock \emph{Journal of Machine Learning Research}, 9\penalty0 (3), 2008.

\bibitem[Sheather and Jones(1991)]{sheather1991reliable}
Simon~J Sheather and Michael~C Jones.
\newblock A reliable data-based bandwidth selection method for kernel density
  estimation.
\newblock \emph{Journal of the Royal Statistical Society: Series B
  (Methodological)}, 53\penalty0 (3):\penalty0 683--690, 1991.

\bibitem[Singh et~al.(2015)Singh, Sandhu, and Kumar]{Sing1503:Comment}
Kamaljot Singh, Ranjeet~Kaur Sandhu, and Dinesh Kumar.
\newblock Comment volume prediction using neural networks and decision trees.
\newblock In \emph{IEEE UKSim-AMSS 17th International Conference on
  Computational Modeling and Simulation, UKSim2015}, March 2015.

\bibitem[Sprangers et~al.(2021)Sprangers, Schelter, and
  de~Rijke]{sprangers2021probabilistic}
Olivier Sprangers, Sebastian Schelter, and Maarten de~Rijke.
\newblock Probabilistic gradient boosting machines for large-scale
  probabilistic regression.
\newblock In \emph{Proceedings of the 27th ACM SIGKDD Conference on Knowledge
  Discovery \& Data Mining}, 2021.

\bibitem[Stock et~al.(2012)Stock, Watson, et~al.]{stock2012introduction}
James~H Stock, Mark~W Watson, et~al.
\newblock \emph{Introduction to Econometrics}, volume~3.
\newblock Pearson New York, 2012.

\bibitem[Suzanne(2018)]{obesity}
Suzanne.
\newblock {CDC} data: Nutrition, physical activity, \& obesity.
\newblock
  \url{https://www.kaggle.com/spittman1248/cdc-data-nutrition-physical-activity-obesity},
  2018.
\newblock [Online; accessed 12-September-2021].

\bibitem[Taieb et~al.(2015)Taieb, Huser, Hyndman, Genton,
  et~al.]{taieb2015probabilistic}
Souhaib~Ben Taieb, Raphael Huser, Rob Hyndman, Marc Genton, et~al.
\newblock Probabilistic time series forecasting with boosted additive models:
  An application to smart meter data.
\newblock \emph{Department of Economics and Business Statistics, Monash
  University}, 2015.

\bibitem[Taylor and Letham(2018)]{taylor2018forecasting}
Sean~J Taylor and Benjamin Letham.
\newblock Forecasting at scale.
\newblock \emph{The American Statistician}, 72\penalty0 (1):\penalty0 37--45,
  2018.

\bibitem[Tjoa and Guan(2020)]{tjoa2020survey}
Erico Tjoa and Cuntai Guan.
\newblock A survey on explainable artificial intelligence ({XAI}): Toward
  medical {XAI}.
\newblock \emph{IEEE Transactions on Neural Networks and Learning Systems},
  2020.

\bibitem[Tsanas and Xifara(2012)]{tsanas2012accurate}
Athanasios Tsanas and Angeliki Xifara.
\newblock Accurate quantitative estimation of energy performance of residential
  buildings using statistical machine learning tools.
\newblock \emph{Energy and Buildings}, 49:\penalty0 560--567, 2012.

\bibitem[T{\"u}fekci(2014)]{tufekci2014prediction}
P{\i}nar T{\"u}fekci.
\newblock Prediction of full load electrical power output of a base load
  operated combined cycle power plant using machine learning methods.
\newblock \emph{International Journal of Electrical Power \& Energy Systems},
  60:\penalty0 126--140, 2014.

\bibitem[Ustimenko and Prokhorenkova(2021)]{ustimenko2021sglb}
Aleksei Ustimenko and Liudmila Prokhorenkova.
\newblock {SGLB}: Stochastic gradient langevin boosting.
\newblock In \emph{International Conference on Machine Learning}, pages
  10487--10496. PMLR, 2021.

\bibitem[van Rijn(2014)]{kin8nm}
Jan van Rijn.
\newblock Kin8nm.
\newblock \url{https://www.openml.org/d/189}, 2014.
\newblock [Online; accessed 20-January-2022].

\bibitem[Wang et~al.(2019)Wang, Smola, Maddix, Gasthaus, Foster, and
  Januschowski]{wang2019deep}
Yuyang Wang, Alex Smola, Danielle Maddix, Jan Gasthaus, Dean Foster, and Tim
  Januschowski.
\newblock Deep factors for forecasting.
\newblock In \emph{Proceedings of the 36th International Conference on Machine
  Learning}, pages 6607--6617. PMLR, 2019.

\bibitem[Yeh(1998)]{yeh1998modeling}
I-C Yeh.
\newblock Modeling of strength of high-performance concrete using artificial
  neural networks.
\newblock \emph{Cement and Concrete Research}, 28\penalty0 (12):\penalty0
  1797--1808, 1998.

\bibitem[Zamo and Naveau(2018)]{zamo2018estimation}
Micha{\"e}l Zamo and Philippe Naveau.
\newblock Estimation of the continuous ranked probability score with limited
  information and applications to ensemble weather forecasts.
\newblock \emph{Mathematical Geosciences}, 50\penalty0 (2):\penalty0 209--234,
  2018.

\end{thebibliography}

\section*{Checklist}

\begin{enumerate}

\item For all authors...
\begin{enumerate}
  \item Do the main claims made in the abstract and introduction accurately reflect the paper's contributions and scope?
    \answerYes{}
  \item Did you describe the limitations of your work?
    \answerYes{See \S\ref{sec:conclusion}.}
  \item Did you discuss any potential negative societal impacts of your work?
    \answerYes{See \S\ref{app_sec:ethical_statement}.}
  \item Have you read the ethics review guidelines and ensured that your paper conforms to them?
    \answerYes{}
\end{enumerate}

\item If you are including theoretical results...
\begin{enumerate}
  \item Did you state the full set of assumptions of all theoretical results?
    \answerNA{}
	\item Did you include complete proofs of all theoretical results?
    \answerNA{}
\end{enumerate}

\item If you ran experiments...
\begin{enumerate}
  \item Did you include the code, data, and instructions needed to reproduce the main experimental results (either in the supplemental material or as a URL)?
    \answerYes{See \S\ref{sec:experiments}}.
  \item Did you specify all the training details (e.g., data splits, hyperparameters, how they were chosen)?
    \answerYes{See \S\ref{sec:methodology}.}
	\item Did you report error bars (e.g., with respect to the random seed after running experiments multiple times)?
    \answerYes{}
	\item Did you include the total amount of compute and the type of resources used (e.g., type of GPUs, internal cluster, or cloud provider)?
    \answerYes{See \S\ref{sec:experiments}.}
\end{enumerate}

\item If you are using existing assets (e.g., code, data, models) or curating/releasing new assets...
\begin{enumerate}
  \item If your work uses existing assets, did you cite the creators?
    \answerYes{See \S\ref{app_sec:datasets}.}
  \item Did you mention the license of the assets?
    \answerNo{The license of each asset can be found by following its corresponding citation.}
  \item Did you include any new assets either in the supplemental material or as a URL?
    \answerYes{See \S\ref{sec:experiments}.}
  \item Did you discuss whether and how consent was obtained from people whose data you're using/curating?
    \answerNA{}
  \item Did you discuss whether the data you are using/curating contains personally identifiable information or offensive content?
    \answerNA{}
\end{enumerate}

\item If you used crowdsourcing or conducted research with human subjects...
\begin{enumerate}
  \item Did you include the full text of instructions given to participants and screenshots, if applicable?
    \answerNA{}
  \item Did you describe any potential participant risks, with links to Institutional Review Board (IRB) approvals, if applicable?
    \answerNA{}
  \item Did you include the estimated hourly wage paid to participants and the total amount spent on participant compensation?
    \answerNA{}
\end{enumerate}

\end{enumerate}

%%%%%%%%%%%%%%%%%%%%%%%%%%%%%%%%%%%%%%%%%%%%%%%%%%%%%%%%%%%%

\appendix

\newpage

\section{Algorithmic Details}
\label{app_sec:alg_details}

\begin{figure}[h]
\centering
\includegraphics[width=0.85\textwidth]{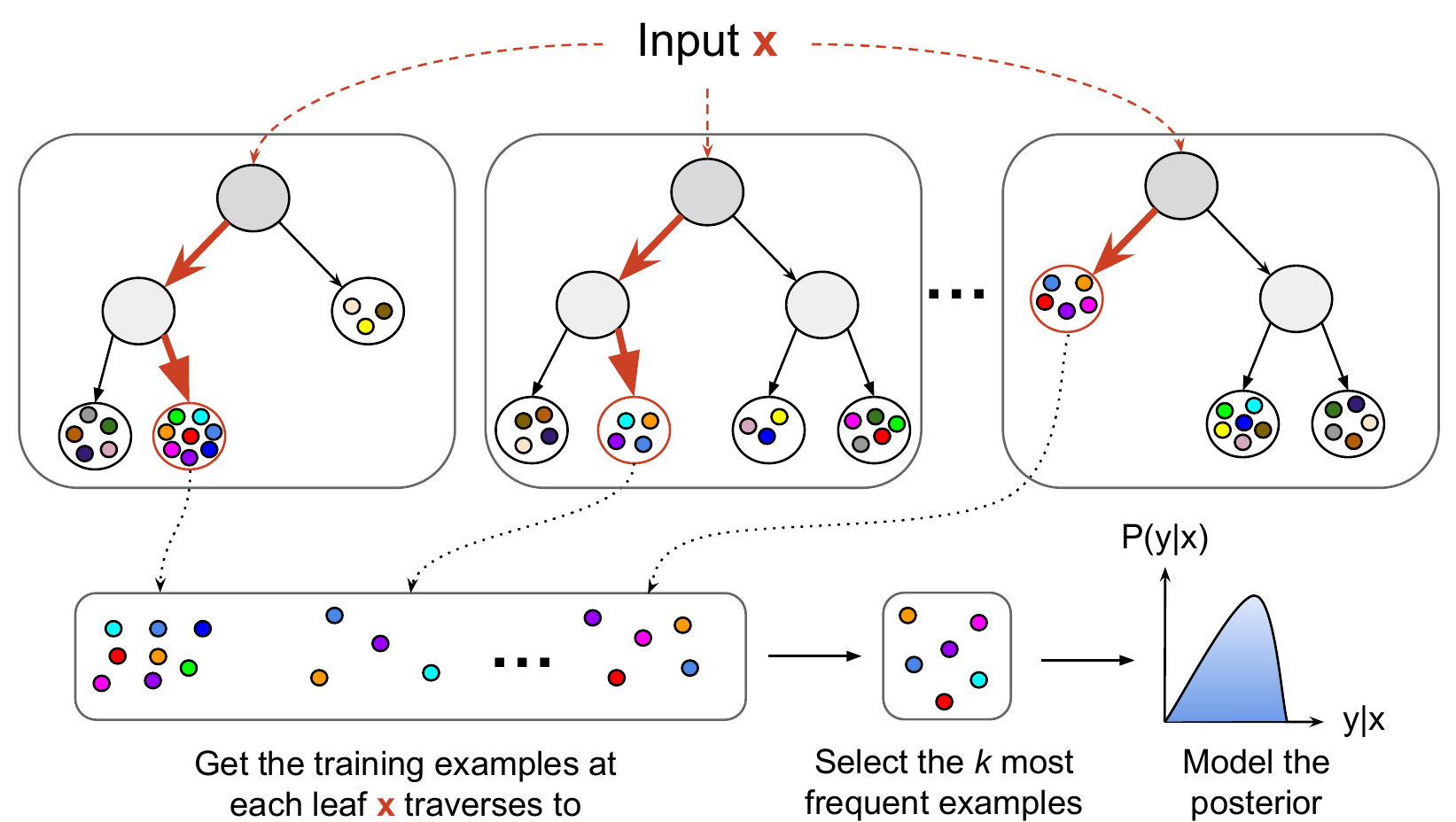}
\caption{IBUG workflow. Given a GBRT model and an input instance~$x$, IBUG collects the training examples at each leaf~$x$ traverses to, keeps the~$k$ most frequent examples, and then uses those examples to model the output distribution.}
\label{app_fig:ibug}
\end{figure}

Figure~\ref{app_fig:ibug} summarizes how IBUG generates a probabilistic prediction for a given input instance.

\subsection{Ethical Statement}
\label{app_sec:ethical_statement}

In general, this work has no foreseeable negative societal impacts; however, users should carefully validate their models as imprecise uncertainty estimates may adversely affect certain domains~(e.g., healthcare, weather).

\newpage

\section{Implementation and Experiment Details}
\label{app_sec:reproducibility}

We implement IBUG in Python, using Cython---a Python package allowing the development of C extensions---to store a unified representation of the model structure. IBUG currently supports all major modern gradient boosting frameworks including XGBoost~\citep{chen2016xgboost}, LightGBM~\citep{ke2017lightgbm}, and CatBoost~\citep{prokhorenkova2018catboost}. Experiments are run on an Intel(R) Xeon(R) CPU E5-2690 v4 @ 2.6GHz with 60GB of RAM @ 2.4GHz. We run our experiments on publicly available datasets. Links to all data sources as well as the code for IBUG and all experiments is currently available online at~\url{https://github.com/jjbrophy47/ibug}.

\paragraph{Metrics.} We use the continuous ranked probability score~(CRPS) and negative log likelihood~(NLL) to measure probabilistic performance. CRPS is a quadratic measure of discrepancy between the cumulative distribution function~(CDF)~$F$ of forecast~$\hat{y}$ and the empirical CDF of the scalar observation~$y$: {\small $\int (F(\hat{y}) - \mathbbm{1}[\hat{y} \geq y])^2 d\hat{y}$} in which~$\mathbbm{1}$ is the indicator function~\cite{gneiting2007strictly,zamo2018estimation}. To evaluate point performance, we use root mean squared error~(RMSE): {\small $\sqrt{\frac{1}{n} \sum_{i=1}^{n}(y_i - \hat{y}_i)^2}$}.

\subsection{Datasets}
\label{app_sec:datasets}

This section gives a detailed description for each dataset we use in our experiments.

\begin{itemize}

    \item \textbf{Ames}~\citep{de2011ames} consists of 2,930 instances of housing prices in the Ames, Iowa area characterized by 80 attributes. The aim is to predict the sale price of a given house.
    
    \item \textbf{Bike}~\citep{bike_sharing,Dua:2019} contains 17,379 measurements of the number of bikes rented per hour characterized by 16 attributes. The aim is to predict the number of bikes rented for a given hour.
    
    \item \textbf{California}~\citep{pace1997sparse} consists of 20,640 instances of median housing prices in various California districts characterized by 8 attributes. The aim is to predict the median housing price for the given district.
    
    \item \textbf{Communities}~\citep{redmond2002data,Dua:2019} consists of 1,994 measurements of violent crime statistics based on crime, survey, and census data. The dataset is characterized by 100 attributes, and the aim is to predict the violent crime rate for a given population.

    \item \textbf{Concrete}~\citep{yeh1998modeling,Dua:2019} consists of 1,030 instances of concrete characterized by 8 attributes. The aim is to predict the compressive strength of the concrete.

    \item \textbf{Energy}~\citep{tsanas2012accurate,Dua:2019} consists of 768 buildings in which each building is one of 12 different shapes and is characterized by 8 features. The aim is to predict the cooling load associated with the building.
    
    \item \textbf{Facebook}~\citep{Sing1503:Comment,Dua:2019} consists of 40,949 Facebook posts characterized by 53 attributes. The aim is to predict the number of comments for a given post.
    
    \item \textbf{Kin8nm}~\citep{kin8nm} consists of 8,192 instances of the forward kinematics of an 8 link robotic arm. The aim is to predict the forward kinematics of the robotic arm.
    
    \item \textbf{Life}~\citep{life} consists of 2,928 instances of life expectancy estimates for various countries during one year. Each instance is characterized by 20 attributes, and the aim is to predict the life expectancy of the country during a specific year.
    
    \item \textbf{MEPS}~\citep{cohen2009medical} consists of 16,656 instances of medical expenditure survey data. Each instance is characterized by 139 attributes, and the aim is to predict the insurance utilization for the given medical expenditure.
    
    \item \textbf{MSD}~\citep{Bertin-Mahieux2011} consists of 515,345 songs characterized by 90 audio features constructed from each song. The aim is to predict what year the song was released based on the audio features.
    
    \item \textbf{Naval}~\citep{coraddu2016machine,Dua:2019} consists of 11,934 instances extracted from a high-performing gas turbine simulation. Each instance is characterized by 16 features. The aim is to predict the gas turbine decay coefficient.
    
    \item \textbf{News}~\citep{Dua:2019,fernandes2015proactive} consists of 39,644 Mashable articles characterized by 60 features. The aim is to predict the number of shares for a given article.
    
    \item \textbf{Obesity}~\citep{obesity} contains 48,346 instances of obesity rates for different states and regions with differing socioeconomic backgrounds. Each instance is characterized by 32 attributes. The aim is to predict the obesity rate of the region.
    
    \item \textbf{Power}~\citep{Dua:2019,kaya2012local,tufekci2014prediction} contains 9,568 readings of a Combined Cycle Power Plant~(CCPP) at full work load. Each reading is characterized by 4 features. The aim is to predict the net hourly electrical energy output.
    
    \item \textbf{Protein}~\citep{Dua:2019} contains 45,730 tertiary-protein-structure instances characterized by 9 attributes. The aim is to predict the armstrong coefficient of the protein structure.
    
    \item \textbf{STAR}~\citep{Dua:2019,stock2012introduction} contains 2,161 student-teacher achievement scores characterized by 39 attributes. The aim is to predict the student-teacher achievement based on the given intervention.
    
    \item \textbf{Superconductor}~\citep{Dua:2019,hamidieh2018data} contains 21,263 potential superconductors characterized by 81 attributes. The aim is to predict the critical temperature of the given superconductor.
    
    \item \textbf{Synthetic}~\citep{breiman1996bagging,friedman1991multivariate} is a non-linear synthetic regression dataset in which the inputs are independent and uniformly distributed on the interval~$[0, 1]$; the dataset contains 10,000 instances characterized by 100 attributes.
    
    \item \textbf{Wave}~\citep{Dua:2019} consists of 287,999 positions and absorbed power outputs of wave energy converters (WECs) in four real wave scenarios off the southern coast of Australia (Sydney, Adelaide, Perth and Tasmania). The aim is to predict the total power output of a given WEC.
    
    \item \textbf{Wine}~\citep{cortez2009modeling,Dua:2019} consists of 6,497 instances of Portuguese ``Vinho Verde'' red and white wine characterized by 11 features. The aim is to predict the quality of the wine from 0-10.
    
    \item \textbf{Yacht}~\citep{Dua:2019} consists of 308 instances of yacht-sailing performance characterized by 6 attributes. The aim is to predict the residual resistance per unit weight of displacement.

\end{itemize}

For each dataset, we generate one-hot encodings for any categorical variable and leave all numeric and binary variables as is. Table~\ref{app_tab:dataset_summary} shows a summary of the datasets after preprocessing.

\begin{table}[h]
    \caption{Dataset summary after preprocessing.}
    \label{app_tab:dataset_summary}
    \centering
    \small
    \vspace{0.5em}
    \begin{tabular}{@{}lcrr@{}}
        \toprule
        Dataset & Source & $n$ & $p$ \\
        \midrule
        Ames & \cite{de2011ames} & 2,930 & 358 \\
        Bike & \cite{bike_sharing,Dua:2019} & 17,379 & 37 \\
        California & \cite{pace1997sparse} & 20,640 & 100 \\
        Communities & \cite{Dua:2019,redmond2002data} & 1,994 & 100 \\
        Concrete & \cite{yeh1998modeling,Dua:2019} & 1,030 & 8 \\
        Energy & \cite{tsanas2012accurate,Dua:2019} & 768 & 16 \\
        Facebook & \cite{Sing1503:Comment,Dua:2019} & 40,949 & 133 \\
        Kin8nm & \cite{kin8nm} & 8,192 & 8 \\
        Life & \cite{life} & 2,928 & 204 \\
        MEPS & \cite{cohen2009medical} & 15,656 & 139 \\
        MSD & \cite{Bertin-Mahieux2011} & 515,345 & 90 \\
        Naval & \cite{Dua:2019,coraddu2016machine} & 11,934 & 17 \\
        News & \cite{Dua:2019,fernandes2015proactive} & 39,644 & 58 \\
        Obesity & \cite{obesity} & 48,346 & 100 \\
        Power & \cite{Dua:2019,kaya2012local,tufekci2014prediction} & 9,568 & 4 \\
        Protein & \cite{Dua:2019} & 45,730 & 9 \\
        STAR & \cite{Dua:2019,stock2012introduction} & 2,161 & 95 \\
        Superconductor & \cite{Dua:2019,hamidieh2018data} & 21,263 & 82 \\
        Synthetic & \cite{breiman1996bagging,friedman1991multivariate} & 10,000 & 100 \\
        Wave & \cite{Dua:2019} & 287,999 & 48 \\
        Wine & \cite{cortez2009modeling,Dua:2019} & 6,497 & 11 \\
        Yacht & \cite{Dua:2019} & 308 & 6 \\
        \bottomrule
    \end{tabular}
\end{table}

\newpage

\subsection{Hyperparameters}
\label{app_sec:hyperparameters}

Tables~\ref{tab:hyperparameters_crps} and~\ref{tab:hyperparameters_nll} show hyperparameter values selected most often for each dataset when optimizing CRPS and NLL, respectively. We tune nearest-neighbor hyperparameter~$k$ using values [3, 5, 7, 9, 11, 15, 31, 61, 91, 121, 151, 201, 301, 401, 501, 601, 701], $\gamma$ and $\delta$ using values [1e-8, 1e-7, 1e-6, 1e-5, 1e-4, 1e-3, 1e-2, 1e-1, 0, 1e0, 1e1, 1e2, 1e3] with multipliers [1.0, 2.5, 5.0], number of trees~$T$ using values [10, 25, 50, 100, 250, 500, 1000, 2000]~(since NGBoost has no hyperparameters to tune besides~$T$, we tune~$T$ on the validation set using early stopping~\cite{duan2020ngboost}), learning rate~$\eta$ using values~[0.01, 0.1], maximum number of leaves~$h$ using values~[15, 31, 61, 91], minimum number of leaves~$n_{\ell_0}$ using values [1, 20], maximum depth~$d$ using values~[2, 3, 5, 7, -1 (unlimited)], and~$\rho$ which selects the minimum variance computed from the validation set predictions. For the MSD and Wave datasets, we use a bagging fraction of 0.1~\cite{duan2020ngboost,sprangers2021probabilistic}.

\begin{table}[ht]
\caption{Hyperparameters selected most often over 10 folds for each dataset when optimizing CRPS.}
\label{tab:hyperparameters_crps}
\vspace{0.5em}
% \centering
\scriptsize
\begin{tabular}{lcc|ccccc|ccccc}
\toprule
& \multicolumn{2}{c}{NGBoost} & \multicolumn{5}{c}{PGBM}
& \multicolumn{5}{c}{CBU} \\
\cmidrule(lr){2-3}\cmidrule(lr){4-8}\cmidrule(lr){9-13}
Dataset & $T$ & $\gamma/\delta$ & $T$ & $\eta$ & $h$ & $n_{\ell_0}$ & $\gamma/\delta$ & $T$ & $\eta$ & $d$ & $n_{\ell_0}$ & $\gamma/\delta$ \\
\midrule
Ames & 2000 & $\gamma$:1e+00 & 2000 & 0.1 & 15 & 1 & $\gamma$:2e+00 & 2000 & 0.1 & 7 & 1 & $\delta$:5e+03 \\
Bike & 2000 & $\gamma$:5e-01 & 2000 & 0.01 & 61 & 1 & $\delta$:1e-08 & 2000 & 0.1 & 2 & 1 & $\delta$:5e-02 \\
California & 2000 & $\delta$:3e-02 & 1000 & 0.1 & 31 & 20 & $\delta$:3e-02 & 2000 & 0.1 & -1 & 1 & $\delta$:1e-01 \\
Communities & 223 & $\delta$:1e-02 & 500 & 0.01 & 15 & 20 & $\gamma$:1e+01 & 2000 & 0.01 & 7 & 1 & $\delta$:5e-02 \\
Concrete & 2000 & $\delta$:1e+00 & 2000 & 0.1 & 15 & 20 & $\delta$:1e-08 & 2000 & 0.1 & 5 & 1 & $\delta$:2e+00 \\
Energy & 2000 & $\gamma$:5e-01 & 2000 & 0.1 & 15 & 1 & $\gamma$:5e-01 & 2000 & 0.1 & 3 & 1 & $\delta$:1e-01 \\
Facebook & 2000 & $\gamma$:1e+00 & 2000 & 0.01 & 15 & 1 & $\gamma$:2e+00 & 2000 & 0.1 & 5 & 1 & $\gamma$:1e+00 \\
Kin8nm & 581 & $\delta$:3e-02 & 2000 & 0.1 & 61 & 20 & $\delta$:5e-02 & 2000 & 0.1 & 7 & 1 & $\delta$:5e-02 \\
Life & 2000 & $\gamma$:1e+00 & 2000 & 0.1 & 15 & 1 & $\delta$:2e-01 & 2000 & 0.1 & 5 & 1 & $\delta$:1e+00 \\
MEPS & 583 & $\delta$:1e-01 & 50 & 0.1 & 15 & 1 & $\delta$:1e-08 & 100 & 0.01 & -1 & 1 & $\gamma$:1e+00 \\
MSD & 2000 & $\delta$:1e-01 & 2000 & 0.01 & 91 & 20 & $\gamma$:1e+01 & 2000 & 0.1 & 7 & 1 & $\delta$:5e-01 \\
Naval & 2000 & $\delta$:0e+00 & 2000 & 0.1 & 61 & 20 & $\gamma$:3e-02 & 2000 & 0.1 & 7 & 1 & $\delta$:3e-04 \\
News & 2000 & $\gamma$:5e-01 & 100 & 0.01 & 15 & 20 & $\delta$:2e+03 & 100 & 0.01 & 2 & 1 & $\gamma$:5e-01 \\
Obesity & 2000 & $\delta$:1e-01 & 500 & 0.1 & 91 & 20 & $\delta$:1e-08 & 2000 & 0.1 & 7 & 1 & $\delta$:5e-01 \\
Power & 2000 & $\delta$:2e-01 & 500 & 0.1 & 91 & 1 & $\delta$:5e-01 & 2000 & 0.1 & 7 & 1 & $\delta$:1e+00 \\
Protein & 2000 & $\delta$:1e-01 & 2000 & 0.1 & 91 & 20 & $\gamma$:2e+00 & 2000 & 0.1 & 7 & 1 & $\delta$:1e+00 \\
STAR & 187 & $\delta$:2e+01 & 1000 & 0.01 & 15 & 1 & $\gamma$:1e+01 & 2000 & 0.01 & -1 & 1 & $\delta$:5e+01 \\
Superconductor & 162 & $\gamma$:5e-01 & 1000 & 0.01 & 15 & 20 & $\delta$:5e-02 & 2000 & 0.1 & -1 & 1 & $\delta$:3e-02 \\
Synthetic & 208 & $\delta$:5e-01 & 500 & 0.01 & 15 & 20 & $\delta$:1e+01 & 2000 & 0.01 & 3 & 1 & $\delta$:1e+00 \\
Wave & 2000 & $\delta$:5e+03 & 2000 & 0.1 & 15 & 1 & $\delta$:1e-08 & 2000 & 0.1 & -1 & 1 & $\delta$:2e+02 \\
Wine & 309 & $\delta$:5e-02 & 2000 & 0.01 & 91 & 20 & $\delta$:5e-01 & 2000 & 0.1 & 7 & 1 & $\delta$:2e-01 \\
Yacht & 2000 & $\gamma$:5e-01 & 2000 & 0.1 & 15 & 1 & $\delta$:0e+00 & 2000 & 0.1 & 3 & 1 & $\gamma$:2e+00 \\
% \bottomrule
\midrule
\end{tabular}
\quad
\begin{tabular}{lccccccc}
% \vspace{0.5em}
% \toprule
% \midrule
& \multicolumn{4}{c}{CatBoost} & \multicolumn{3}{c}{IBUG} \\
\cmidrule(lr){2-5}\cmidrule(lr){6-8}
Dataset &
$T$ & $\eta$ & $d$ & $n_{\ell_0}$ & $k$ & $\rho$ & $\gamma/\delta$ \\
\midrule
Ames & 2000 & 0.1 & -1 & 1 & 5 & 2206 & $\delta$:3e-04 \\
Bike & 2000 & 0.1 & 2 & 1 & 3 & 0.471 & $\gamma$:2e-01 \\
California & 2000 & 0.1 & -1 & 1 & 7 & 2e-15 & $\delta$:1e-08 \\
Communities & 2000 & 0.01 & -1 & 1 & 15 & 0.017 & $\delta$:0e+00 \\
Concrete & 2000 & 0.1 & 5 & 1 & 3 & 0.049 & $\gamma$:5e-01 \\
Energy & 2000 & 0.1 & 5 & 1 & 3 & 0.035 & $\gamma$:1e-01 \\
Facebook & 2000 & 0.1 & -1 & 1 & 15 & 0.213 & $\delta$:1e-01 \\
Kin8nm & 2000 & 0.1 & 7 & 1 & 3 & 0.003 & $\gamma$:5e-01 \\
Life & 2000 & 0.1 & 5 & 1 & 3 & 0.047 & $\gamma$:5e-01 \\
MEPS & 250 & 0.01 & 5 & 1 & 201 & 1.08 & $\delta$:1e-07 \\
MSD & 2000 & 0.1 & 7 & 1 & 31 & 1.25 & $\delta$:1e-07 \\
Naval & 2000 & 0.1 & 7 & 1 & 3 & 1e-15 & $\gamma$:5e-01 \\
News & 1000 & 0.01 & 2 & 1 & 15 & 163 & $\gamma$:5e-01 \\
Obesity & 2000 & 0.1 & 7 & 1 & 5 & 0.306 & $\gamma$:5e-01 \\
Power & 2000 & 0.1 & 7 & 1 & 5 & 0.220 & $\delta$:1e-01 \\
Protein & 2000 & 0.1 & 7 & 1 & 31 & 0.028 & $\delta$:1e-01 \\
STAR & 250 & 0.01 & 5 & 1 & 121 & 192 & $\delta$:1e-08 \\
Superconductor & 2000 & 0.1 & 5 & 1 & 3 & 5e-15 & $\gamma$:1e-01 \\
Synthetic & 1000 & 0.01 & 7 & 1 & 401 & 9.34 & $\delta$:1e-08 \\
Wave & 2000 & 0.1 & -1 & 1 & 3 & 2e-10 & $\gamma$:2e-01 \\
Wine & 2000 & 0.1 & 7 & 1 & 15 & 0.268 & $\delta$:2e-08 \\
Yacht & 2000 & 0.1 & 2 & 1 & 3 & 0.196 & $\gamma$:1e-01 \\
\bottomrule
\end{tabular}
\end{table}

\begin{table}[ht]
\caption{Hyperparameters selected most often over 10 folds for each dataset when optimizing NLL.}
\label{tab:hyperparameters_nll}
\vspace{0.5em}
% \centering
\scriptsize
\begin{tabular}{lcc|ccccc|ccccc}
\toprule
& \multicolumn{2}{c}{NGBoost} & \multicolumn{5}{c}{PGBM}
& \multicolumn{5}{c}{CBU} \\
\cmidrule(lr){2-3}\cmidrule(lr){4-8}\cmidrule(lr){9-13}
Dataset & $T$ & $\gamma/\delta$ & $T$ & $\eta$ & $h$ & $n_{\ell_0}$ & $\gamma/\delta$ & $T$ & $\eta$ & $d$ & $n_{\ell_0}$ & $\gamma/\delta$ \\
\midrule
Ames & 373 & $\delta$:2e+03 & 2000 & 0.1 & 15 & 1 & $\gamma$:2e+00 & 2000 & 0.1 & 7 & 1 & $\gamma$:1e+01 \\
Bike & 926 & $\delta$:0e+00 & 2000 & 0.01 & 61 & 1 & $\delta$:1e-08 & 2000 & 0.1 & 2 & 1 & $\delta$:0e+00 \\
California & 2000 & $\delta$:5e-02 & 1000 & 0.1 & 31 & 20 & $\delta$:1e-01 & 2000 & 0.1 & -1 & 1 & $\delta$:2e-01 \\
Communities & 156 & $\delta$:1e-02 & 500 & 0.01 & 15 & 20 & $\gamma$:1e+01 & 2000 & 0.01 & 7 & 1 & $\delta$:1e-01 \\
Concrete & 383 & $\delta$:1e+00 & 2000 & 0.1 & 15 & 20 & $\delta$:1e+00 & 2000 & 0.1 & 5 & 1 & $\delta$:2e+00 \\
Energy & 422 & $\delta$:1e-02 & 2000 & 0.1 & 15 & 1 & $\delta$:1e-08 & 2000 & 0.1 & 3 & 1 & $\delta$:1e-01 \\
Facebook & 549 & $\delta$:0e+00 & 2000 & 0.01 & 15 & 1 & $\gamma$:5e+00 & 2000 & 0.1 & 5 & 1 & $\gamma$:2e+00 \\
Kin8nm & 975 & $\delta$:1e-02 & 2000 & 0.1 & 61 & 20 & $\delta$:5e-02 & 2000 & 0.1 & 7 & 1 & $\delta$:1e-01 \\
Life & 366 & $\delta$:2e-01 & 2000 & 0.1 & 15 & 1 & $\delta$:1e+00 & 2000 & 0.1 & 5 & 1 & $\delta$:1e+00 \\
MEPS & 188 & $\delta$:1e+00 & 50 & 0.1 & 15 & 1 & $\delta$:1e-08 & 100 & 0.01 & -1 & 1 & $\delta$:1e+00 \\
MSD & 2000 & $\delta$:3e-02 & 2000 & 0.01 & 91 & 20 & $\gamma$:1e+01 & 2000 & 0.1 & 7 & 1 & $\delta$:1e+00 \\
Naval & 2000 & $\delta$:5e-05 & 2000 & 0.1 & 61 & 20 & $\gamma$:5e-02 & 2000 & 0.1 & 7 & 1 & $\delta$:3e-04 \\
News & 38 & $\delta$:1e+03 & 100 & 0.01 & 15 & 20 & $\gamma$:1e+01 & 100 & 0.01 & 2 & 1 & $\delta$:2e+03 \\
Obesity & 2000 & $\delta$:0e+00 & 500 & 0.1 & 91 & 20 & $\delta$:1e-08 & 2000 & 0.1 & 7 & 1 & $\delta$:5e-01 \\
Power & 275 & $\delta$:2e-01 & 500 & 0.1 & 91 & 1 & $\delta$:1e+00 & 2000 & 0.1 & 7 & 1 & $\delta$:2e+00 \\
Protein & 2000 & $\delta$:2e-01 & 2000 & 0.1 & 91 & 20 & $\delta$:2e+00 & 2000 & 0.1 & 7 & 1 & $\delta$:1e+00 \\
STAR & 176 & $\delta$:1e+01 & 1000 & 0.01 & 15 & 1 & $\delta$:2e+02 & 2000 & 0.01 & -1 & 1 & $\delta$:5e+01 \\
Superconductor & 378 & $\gamma$:1e+00 & 1000 & 0.01 & 15 & 20 & $\gamma$:2e+00 & 2000 & 0.1 & -1 & 1 & $\delta$:1e-01 \\
Synthetic & 284 & $\delta$:5e-01 & 500 & 0.01 & 15 & 20 & $\delta$:1e+01 & 2000 & 0.01 & 3 & 1 & $\delta$:1e+00 \\
Wave & 2000 & $\gamma$:1e+00 & 2000 & 0.1 & 15 & 1 & $\delta$:5e+02 & 2000 & 0.1 & -1 & 1 & $\delta$:2e+02 \\
Wine & 390 & $\delta$:5e-02 & 2000 & 0.01 & 91 & 20 & $\gamma$:2e+01 & 2000 & 0.1 & 7 & 1 & $\delta$:5e-01 \\
Yacht & 356 & $\delta$:0e+00 & 2000 & 0.1 & 15 & 1 & $\delta$:5e-02 & 2000 & 0.1 & 3 & 1 & $\delta$:5e-01 \\
% \bottomrule
\midrule
\end{tabular}
\quad
\begin{tabular}{lccccccc}
% \vspace{0.5em}
% \toprule
% \midrule
& \multicolumn{4}{c}{CatBoost} & \multicolumn{3}{c}{IBUG} \\
\cmidrule(lr){2-5}\cmidrule(lr){6-8}
Dataset &
$T$ & $\eta$ & $d$ & $n_{\ell_0}$ & $k$ & $\rho$ & $\gamma/\delta$ \\
\midrule
Ames & 2000 & 0.1 & -1 & 1 & 11 & 4673 & $\delta$:1e-08 \\
Bike & 2000 & 0.1 & 2 & 1 & 5 & 0.4 & $\gamma$:2e-01 \\
California & 2000 & 0.1 & -1 & 1 & 31 & 0.063 & $\delta$:0e+00 \\
Communities & 2000 & 0.01 & -1 & 1 & 61 & 0.026 & $\delta$:0e+00 \\
Concrete & 2000 & 0.1 & 5 & 1 & 5 & 0.56 & $\delta$:1e-08 \\
Energy & 2000 & 0.1 & 5 & 1 & 3 & 0.087 & $\gamma$:2e-01 \\
Facebook & 2000 & 0.1 & -1 & 1 & 301 & 0.175 & $\delta$:1e-01 \\
Kin8nm & 2000 & 0.1 & 7 & 1 & 7 & 0.031 & $\delta$:0e+00 \\
Life & 2000 & 0.1 & 5 & 1 & 7 & 0.22 & $\delta$:2e-08 \\
MEPS & 250 & 0.01 & 5 & 1 & 301 & 1.76 & $\delta$:1e+00 \\
MSD & 2000 & 0.1 & 7 & 1 & 61 & 1.75 & $\delta$:1e-07 \\
Naval & 2000 & 0.1 & 7 & 1 & 5 & 4e-04 & $\gamma$:5e-01 \\
News & 1000 & 0.01 & 2 & 1 & 301 & 994 & $\delta$:2e+03 \\
Obesity & 2000 & 0.1 & 7 & 1 & 9 & 0.529 & $\delta$:1e-07 \\
Power & 2000 & 0.1 & 7 & 1 & 15 & 0.861 & $\delta$:1e-07 \\
Protein & 2000 & 0.1 & 7 & 1 & 121 & 0.218 & $\delta$:5e-08 \\
STAR & 250 & 0.01 & 5 & 1 & 121 & 189 & $\delta$:1e-05 \\
Superconductor & 2000 & 0.1 & 5 & 1 & 7 & 0.019 & $\gamma$:2e-01 \\
Synthetic & 1000 & 0.01 & 7 & 1 & 401 & 9.39 & $\delta$:1e-08 \\
Wave & 2000 & 0.1 & -1 & 1 & 31 & 349 & $\gamma$:2e-01 \\
Wine & 2000 & 0.1 & 7 & 1 & 61 & 0.297 & $\delta$:2e-08 \\
Yacht & 2000 & 0.1 & 2 & 1 & 3 & 0.196 & $\gamma$:2e-01 \\
\bottomrule
\end{tabular}
\end{table}

\newpage
\hphantom{dummy}
\newpage

\subsection{Additional Metrics}
\label{app_sec:additional_metrics}

In this section, we show results for point performance and probabilistic performance with additional metrics. Each table shows average results over the 10 random folds for each dataset, with standard errors in subscripted parentheses. We use the \emph{Uncertainty Toolbox}\footnote{\url{https://uncertainty-toolbox.github.io/}}~\cite{chung2021uncertainty} to compute each metric. Lower is better for all metrics.

\paragraph{Point performance and negative-log likelihood.}

Tables~\ref{app_tab:results_rmse} and~\ref{app_tab:results_nll} show point~(RMSE) and probabilistic~(NLL) performance of each method.

\begin{table}[h]
\caption{Point~(RMSE~$\downarrow$) performance for each method on each dataset.}
\label{app_tab:results_rmse}
\centering
\scriptsize
\vspace{0.5em}
\begin{tabular}{lccccc}
\toprule
Dataset & NGBoost & PGBM & CBU & IBUG & IBUG+CBU \\
\midrule
Ames & 24580$_{(804)}$ & {\bfseries 23541}$_{(1225)}$ & {\bfseries 22576}$_{(924)}$ & {\bfseries 22942}$_{(1388)}$ & {\bfseries 22391}$_{(1119)}$ \\
Bike & 4.173$_{(0.076)}$ & 3.812$_{(0.225)}$ & {\bfseries 2.850}$_{(0.192)}$ & {\bfseries 2.826}$_{(0.200)}$ & {\bfseries 2.708}$_{(0.202)}$ \\
California & 0.503$_{(0.003)}$ & 0.445$_{(0.001)}$ & 0.449$_{(0.002)}$ & {\bfseries 0.432}$_{(0.001)}$ & 0.434$_{(0.002)}$ \\
Communities & 0.137$_{(0.004)}$ & {\bfseries 0.135}$_{(0.004)}$ & {\bfseries 0.133}$_{(0.004)}$ & {\bfseries 0.133}$_{(0.004)}$ & {\bfseries 0.132}$_{(0.004)}$ \\
Concrete & 5.485$_{(0.182)}$ & 3.840$_{(0.209)}$ & {\bfseries 3.682}$_{(0.202)}$ & {\bfseries 3.629}$_{(0.183)}$ & {\bfseries 3.617}$_{(0.188)}$ \\
Energy & 0.461$_{(0.030)}$ & 0.291$_{(0.022)}$ & 0.381$_{(0.023)}$ & {\bfseries 0.264}$_{(0.023)}$ & 0.303$_{(0.023)}$ \\
Facebook & {\bfseries 20.8}$_{(1.102)}$ & {\bfseries 20.5}$_{(0.867)}$ & {\bfseries 20.1}$_{(0.913)}$ & {\bfseries 20.0}$_{(0.903)}$ & {\bfseries 19.9}$_{(0.929)}$ \\
Kin8nm & 0.176$_{(0.001)}$ & 0.108$_{(0.001)}$ & 0.103$_{(0.001)}$ & {\bfseries 0.086}$_{(0.001)}$ & 0.091$_{(0.001)}$ \\
Life & 2.280$_{(0.032)}$ & 1.678$_{(0.059)}$ & {\bfseries 1.637}$_{(0.058)}$ & {\bfseries 1.652}$_{(0.055)}$ & {\bfseries 1.610}$_{(0.056)}$ \\
MEPS & {\bfseries 23.7}$_{(0.955)}$ & {\bfseries 24.1}$_{(0.760)}$ & {\bfseries 23.5}$_{(0.950)}$ & {\bfseries 23.7}$_{(0.932)}$ & {\bfseries 23.6}$_{(0.945)}$ \\
MSD & 9.121$_{(0.010)}$ & 8.804$_{(0.008)}$ & 8.743$_{(0.008)}$ & 8.747$_{(0.008)}$ & {\bfseries 8.722}$_{(0.008)}$ \\
Naval & 0.002$_{(0.000)}$ & 0.001$_{(0.000)}$ & 0.001$_{(0.000)}$ & {\bfseries 0.000}$_{(0.000)}$ & {\bfseries 0.000}$_{(0.000)}$ \\
News & {\bfseries 11162}$_{(1153)}$ & {\bfseries 11047}$_{(1106)}$ & {\bfseries 11036}$_{(1118)}$ & {\bfseries 11036}$_{(1116)}$ & {\bfseries 11032}$_{(1118)}$ \\
Obesity & 5.315$_{(0.022)}$ & 3.658$_{(0.033)}$ & {\bfseries 3.572}$_{(0.038)}$ & {\bfseries 3.576}$_{(0.037)}$ & {\bfseries 3.567}$_{(0.037)}$ \\
Power & 3.836$_{(0.045)}$ & 3.017$_{(0.056)}$ & {\bfseries 2.924}$_{(0.065)}$ & {\bfseries 2.941}$_{(0.059)}$ & {\bfseries 2.912}$_{(0.063)}$ \\
Protein & 4.525$_{(0.040)}$ & {\bfseries 3.455}$_{(0.021)}$ & 3.520$_{(0.019)}$ & 3.512$_{(0.017)}$ & 3.493$_{(0.018)}$ \\
STAR & 233$_{(2.388)}$ & {\bfseries 229}$_{(2.076)}$ & {\bfseries 229}$_{(1.850)}$ & {\bfseries 228}$_{(1.985)}$ & {\bfseries 228}$_{(1.857)}$ \\
Superconductor & {\bfseries 0.170}$_{(0.101)}$ & 0.425$_{(0.091)}$ & 0.463$_{(0.087)}$ & 0.427$_{(0.088)}$ & 0.419$_{(0.089)}$ \\
Synthetic & {\bfseries 10.2}$_{(0.068)}$ & {\bfseries 10.1}$_{(0.072)}$ & {\bfseries 10.2}$_{(0.072)}$ & {\bfseries 10.1}$_{(0.073)}$ & {\bfseries 10.1}$_{(0.073)}$ \\
Wave & 13537$_{(32.7)}$ & 7895$_{(86.0)}$ & 4803$_{(37.5)}$ & 4899$_{(55.0)}$ & {\bfseries 4020}$_{(33.5)}$ \\
Wine & 0.693$_{(0.010)}$ & {\bfseries 0.603}$_{(0.010)}$ & 0.626$_{(0.010)}$ & {\bfseries 0.596}$_{(0.012)}$ & {\bfseries 0.598}$_{(0.011)}$ \\
Yacht & 0.761$_{(0.106)}$ & 0.809$_{(0.103)}$ & {\bfseries 0.677}$_{(0.124)}$ & {\bfseries 0.668}$_{(0.125)}$ & {\bfseries 0.645}$_{(0.124)}$ \\
\midrule
% NGBoost W-T-L & - & 1-6-15 & 1-4-17 & 1-5-16 & 1-3-18 \\
% PGBM W-T-L & 15-6-1 & - & 6-10-6 & 1-8-13 & 1-9-12 \\
% CBU W-T-L & 17-4-1 & 6-10-6 & - & 0-16-6 & 0-6-16 \\
IBUG W-T-L & 16-5-1 & 13-8-1 & 6-16-0 & - & 2-13-7 \\
IBUG+CBU W-T-L & 18-3-1 & 12-9-1 & 16-6-0 & 7-13-2 & - \\
\bottomrule
\end{tabular}
\end{table}

\begin{table}[h]
\caption{Probabilistic~(NLL~$\downarrow$) performance for each method on each dataset.
}
\label{app_tab:results_nll}
\centering
\scriptsize
\vspace{0.5em}
\begin{tabular}{lccccc}
\toprule
Dataset & NGBoost & PGBM & CBU & IBUG & IBUG+CBU \\
\midrule
Ames & 11.3$_{(0.018)}$ & 11.3$_{(0.029)}$ & 11.9$_{(0.140)}$ & {\bfseries 11.2}$_{(0.030)}$ & 11.5$_{(0.092)}$ \\
Bike & 1.942$_{(0.024)}$ & 1.929$_{(0.078)}$ & {\bfseries 1.184}$_{(0.034)}$ & 1.886$_{(0.056)}$ & 1.382$_{(0.042)}$ \\
California & 0.545$_{(0.007)}$ & 0.580$_{(0.005)}$ & 0.524$_{(0.004)}$ & 0.477$_{(0.010)}$ & {\bfseries 0.437}$_{(0.016)}$ \\
Communities & {\bfseries -0.697}$_{(0.045)}$ & {\bfseries -0.666}$_{(0.034)}$ & {\bfseries -0.614}$_{(0.109)}$ & {\bfseries -0.639}$_{(0.135)}$ & {\bfseries -0.665}$_{(0.116)}$ \\
Concrete & 3.043$_{(0.030)}$ & 2.802$_{(0.083)}$ & {\bfseries 2.766}$_{(0.086)}$ & 2.980$_{(0.146)}$ & {\bfseries 2.695}$_{(0.060)}$ \\
Energy & 0.604$_{(0.192)}$ & {\bfseries 0.322}$_{(0.182)}$ & {\bfseries 0.406}$_{(0.116)}$ & 1.644$_{(0.514)}$ & 0.658$_{(0.165)}$ \\
Facebook & {\bfseries 2.102}$_{(0.026)}$ & 3.116$_{(0.077)}$ & 2.574$_{(0.191)}$ & 2.175$_{(0.067)}$ & 2.276$_{(0.140)}$ \\
Kin8nm & -0.414$_{(0.007)}$ & -0.774$_{(0.034)}$ & -0.772$_{(0.008)}$ & {\bfseries -0.841}$_{(0.008)}$ & {\bfseries -0.847}$_{(0.010)}$ \\
Life & 2.163$_{(0.029)}$ & 1.943$_{(0.033)}$ & 1.932$_{(0.079)}$ & 1.858$_{(0.033)}$ & {\bfseries 1.783}$_{(0.041)}$ \\
MEPS & {\bfseries 3.722}$_{(0.050)}$ & 3.902$_{(0.049)}$ & {\bfseries 3.699}$_{(0.038)}$ & 3.793$_{(0.052)}$ & {\bfseries 3.675}$_{(0.041)}$ \\
MSD & 3.454$_{(0.002)}$ & 3.571$_{(0.002)}$ & 3.415$_{(0.001)}$ & 3.415$_{(0.002)}$ & {\bfseries 3.393}$_{(0.001)}$ \\
Naval & -5.408$_{(0.007)}$ & -5.064$_{(0.338)}$ & -6.141$_{(0.013)}$ & -6.208$_{(0.010)}$ & {\bfseries -6.284}$_{(0.007)}$ \\
News & 10.9$_{(0.268)}$ & {\bfseries 10.7}$_{(0.339)}$ & {\bfseries 10.6}$_{(0.205)}$ & {\bfseries 10.6}$_{(0.208)}$ & {\bfseries 10.6}$_{(0.192)}$ \\
Obesity & 2.940$_{(0.003)}$ & 2.604$_{(0.015)}$ & {\bfseries 2.439}$_{(0.009)}$ & 2.646$_{(0.009)}$ & 2.515$_{(0.010)}$ \\
Power & 2.752$_{(0.032)}$ & {\bfseries 2.518}$_{(0.021)}$ & 2.538$_{(0.019)}$ & 2.575$_{(0.036)}$ & {\bfseries 2.514}$_{(0.017)}$ \\
Protein & 2.840$_{(0.014)}$ & 2.661$_{(0.005)}$ & 2.553$_{(0.009)}$ & 2.653$_{(0.054)}$ & {\bfseries 2.516}$_{(0.010)}$ \\
STAR & 6.869$_{(0.013)}$ & 6.866$_{(0.012)}$ & {\bfseries 6.866}$_{(0.014)}$ & {\bfseries 6.853}$_{(0.008)}$ & {\bfseries 6.852}$_{(0.009)}$ \\
Superconductor & {\bfseries 12.2}$_{(13.1)}$ & {\bfseries 0.035}$_{(0.095)}$ & {\bfseries -0.014}$_{(0.078)}$ & 0.783$_{(0.181)}$ & 0.108$_{(0.036)}$ \\
Synthetic & 3.745$_{(0.007)}$ & {\bfseries 3.742}$_{(0.006)}$ & {\bfseries 3.741}$_{(0.008)}$ & {\bfseries 3.738}$_{(0.007)}$ & {\bfseries 3.738}$_{(0.007)}$ \\
Wave & 10.7$_{(0.002)}$ & 10.3$_{(0.021)}$ & {\bfseries 9.675}$_{(0.003)}$ & 10.5$_{(0.030)}$ & 9.760$_{(0.046)}$ \\
Wine & 1.025$_{(0.013)}$ & 0.952$_{(0.020)}$ & 1.025$_{(0.028)}$ & {\bfseries 0.910}$_{(0.016)}$ & 0.933$_{(0.012)}$ \\
Yacht & 0.905$_{(0.232)}$ & {\bfseries 0.357}$_{(0.162)}$ & 0.951$_{(0.252)}$ & 1.799$_{(1.307)}$ & 0.840$_{(0.310)}$ \\
\midrule
% NGBoost W-T-L & - & 4-8-10 & 2-7-13 & 0-10-12 & 1-6-15 \\
% PGBM W-T-L & 10-8-4 & - & 3-10-9 & 4-11-7 & 2-10-10 \\
% CBU W-T-L & 13-7-2 & 9-10-3 & - & 6-11-5 & 3-6-13 \\
IBUG W-T-L & 12-10-0 & 7-11-4 & 5-11-6 & - & 2-8-12 \\
IBUG+CBU W-T-L & 15-6-1 & 10-10-2 & 13-6-3 & 12-8-2 & - \\
\bottomrule
\end{tabular}
\end{table}

\newpage
\paragraph{Check and interval scores.}

Tables~\ref{app_tab:results_check} and~\ref{app_tab:results_interval} show results when measuring performance with two additional proper scoring rules~\citep{gneiting2007strictly}, \emph{check score} (a.k.a. ``pinball loss'') and \emph{interval score} (evaluation using a pair of quantiles with expected coverage). Under these additional metrics, IBUG+CBU still outperform all other approaches.

\begin{table}[h]
\caption{Probabilistic (check score a.k.a. ``pinball loss''~$\downarrow$) performance.}
\label{app_tab:results_check}
\centering
\scriptsize
\vspace{0.5em}
\begin{tabular}{lccccc}
\toprule
Dataset & NGBoost & PGBM & CBU & IBUG & IBUG+CBU \\
\midrule
Ames & 19358$_{(276)}$ & 5487$_{(179)}$ & 5551$_{(167)}$ & {\bfseries 5266}$_{(185)}$ & {\bfseries 5145}$_{(186)}$ \\
Bike & 6.264$_{(0.482)}$ & 0.597$_{(0.020)}$ & 0.420$_{(0.018)}$ & 0.490$_{(0.024)}$ & {\bfseries 0.386}$_{(0.016)}$ \\
California & 8e+10$_{(8e+10)}$ & 0.112$_{(4e-04)}$ & 0.110$_{(4e-04)}$ & 0.107$_{(5e-04)}$ & {\bfseries 0.104}$_{(4e-04)}$ \\
Communities & 0.034$_{(0.001)}$ & 0.034$_{(1e-03)}$ & 0.034$_{(9e-04)}$ & {\bfseries 0.033}$_{(9e-04)}$ & {\bfseries 0.033}$_{(9e-04)}$ \\
Concrete & 1.722$_{(0.092)}$ & 0.972$_{(0.043)}$ & {\bfseries 0.902}$_{(0.039)}$ & 0.932$_{(0.049)}$ & {\bfseries 0.878}$_{(0.041)}$ \\
Energy & 0.262$_{(0.022)}$ & {\bfseries 0.074}$_{(0.003)}$ & 0.099$_{(0.005)}$ & {\bfseries 0.072}$_{(0.005)}$ & 0.079$_{(0.004)}$ \\
Facebook & 2.024$_{(0.049)}$ & 1.788$_{(0.047)}$ & 1.617$_{(0.030)}$ & 1.551$_{(0.033)}$ & {\bfseries 1.502}$_{(0.035)}$ \\
Kin8nm & 0.048$_{(3e-04)}$ & 0.031$_{(5e-04)}$ & 0.029$_{(3e-04)}$ & {\bfseries 0.026}$_{(3e-04)}$ & {\bfseries 0.026}$_{(3e-04)}$ \\
Life & 1.462$_{(0.739)}$ & 0.411$_{(0.014)}$ & 0.389$_{(0.012)}$ & 0.400$_{(0.011)}$ & {\bfseries 0.368}$_{(0.011)}$ \\
MEPS & {\bfseries 2.779}$_{(0.098)}$ & 3.246$_{(0.046)}$ & 3.050$_{(0.055)}$ & 3.100$_{(0.057)}$ & 3.033$_{(0.056)}$ \\
MSD & 2.283$_{(0.003)}$ & 2.310$_{(0.002)}$ & 2.203$_{(0.002)}$ & 2.226$_{(0.002)}$ & {\bfseries 2.195}$_{(0.002)}$ \\
Naval & 0.002$_{(3e-05)}$ & 2e-04$_{(2e-05)}$ & 2e-04$_{(2e-06)}$ & 1e-04$_{(1e-06)}$ & {\bfseries 1e-04}$_{(8e-07)}$ \\
News & {\bfseries 1102}$_{(23.7)}$ & 1188$_{(26.3)}$ & 1181$_{(26.2)}$ & 1280$_{(20.5)}$ & 1198$_{(26.0)}$ \\
Obesity & 1.620$_{(0.014)}$ & 0.939$_{(0.011)}$ & {\bfseries 0.879}$_{(0.009)}$ & 0.941$_{(0.010)}$ & 0.894$_{(0.009)}$ \\
Power & 1.063$_{(0.012)}$ & 0.773$_{(0.010)}$ & {\bfseries 0.744}$_{(0.011)}$ & 0.778$_{(0.010)}$ & {\bfseries 0.743}$_{(0.011)}$ \\
Protein & 2739$_{(2730)}$ & 0.920$_{(0.006)}$ & 0.902$_{(0.005)}$ & 0.900$_{(0.004)}$ & {\bfseries 0.880}$_{(0.004)}$ \\
STAR & 66.6$_{(0.803)}$ & {\bfseries 65.9}$_{(0.697)}$ & {\bfseries 65.7}$_{(0.647)}$ & {\bfseries 65.4}$_{(0.613)}$ & {\bfseries 65.4}$_{(0.605)}$ \\
Superconductor & 1.215$_{(0.014)}$ & {\bfseries 0.064}$_{(0.002)}$ & 0.076$_{(0.002)}$ & 0.077$_{(0.003)}$ & {\bfseries 0.064}$_{(0.002)}$ \\
Synthetic & 2.918$_{(0.021)}$ & {\bfseries 2.897}$_{(0.020)}$ & {\bfseries 2.898}$_{(0.020)}$ & {\bfseries 2.894}$_{(0.020)}$ & {\bfseries 2.894}$_{(0.020)}$ \\
Wave & 2.9e+05$_{(446)}$ & 1964$_{(37.3)}$ & 1186$_{(5.194)}$ & 1350$_{(8.028)}$ & {\bfseries 1023}$_{(4.813)}$ \\
Wine & 0.194$_{(0.002)}$ & {\bfseries 0.163}$_{(0.003)}$ & 0.170$_{(0.003)}$ & {\bfseries 0.162}$_{(0.003)}$ & {\bfseries 0.162}$_{(0.003)}$ \\
Yacht & 0.594$_{(0.080)}$ & {\bfseries 0.147}$_{(0.021)}$ & {\bfseries 0.142}$_{(0.024)}$ & {\bfseries 0.139}$_{(0.024)}$ & {\bfseries 0.128}$_{(0.023)}$ \\
\midrule
% NGBoost W-T-L & - & 3-4-15 & 2-4-16 & 2-3-17 & 2-3-17 \\
% PGBM W-T-L & 15-4-3 & - & 3-7-12 & 2-9-11 & 1-6-15 \\
% CBU W-T-L & 16-4-2 & 12-7-3 & - & 8-5-9 & 2-2-18 \\
IBUG W-T-L & 17-3-2 & 11-9-2 & 9-5-8 & - & 1-6-15 \\
IBUG+CBU W-T-L & 17-3-2 & 15-6-1 & 18-2-2 & 15-6-1 & - \\
\bottomrule
\end{tabular}
\end{table}

\begin{table}[h]
\caption{Probabilistic (interval score~$\downarrow$) performance.}
\label{app_tab:results_interval}
\centering
\scriptsize
\vspace{0.5em}
\begin{tabular}{lccccc}
\toprule
Dataset & NGBoost & PGBM & CBU & IBUG & IBUG+CBU \\
\midrule
Ames & 2.0e+05$_{(3492)}$ & 59165$_{(1952)}$ & 66337$_{(2499)}$ & {\bfseries 57219}$_{(1941)}$ & {\bfseries 55551}$_{(1994)}$ \\
Bike & 66.4$_{(6.425)}$ & 7.048$_{(0.411)}$ & {\bfseries 4.270}$_{(0.136)}$ & 6.775$_{(0.324)}$ & {\bfseries 4.263}$_{(0.191)}$ \\
California & 1e+12$_{(1e+12)}$ & 1.257$_{(0.008)}$ & 1.168$_{(0.008)}$ & 1.230$_{(0.020)}$ & {\bfseries 1.119}$_{(0.006)}$ \\
Communities & 0.361$_{(0.013)}$ & 0.366$_{(0.012)}$ & 0.352$_{(0.011)}$ & {\bfseries 0.343}$_{(0.013)}$ & {\bfseries 0.339}$_{(0.011)}$ \\
Concrete & 17.2$_{(0.917)}$ & 11.1$_{(0.698)}$ & {\bfseries 10.3}$_{(0.491)}$ & 12.1$_{(0.800)}$ & {\bfseries 10.1}$_{(0.523)}$ \\
Energy & 2.711$_{(0.198)}$ & {\bfseries 0.814}$_{(0.050)}$ & 0.998$_{(0.067)}$ & 0.912$_{(0.083)}$ & {\bfseries 0.819}$_{(0.064)}$ \\
Facebook & 28.4$_{(0.909)}$ & 26.8$_{(1.125)}$ & 21.6$_{(0.692)}$ & {\bfseries 17.4}$_{(0.476)}$ & {\bfseries 17.1}$_{(0.509)}$ \\
Kin8nm & 0.458$_{(0.003)}$ & 0.311$_{(0.007)}$ & 0.292$_{(0.005)}$ & 0.302$_{(0.009)}$ & {\bfseries 0.262}$_{(0.005)}$ \\
Life & 17.5$_{(9.900)}$ & 5.051$_{(0.239)}$ & 4.617$_{(0.198)}$ & 5.093$_{(0.264)}$ & {\bfseries 4.332}$_{(0.207)}$ \\
MEPS & 42.2$_{(1.973)}$ & 44.3$_{(1.294)}$ & {\bfseries 37.7}$_{(1.223)}$ & {\bfseries 38.3}$_{(1.375)}$ & {\bfseries 37.2}$_{(1.254)}$ \\
MSD & 24.5$_{(0.039)}$ & 24.8$_{(0.035)}$ & 22.3$_{(0.020)}$ & 22.4$_{(0.029)}$ & {\bfseries 22.0}$_{(0.025)}$ \\
Naval & 0.014$_{(3e-04)}$ & 0.003$_{(3e-04)}$ & 0.002$_{(2e-05)}$ & 0.001$_{(3e-05)}$ & {\bfseries 0.001}$_{(1e-05)}$ \\
News & {\bfseries 16557}$_{(519)}$ & {\bfseries 16242}$_{(556)}$ & {\bfseries 16166}$_{(580)}$ & 18694$_{(373)}$ & {\bfseries 16426}$_{(551)}$ \\
Obesity & 15.5$_{(0.153)}$ & 9.731$_{(0.125)}$ & {\bfseries 8.747}$_{(0.083)}$ & 10.7$_{(0.139)}$ & 9.162$_{(0.086)}$ \\
Power & 10.6$_{(0.136)}$ & 8.146$_{(0.122)}$ & {\bfseries 7.837}$_{(0.165)}$ & 8.512$_{(0.152)}$ & {\bfseries 7.803}$_{(0.156)}$ \\
Protein & 36689$_{(36570)}$ & 10.1$_{(0.149)}$ & 9.277$_{(0.062)}$ & 9.322$_{(0.045)}$ & {\bfseries 8.853}$_{(0.052)}$ \\
STAR & 642$_{(7.014)}$ & 637$_{(6.564)}$ & {\bfseries 636}$_{(6.131)}$ & {\bfseries 630}$_{(4.545)}$ & {\bfseries 630}$_{(4.968)}$ \\
Superconductor & 12.0$_{(0.133)}$ & 0.776$_{(0.023)}$ & 0.755$_{(0.030)}$ & 1.150$_{(0.060)}$ & {\bfseries 0.692}$_{(0.033)}$ \\
Synthetic & 28.4$_{(0.228)}$ & {\bfseries 28.1}$_{(0.188)}$ & {\bfseries 28.1}$_{(0.211)}$ & {\bfseries 28.0}$_{(0.197)}$ & {\bfseries 28.0}$_{(0.199)}$ \\
Wave & 3e+06$_{(3727)}$ & 20256$_{(323)}$ & 11748$_{(55.8)}$ & 16669$_{(117)}$ & {\bfseries 10569}$_{(47.4)}$ \\
Wine & 1.930$_{(0.023)}$ & {\bfseries 1.723}$_{(0.032)}$ & 1.793$_{(0.035)}$ & {\bfseries 1.716}$_{(0.030)}$ & {\bfseries 1.692}$_{(0.031)}$ \\
Yacht & 5.798$_{(0.808)}$ & {\bfseries 1.621}$_{(0.248)}$ & {\bfseries 1.796}$_{(0.419)}$ & {\bfseries 1.955}$_{(0.433)}$ & {\bfseries 1.619}$_{(0.406)}$ \\
\midrule
% NGBoost W-T-L & - & 1-8-13 & 0-5-17 & 1-3-18 & 0-4-18 \\
% PGBM W-T-L & 13-8-1 & - & 3-7-12 & 5-9-8 & 0-6-16 \\
% CBU W-T-L & 17-5-0 & 12-7-3 & - & 10-8-4 & 2-4-16 \\
IBUG W-T-L & 18-3-1 & 8-9-5 & 4-8-10 & - & 0-6-16 \\
IBUG+CBU W-T-L & 18-4-0 & 16-6-0 & 16-4-2 & 16-6-0 & - \\
\bottomrule
\end{tabular}
\end{table}

\newpage
\paragraph{Calibration error.}

Table~\ref{app_tab:calibration} shows the average MACE~(mean absolute calibration error) and sharpness scores. Sharpness quantifies the average of the standard deviations and thus does not depend on the actual ground-truth label; therefore, MACE and sharpness are shown together, with better methods having both low calibration error and low sharpness scores.

We observe that NGBoost is particularly well-calibrated, but lacks sharpness, meaning the prediction intervals of NGBoost are generally too wide. PGBM tends to have very sharp prediction intervals, but high calibration error. In contrast, CBU tends to achieve both low calibration error and high sharpness in relation to the other methods. However, these results are with variance calibration~(\S\ref{sec:posterior}), which we note has a significant impact on the CBU approach. For example, the median improvement in MACE score~(over datasets) for CBU when using variance calibration vs. without is greater than 3x.

\begin{table}[h]
\caption{Probabilistic (MACE~$\downarrow$ / sharpness~$\downarrow$) performance. Standard errors are omitted for brevity.}
\label{app_tab:calibration}
\centering
\small
\vspace{0.5em}
\begin{tabular}{lccccc}
\toprule
Dataset & NGBoost & PGBM & CBU & IBUG & IBUG+CBU \\
\midrule
Ames & 0.082/74148 & {\bfseries 0.040}/{\bfseries 18432} & 0.073/18867 & 0.068/23186 & 0.063/19791 \\
Bike & 0.070/190 & 0.140/2.077 & {\bfseries 0.045}/2.136 & 0.096/{\bfseries 1.272} & 0.051/1.595 \\
California & {\bfseries 0.014}/3e+13 & 0.053/{\bfseries 0.344} & 0.021/0.367 & 0.089/0.382 & 0.037/0.364 \\
Communities & 0.039/0.129 & 0.067/{\bfseries 0.120} & 0.051/0.136 & {\bfseries 0.035}/0.133 & 0.048/0.133 \\
Concrete & 0.056/6.889 & 0.068/3.002 & 0.096/3.177 & 0.115/{\bfseries 2.503} & {\bfseries 0.054}/2.708 \\
Energy & 0.127/1.497 & 0.093/0.252 & 0.054/0.373 & 0.103/{\bfseries 0.249} & {\bfseries 0.053}/0.296 \\
Facebook & 0.094/9.171 & 0.206/{\bfseries 4.309} & 0.072/7.332 & {\bfseries 0.061}/18.9 & 0.091/12.5 \\
Kin8nm & 0.020/0.182 & 0.037/0.108 & {\bfseries 0.020}/0.096 & 0.126/{\bfseries 0.071} & 0.045/0.081 \\
Life & {\bfseries 0.039}/111 & 0.069/{\bfseries 1.103} & 0.079/1.189 & 0.115/1.401 & 0.069/1.216 \\
MEPS & {\bfseries 0.030}/{\bfseries 6.680} & 0.074/8.200 & 0.119/14.1 & 0.086/17.2 & 0.106/15.3 \\
MSD & {\bfseries 0.007}/7.749 & 0.036/{\bfseries 7.436} & 0.012/8.137 & 0.039/9.088 & 0.031/8.519 \\
Naval & {\bfseries 0.032}/0.006 & 0.279/1e-03 & 0.048/6e-04 & 0.059/{\bfseries 5e-04} & 0.086/5e-04 \\
News & 0.104/{\bfseries 2170} & {\bfseries 0.085}/3289 & 0.101/2975 & 0.202/4803 & 0.109/3498 \\
Obesity & 0.012/5.996 & 0.065/3.451 & {\bfseries 0.006}/3.102 & 0.095/2.957 & 0.043/{\bfseries 2.956} \\
Power & 0.020/3.761 & 0.026/2.558 & {\bfseries 0.018}/{\bfseries 2.299} & 0.030/3.328 & 0.019/2.729 \\
Protein & 0.029/2e+06 & 0.076/{\bfseries 2.823} & 0.037/3.144 & {\bfseries 0.016}/3.977 & 0.046/3.498 \\
STAR & 0.025/248 & 0.031/250 & 0.030/{\bfseries 242} & {\bfseries 0.023}/245 & 0.025/243 \\
Superconductor & 0.074/7.993 & 0.102/0.240 & {\bfseries 0.028}/0.322 & 0.205/{\bfseries 0.208} & 0.041/0.240 \\
Synthetic & {\bfseries 0.012}/10.4 & 0.023/10.9 & 0.019/{\bfseries 10.4} & 0.012/10.4 & 0.014/10.4 \\
Wave & 0.129/1e+06 & 0.018/6403 & {\bfseries 0.007}/{\bfseries 4310} & 0.089/6496 & 0.042/5127 \\
Wine & {\bfseries 0.017}/0.694 & 0.070/{\bfseries 0.540} & 0.027/0.575 & 0.091/0.643 & 0.061/0.600 \\
Yacht & 0.115/4.057 & 0.174/0.690 & 0.098/0.508 & 0.124/{\bfseries 0.371} & {\bfseries 0.078}/0.412 \\
% \midrule
% NGBoost W-T-L & -/- & 13-7-2/2-7-13 & 7-11-4/3-8-11 & 13-6-3/4-7-11 & 9-10-3/4-7-11 \\
% PGBM W-T-L & 2-7-13/13-7-2 & -/- & 3-8-11/6-12-4 & 8-8-6/8-9-5 & 2-10-10/6-9-7 \\
% CBU W-T-L & 4-11-7/11-8-3 & 11-8-3/4-12-6 & -/- & 9-10-3/8-8-6 & 9-10-3/8-7-7 \\
% IBUG W-T-L & 3-6-13/11-7-4 & 6-8-8/5-9-8 & 3-10-9/6-8-8 & -/- & 3-7-12/5-7-10 \\
% IBUG+CBU W-T-L & 3-10-9/11-7-4 & 10-10-2/7-9-6 & 3-10-9/7-7-8 & 12-7-3/10-7-5 & -/- \\
\bottomrule
\end{tabular}
\end{table}

\newpage

\subsection{Runtime}
\label{app_sec:runtime}

Tables~\ref{app_tab:runtime_train} and~\ref{app_tab:runtime_prediction} provide detailed runtime results for each method. Results are averaged over 10 folds, and standard deviations are shown in subscripted parentheses; lower is better. The last row in each table shows the Geometric mean over all datasets.

\begin{table}[h]
\caption{Total train~(including tuning) time~(in seconds).}
\label{app_tab:runtime_train}
\centering
\scriptsize
\vspace{0.5em}
\begin{tabular}{lcccc}
\toprule
Dataset & NGBoost & PGBM & CBU & IBUG \\
\midrule
Ames & {\bfseries 417}$_{(587)}$ & 1.4e+05$_{(25170)}$ & 23181$_{(258)}$ & 22264$_{(796)}$ \\
Bike & {\bfseries 195}$_{(143)}$ & 58246$_{(8207)}$ & 23207$_{(239)}$ & 22417$_{(794)}$ \\
California & {\bfseries 315}$_{(90.8)}$ & 16958$_{(1173)}$ & 23141$_{(253)}$ & 22530$_{(649)}$ \\
Communities & {\bfseries 38.0}$_{(21.2)}$ & 24491$_{(4246)}$ & 23023$_{(260)}$ & 22429$_{(483)}$ \\
Concrete & {\bfseries 57.1}$_{(22.9)}$ & 4130$_{(3621)}$ & 22953$_{(265)}$ & 22402$_{(577)}$ \\
Energy & {\bfseries 35.3}$_{(33.0)}$ & 2706$_{(601)}$ & 22783$_{(278)}$ & 22423$_{(602)}$ \\
Facebook & {\bfseries 731}$_{(659)}$ & 3.5e+05$_{(58586)}$ & 23061$_{(310)}$ & 23145$_{(517)}$ \\
Kin8nm & {\bfseries 77.8}$_{(39.6)}$ & 10489$_{(2181)}$ & 23142$_{(296)}$ & 22694$_{(529)}$ \\
Life & {\bfseries 105}$_{(87.4)}$ & 83814$_{(25313)}$ & 23082$_{(273)}$ & 20531$_{(7050)}$ \\
MEPS & {\bfseries 351}$_{(477)}$ & 2.3e+05$_{(41039)}$ & 23139$_{(327)}$ & 20491$_{(7004)}$ \\
MSD & {\bfseries 11720}$_{(1022)}$ & 2.2e+05$_{(34478)}$ & 23972$_{(258)}$ & 52760$_{(16670)}$ \\
Naval & {\bfseries 847}$_{(1804)}$ & 38882$_{(11481)}$ & 23133$_{(210)}$ & 20607$_{(7059)}$ \\
News & {\bfseries 2275}$_{(138)}$ & 2.4e+05$_{(60642)}$ & 22960$_{(258)}$ & 22492$_{(448)}$ \\
Obesity & {\bfseries 1569}$_{(2208)}$ & 3.2e+05$_{(64847)}$ & 23169$_{(259)}$ & 21040$_{(7086)}$ \\
Power & {\bfseries 107}$_{(53.0)}$ & 12556$_{(1459)}$ & 23042$_{(338)}$ & 22445$_{(298)}$ \\
Protein & {\bfseries 1430}$_{(2298)}$ & 40132$_{(5779)}$ & 23043$_{(277)}$ & 22865$_{(344)}$ \\
STAR & {\bfseries 17.0}$_{(5.788)}$ & 20797$_{(3481)}$ & 22852$_{(263)}$ & 22074$_{(432)}$ \\
Superconductor & {\bfseries 215}$_{(29.3)}$ & 2.1e+05$_{(42859)}$ & 23291$_{(706)}$ & 22503$_{(426)}$ \\
Synthetic & {\bfseries 439}$_{(351)}$ & 1.0e+05$_{(15729)}$ & 23068$_{(333)}$ & 22394$_{(532)}$ \\
Wave & {\bfseries 3487}$_{(342)}$ & 1.0e+05$_{(16204)}$ & 23394$_{(173)}$ & 44282$_{(16994)}$ \\
Wine & {\bfseries 33.1}$_{(19.2)}$ & 15067$_{(2804)}$ & 20942$_{(7191)}$ & 22269$_{(451)}$ \\
Yacht & {\bfseries 51.3}$_{(41.8)}$ & 1965$_{(87.7)}$ & 22915$_{(239)}$ & 22184$_{(433)}$ \\
\midrule
% NGBoost W-L & - & 22-0 & 20-2 & 10-12 \\
% PGBM W-L & 0-22 & - & 0-22 & 0-22 \\
% CBU W-L & 2-20 & 22-0 & - & 2-20 \\
% IBUG W-L & 12-10 & 22-0 & 20-2 & - \\
Geo. mean & {\bfseries 265} & 43604 & 23017 & 23726 \\
\bottomrule
\end{tabular}
\end{table}

\begin{table}[h]
\caption{Average prediction time per text example~(in milliseconds).}
\label{app_tab:runtime_prediction}
\centering
\scriptsize
\vspace{0.5em}
\begin{tabular}{lcccc}
\toprule
Dataset & NGBoost & PGBM & CBU & IBUG \\
\midrule
Ames & {\bfseries 5.583}$_{(5.778)}$ & 9.505$_{(2.426)}$ & {\bfseries 0.066}$_{(0.010)}$ & 4.851$_{(2.766)}$ \\
Bike & {\bfseries 0.514}$_{(0.815)}$ & {\bfseries 7.705}$_{(8.198)}$ & {\bfseries 0.010}$_{(0.002)}$ & 61.6$_{(21.1)}$ \\
California & 0.243$_{(0.082)}$ & {\bfseries 5.659}$_{(9.562)}$ & {\bfseries 0.004}$_{(0.001)}$ & 23.4$_{(5.265)}$ \\
Communities & 0.393$_{(0.170)}$ & 11.5$_{(0.941)}$ & {\bfseries 0.027}$_{(0.010)}$ & 1.803$_{(1.118)}$ \\
Concrete & 2.154$_{(0.884)}$ & {\bfseries 44.8}$_{(57.5)}$ & {\bfseries 0.043}$_{(0.019)}$ & 1.876$_{(0.726)}$ \\
Energy & 1.830$_{(1.369)}$ & 32.9$_{(12.1)}$ & {\bfseries 0.053}$_{(0.027)}$ & 1.135$_{(0.300)}$ \\
Facebook & 0.533$_{(0.469)}$ & {\bfseries 5.148}$_{(7.166)}$ & {\bfseries 0.024}$_{(0.004)}$ & 105$_{(62.4)}$ \\
Kin8nm & 0.194$_{(0.107)}$ & 168$_{(89.7)}$ & {\bfseries 0.008}$_{(0.002)}$ & 4.713$_{(0.454)}$ \\
Life & 1.466$_{(1.171)}$ & 31.5$_{(28.1)}$ & {\bfseries 0.064}$_{(0.037)}$ & 6.376$_{(1.275)}$ \\
MEPS & 0.465$_{(0.121)}$ & {\bfseries 9.510}$_{(23.6)}$ & {\bfseries 0.005}$_{(0.002)}$ & 8.845$_{(7.866)}$ \\
MSD & 1.712$_{(0.347)}$ & 25.3$_{(1.933)}$ & {\bfseries 0.003}$_{(7e-04)}$ & 603$_{(97.4)}$ \\
Naval & 0.280$_{(0.129)}$ & {\bfseries 187}$_{(253)}$ & {\bfseries 0.010}$_{(0.007)}$ & 41.9$_{(22.3)}$ \\
News & 0.577$_{(0.100)}$ & 0.771$_{(0.403)}$ & {\bfseries 0.002}$_{(1e-03)}$ & 40.0$_{(38.0)}$ \\
Obesity & 0.988$_{(0.475)}$ & 10.1$_{(6.503)}$ & {\bfseries 0.020}$_{(0.003)}$ & 110$_{(9.535)}$ \\
Power & 0.252$_{(0.115)}$ & 6.904$_{(4.256)}$ & {\bfseries 0.007}$_{(0.002)}$ & 18.3$_{(17.5)}$ \\
Protein & 0.154$_{(0.080)}$ & 130$_{(73.7)}$ & {\bfseries 0.004}$_{(7e-04)}$ & 90.8$_{(25.7)}$ \\
STAR & 0.353$_{(0.121)}$ & 10.0$_{(1.191)}$ & {\bfseries 0.038}$_{(0.010)}$ & 0.937$_{(0.369)}$ \\
Superconductor & 0.096$_{(0.055)}$ & {\bfseries 27.1}$_{(79.5)}$ & {\bfseries 0.005}$_{(0.002)}$ & 52.6$_{(25.8)}$ \\
Synthetic & {\bfseries 0.482}$_{(0.541)}$ & 2.461$_{(0.524)}$ & {\bfseries 0.009}$_{(0.005)}$ & {\bfseries 7.595}$_{(10.1)}$ \\
Wave & {\bfseries 1.018}$_{(1.339)}$ & {\bfseries 9.116}$_{(17.0)}$ & {\bfseries 0.003}$_{(3e-04)}$ & 719$_{(106)}$ \\
Wine & 0.135$_{(0.061)}$ & {\bfseries 280}$_{(281)}$ & {\bfseries 0.009}$_{(0.002)}$ & 6.001$_{(1.696)}$ \\
Yacht & 4.192$_{(2.669)}$ & 71.5$_{(12.1)}$ & {\bfseries 0.124}$_{(0.081)}$ & 1.237$_{(0.334)}$ \\
\midrule
% NGBoost W-T-L & - & 15-7-0 & 0-1-21 & 17-4-1 \\
% PGBM W-T-L & 0-7-15 & - & 0-5-17 & 7-6-9 \\
% CBU W-T-L & 21-1-0 & 17-5-0 & - & 22-0-0 \\
% IBUG-CB W-T-L & 1-4-17 & 9-6-7 & 0-0-22 & - \\
Geo. mean & 0.585 & 18.5 & {\bfseries 0.013} & 15.9 \\
\bottomrule
\end{tabular}
\end{table}

\newpage

\section{Additional Experiments}
\label{app_sec:additional_experiments}

In this section, we present additional experimental results.

\subsection{Probabilistic Performance Without Variance Calibration}
\label{app_sec:no_calibration}

Tables~\ref{app_tab:crps_no_calibration}~and~\ref{app_tab:nll_no_calibration} show the probabilistic performance of each method \emph{without} variance calibration. Even without variance calibration, IBUG+CBU generally outperforms competing methods. Standard errors are shown in subscripted parentheses.

\begin{table}[h]
\caption{Probabilistic~(CRPS~$\downarrow$) performance \emph{without} variance calibration.}
\label{app_tab:crps_no_calibration}
\centering
\scriptsize
\vspace{0.5em}
\begin{tabular}{lccccc}
\toprule
Dataset & NGBoost & PGBM & CBU & IBUG & IBUG+CBU \\
\midrule
Ames & 38279$_{(564)}$ & 12173$_{(484)}$ & 11948$_{(386)}$ & {\bfseries 10442}$_{(373)}$ & {\bfseries 10208}$_{(392)}$ \\
Bike & 13.9$_{(1.856)}$ & 1.274$_{(0.054)}$ & {\bfseries 0.835}$_{(0.035)}$ & 1.899$_{(0.224)}$ & 1.219$_{(0.105)}$ \\
California & 2e+11$_{(2e+11)}$ & 0.227$_{(0.004)}$ & 0.221$_{(0.001)}$ & 0.213$_{(1e-03)}$ & {\bfseries 0.207}$_{(9e-04)}$ \\
Communities & 0.068$_{(0.002)}$ & 0.077$_{(0.004)}$ & 0.070$_{(0.002)}$ & {\bfseries 0.065}$_{(0.002)}$ & {\bfseries 0.065}$_{(0.002)}$ \\
Concrete & 3.395$_{(0.181)}$ & 1.932$_{(0.088)}$ & 1.994$_{(0.095)}$ & 1.938$_{(0.079)}$ & {\bfseries 1.780}$_{(0.085)}$ \\
Energy & 0.539$_{(0.042)}$ & {\bfseries 0.151}$_{(0.007)}$ & 0.207$_{(0.010)}$ & 0.481$_{(0.041)}$ & 0.293$_{(0.022)}$ \\
Facebook & 4.022$_{(0.099)}$ & 3.860$_{(0.149)}$ & 3.214$_{(0.058)}$ & 3.072$_{(0.066)}$ & {\bfseries 2.971}$_{(0.072)}$ \\
Kin8nm & 0.095$_{(6e-04)}$ & 0.069$_{(0.003)}$ & 0.063$_{(8e-04)}$ & 0.052$_{(4e-04)}$ & {\bfseries 0.052}$_{(6e-04)}$ \\
Life & 2.897$_{(1.465)}$ & 0.836$_{(0.035)}$ & 0.852$_{(0.030)}$ & 0.794$_{(0.022)}$ & {\bfseries 0.739}$_{(0.024)}$ \\
MEPS & {\bfseries 5.529}$_{(0.196)}$ & 6.725$_{(0.126)}$ & 6.050$_{(0.109)}$ & 6.146$_{(0.113)}$ & 6.022$_{(0.114)}$ \\
MSD & 4.525$_{(0.005)}$ & 5.767$_{(0.006)}$ & 4.364$_{(0.004)}$ & 4.410$_{(0.005)}$ & {\bfseries 4.342}$_{(0.004)}$ \\
Naval & 0.003$_{(6e-05)}$ & 0.005$_{(0.002)}$ & 3e-04$_{(3e-06)}$ & 3e-04$_{(2e-06)}$ & {\bfseries 3e-04}$_{(2e-06)}$ \\
News & {\bfseries 2476}$_{(38.9)}$ & 2628$_{(94.9)}$ & 2712$_{(59.7)}$ & 2669$_{(43.9)}$ & 2593$_{(49.7)}$ \\
Obesity & 3.208$_{(0.028)}$ & 1.860$_{(0.022)}$ & {\bfseries 1.754}$_{(0.017)}$ & 1.882$_{(0.019)}$ & 1.772$_{(0.018)}$ \\
Power & 2.104$_{(0.024)}$ & 1.585$_{(0.057)}$ & 1.572$_{(0.024)}$ & 1.542$_{(0.020)}$ & {\bfseries 1.488}$_{(0.022)}$ \\
Protein & 5427$_{(5409)}$ & 1.932$_{(0.014)}$ & 1.822$_{(0.010)}$ & 1.785$_{(0.008)}$ & {\bfseries 1.740}$_{(0.009)}$ \\
STAR & 132$_{(1.697)}$ & 157$_{(6.908)}$ & 132$_{(1.540)}$ & {\bfseries 129}$_{(1.225)}$ & {\bfseries 130}$_{(1.327)}$ \\
Superconductor & 3.200$_{(0.031)}$ & {\bfseries 0.134}$_{(0.005)}$ & 0.151$_{(0.004)}$ & 0.303$_{(0.025)}$ & 0.201$_{(0.013)}$ \\
Synthetic & 5.778$_{(0.043)}$ & 6.946$_{(0.242)}$ & {\bfseries 5.769}$_{(0.049)}$ & {\bfseries 5.731}$_{(0.040)}$ & {\bfseries 5.735}$_{(0.042)}$ \\
Wave & 5.7e+05$_{(886)}$ & 4152$_{(247)}$ & {\bfseries 2350}$_{(10.3)}$ & 4905$_{(12.3)}$ & 3112$_{(8.952)}$ \\
Wine & 0.385$_{(0.005)}$ & 0.383$_{(0.015)}$ & 0.355$_{(0.007)}$ & {\bfseries 0.322}$_{(0.006)}$ & {\bfseries 0.321}$_{(0.006)}$ \\
Yacht & 1.187$_{(0.142)}$ & {\bfseries 0.310}$_{(0.056)}$ & {\bfseries 0.291}$_{(0.050)}$ & 0.644$_{(0.068)}$ & 0.394$_{(0.042)}$ \\
\midrule
% NGBoost W-T-L & - & 5-7-10 & 2-6-14 & 2-4-16 & 2-3-17 \\
% PGBM W-T-L & 10-7-5 & - & 2-9-11 & 6-4-12 & 3-4-15 \\
% CBU W-T-L & 14-6-2 & 11-9-2 & - & 8-4-10 & 6-2-14 \\
IBUG W-T-L & 16-4-2 & 12-4-6 & 10-4-8 & - & 0-5-17 \\
IBUG+CBU W-T-L & 17-3-2 & 15-4-3 & 14-2-6 & 17-5-0 & - \\
\bottomrule
\end{tabular}
\end{table}

\begin{table}[h]
\caption{Probabilistic~(NLL~$\downarrow$) performance \emph{without} variance calibration.}
\label{app_tab:nll_no_calibration}
\centering
\scriptsize
\vspace{0.5em}
\begin{tabular}{lccccc}
\toprule
Dataset & NGBoost & PGBM & CBU & IBUG & IBUG+CBU \\
\midrule
Ames & 11.3$_{(0.023)}$ & 23.7$_{(6.813)}$ & 1676$_{(1093)}$ & {\bfseries 11.2}$_{(0.031)}$ & {\bfseries 11.3}$_{(0.070)}$ \\
Bike & 1.942$_{(0.024)}$ & 11.1$_{(4.073)}$ & {\bfseries 1.264}$_{(0.080)}$ & 2.958$_{(0.106)}$ & 2.386$_{(0.091)}$ \\
California & 0.551$_{(0.010)}$ & 7.821$_{(7.026)}$ & 2.261$_{(0.341)}$ & 0.484$_{(0.009)}$ & {\bfseries 0.375}$_{(0.009)}$ \\
Communities & {\bfseries -7e-01}$_{(0.056)}$ & 20.7$_{(7.235)}$ & 2.438$_{(1.168)}$ & {\bfseries -6e-01}$_{(0.136)}$ & {\bfseries -4e-01}$_{(0.269)}$ \\
Concrete & 3.062$_{(0.031)}$ & 3.102$_{(0.277)}$ & 684$_{(358)}$ & {\bfseries 2.848}$_{(0.055)}$ & {\bfseries 2.822}$_{(0.093)}$ \\
Energy & {\bfseries 0.670}$_{(0.250)}$ & {\bfseries 0.481}$_{(0.341)}$ & 6.129$_{(3.597)}$ & 1.461$_{(0.113)}$ & 1.048$_{(0.162)}$ \\
Facebook & 2.099$_{(0.026)}$ & 14.7$_{(5.523)}$ & 5.147$_{(1.834)}$ & 2.195$_{(0.070)}$ & {\bfseries 2.044}$_{(0.045)}$ \\
Kin8nm & -4e-01$_{(0.007)}$ & 35.0$_{(23.1)}$ & 59.5$_{(24.2)}$ & -8e-01$_{(0.009)}$ & {\bfseries -9e-01}$_{(0.024)}$ \\
Life & 2.188$_{(0.044)}$ & 23.5$_{(20.6)}$ & 71.9$_{(38.8)}$ & {\bfseries 1.889}$_{(0.038)}$ & {\bfseries 1.885}$_{(0.100)}$ \\
MEPS & {\bfseries 3.732}$_{(0.056)}$ & 11.3$_{(3.246)}$ & {\bfseries 3.722}$_{(0.044)}$ & 3.820$_{(0.064)}$ & {\bfseries 3.678}$_{(0.054)}$ \\
MSD & 3.454$_{(0.002)}$ & 65.6$_{(0.162)}$ & 3.450$_{(0.004)}$ & 3.415$_{(0.002)}$ & {\bfseries 3.383}$_{(0.002)}$ \\
Naval & -5e+00$_{(0.007)}$ & -4e+00$_{(0.357)}$ & -5e+00$_{(0.057)}$ & -6e+00$_{(0.007)}$ & {\bfseries -6e+00}$_{(0.006)}$ \\
News & {\bfseries 10.9}$_{(0.335)}$ & 130$_{(49.5)}$ & {\bfseries 10.8}$_{(0.368)}$ & {\bfseries 11.0}$_{(0.415)}$ & {\bfseries 10.7}$_{(0.307)}$ \\
Obesity & 2.940$_{(0.003)}$ & 2.603$_{(0.015)}$ & {\bfseries 2.488}$_{(0.009)}$ & 2.646$_{(0.009)}$ & {\bfseries 2.493}$_{(0.009)}$ \\
Power & 2.769$_{(0.042)}$ & 11.0$_{(8.270)}$ & 5.304$_{(0.672)}$ & {\bfseries 2.575}$_{(0.036)}$ & {\bfseries 2.569}$_{(0.057)}$ \\
Protein & 2.841$_{(0.015)}$ & 5.299$_{(0.268)}$ & 3.291$_{(0.042)}$ & 2.747$_{(0.123)}$ & {\bfseries 2.531}$_{(0.028)}$ \\
STAR & 6.872$_{(0.015)}$ & 23.2$_{(5.203)}$ & 6.989$_{(0.040)}$ & {\bfseries 6.853}$_{(0.008)}$ & {\bfseries 6.857}$_{(0.012)}$ \\
Superconductor & {\bfseries 12.1}$_{(13.4)}$ & 10.5$_{(4.631)}$ & {\bfseries -6e-03}$_{(0.093)}$ & 1.151$_{(0.111)}$ & 0.602$_{(0.094)}$ \\
Synthetic & 3.746$_{(0.008)}$ & 27.3$_{(6.288)}$ & 3.782$_{(0.032)}$ & {\bfseries 3.738}$_{(0.007)}$ & {\bfseries 3.744}$_{(0.010)}$ \\
Wave & 10.7$_{(0.002)}$ & 22.2$_{(7.985)}$ & {\bfseries 9.679}$_{(0.004)}$ & 10.9$_{(0.004)}$ & 10.4$_{(0.004)}$ \\
Wine & 1.029$_{(0.014)}$ & 109$_{(24.9)}$ & 578$_{(428)}$ & {\bfseries 0.910}$_{(0.016)}$ & 0.968$_{(0.030)}$ \\
Yacht & {\bfseries 0.904}$_{(0.232)}$ & 7.227$_{(4.096)}$ & 4.770$_{(2.330)}$ & 1.502$_{(0.308)}$ & {\bfseries 1.204}$_{(0.519)}$ \\
\midrule
% NGBoost W-T-L & - & 11-10-1 & 6-13-3 & 4-7-11 & 2-7-13 \\
% PGBM W-T-L & 1-10-11 & - & 0-11-11 & 2-10-10 & 1-10-11 \\
% CBU W-T-L & 3-13-6 & 11-11-0 & - & 5-9-8 & 3-11-8 \\
IBUG W-T-L & 11-7-4 & 10-10-2 & 8-9-5 & - & 1-10-11 \\
IBUG+CBU W-T-L & 13-7-2 & 11-10-1 & 8-11-3 & 11-10-1 & - \\
\bottomrule
\end{tabular}
\end{table}

\newpage

\subsection{Comparison to~$k$-Nearest Neighbors}
\label{app_sec:knn}

In this section, we compare IBUG to $k$-nearest neighbors, in which similarity is defined by Euclidean distance. For the nearest-neighbors approach, we tune two different~$k$ values, one for estimating the conditional mean, and one for estimating the variance. We also apply standard scaling to the data before training, and denote this method~$k$NN in our results. Table~\ref{app_tab:knn} shows that IBUG is consistently better than~$k$NN in terms of probabilistic performance. However, we note that point predictions from GBRTs is typically better than~$k$NNs, thus we also compare IBUG to a variant of~$k$NN that uses CatBoost as a base model to estimate the conditional mean.

\paragraph{Euclidean Distance vs. Affinity.}

To test which similarity measure~(Euclidean distance or affinity) is more effective, we use the output from CatBoost to model the conditional mean, we then use $k$NN or IBUG to find their respective~$k$-nearest training examples to estimate the variance; we denote these methods~$k$NN-CB\footnote{Again, we apply standard scaling to the data before training~$k$NN-CB.} and IBUG-CB. For~$k$NN-CB, we also reduce the dimensionality of the data by only using the most important features identified by the CatBoost model;\footnote{We tune the number of important features to use for~$k$NN-CB using values [5, 10, 20].} this helps~$k$NN-CB combat the curse of dimensionality when computing similarity. Results of this comparison are in Table~\ref{app_tab:knn}, in which we observe IBUG-CB is always statistically the same or better than $k$NN-CB. These results suggest affinity is a more effective similarity measure than Euclidean distance for uncertainty estimation in GBRTs.

\begin{table}[ht]
\caption{Probabilistic~(CRPS) performance comparison of IBUG against two different nearest-neighbor models.~$k$NN estimates the conditional mean and variance using two different~$k$ values; and $k$NN-CB estimates the variance in the same way as~$k$NN, but uses the scalar output from the CatBoost model to estimate the conditional mean. Overall, these results suggest affinity is a better measure of similarity than Euclidean distance for uncertainty estimation in GBRTs.}
\label{app_tab:knn}
\centering
\vspace{0.5em}
\begin{tabular}{lccc}
\toprule
Dataset & $k$NN & $k$NN-CB & IBUG-CB \\
\midrule
Bike & {\bfseries 0.932}$_{(0.029)}$ & {\bfseries 0.978}$_{(0.049)}$ & {\bfseries 0.974}$_{(0.048)}$ \\
California & 0.579$_{(0.001)}$ & 0.219$_{(1e-03)}$ & {\bfseries 0.213}$_{(9e-04)}$ \\
Communities & 0.072$_{(0.002)}$ & {\bfseries 0.065}$_{(0.002)}$ & {\bfseries 0.065}$_{(0.002)}$ \\
Concrete & 4.645$_{(0.140)}$ & {\bfseries 1.872}$_{(0.085)}$ & {\bfseries 1.849}$_{(0.098)}$ \\
Energy & 0.875$_{(0.016)}$ & {\bfseries 0.153}$_{(0.010)}$ & {\bfseries 0.143}$_{(0.009)}$ \\
Facebook & 5.613$_{(0.065)}$ & 3.275$_{(0.068)}$ & {\bfseries 3.073}$_{(0.066)}$ \\
Kin8nm & 0.067$_{(7e-04)}$ & {\bfseries 0.051}$_{(5e-04)}$ & {\bfseries 0.051}$_{(6e-04)}$ \\
Life & 4.738$_{(0.078)}$ & {\bfseries 0.785}$_{(0.024)}$ & {\bfseries 0.794}$_{(0.023)}$ \\
MEPS & 7.283$_{(0.220)}$ & {\bfseries 6.181}$_{(0.107)}$ & {\bfseries 6.150}$_{(0.114)}$ \\
MSD & 5.312$_{(0.006)}$ & 4.446$_{(0.004)}$ & {\bfseries 4.410}$_{(0.005)}$ \\
Naval & 8e-04$_{(2e-05)}$ & 3e-04$_{(2e-06)}$ & {\bfseries 2e-04}$_{(2e-06)}$ \\
News & 2654$_{(52.0)}$ & {\bfseries 2597}$_{(52.3)}$ & {\bfseries 2545}$_{(41.0)}$ \\
Obesity & 5.526$_{(0.013)}$ & {\bfseries 1.900}$_{(0.043)}$ & {\bfseries 1.866}$_{(0.021)}$ \\
Power & 2.074$_{(0.022)}$ & {\bfseries 1.553}$_{(0.020)}$ & {\bfseries 1.542}$_{(0.020)}$ \\
Protein & 3.241$_{(0.010)}$ & {\bfseries 1.787}$_{(0.008)}$ & {\bfseries 1.784}$_{(0.008)}$ \\
STAR & 140$_{(1.553)}$ & {\bfseries 129}$_{(1.204)}$ & {\bfseries 130}$_{(1.214)}$ \\
Superconductor & 3.445$_{(0.041)}$ & {\bfseries 0.156}$_{(0.006)}$ & {\bfseries 0.153}$_{(0.006)}$ \\
Synthetic & 6.136$_{(0.047)}$ & {\bfseries 5.735}$_{(0.039)}$ & {\bfseries 5.731}$_{(0.040)}$ \\
Wave & 11987$_{(36.6)}$ & 2700$_{(17.0)}$ & {\bfseries 2679}$_{(16.0)}$ \\
Wine & 0.445$_{(0.004)}$ & {\bfseries 0.322}$_{(0.006)}$ & {\bfseries 0.322}$_{(0.006)}$ \\
Yacht & 3.354$_{(0.408)}$ & {\bfseries 0.275}$_{(0.048)}$ & {\bfseries 0.276}$_{(0.048)}$ \\
\midrule
% KNN W-T-L & - & 0-1-21 & 0-1-21 \\
% KNN-CB W-T-L & 21-1-0 & - & 0-12-10 \\
IBUG-CB W-T-L & 21-1-0 & 10-12-0 & - \\
\bottomrule
\end{tabular}
\end{table}

\newpage

\subsection{Comparison to Bayesian Additive Regression Trees}
\label{app_sec:bart}

BART~(Bayesian Additive Regression Trees) takes a Bayesian approach to uncertainty estimation in trees~\cite{chipman2010bart}. Although well-grounded theoretically, BART requires expensive sampling techniques such as MCMC~(Markov Chain Monte Carlo) to provide approximate solutions.

In this section, we compare IBUG to BART using a popular open-source implementation.\footnote{\url{https://github.com/JakeColtman/bartpy}} However, due to BART's computational complexity, we tune the number of trees for both IBUG and BART using values [10, 50, 100, 200], set the number of chains for BART to 5, and run our comparison using a subset of the datasets in our empirical evaluation consisting of 11 relatively small datasets.

Tables~\ref{app_tab:bart_crps}~and~\ref{app_tab:bart_rmse} show IBUG consistently outperforms BART in terms of both probabilistic and point performance.

\begin{table}[ht]
\caption{Probabilistic~(CRPS~$\downarrow$) performance comparison between IBUG and BART.}
\label{app_tab:bart_crps}
\centering
\vspace{0.5em}
\begin{tabular}{lcc}
\toprule
Dataset & BART & IBUG \\
\midrule
Bike &        4.521$_{(0.119)}$ & {\bfseries 0.974}$_{(0.048)}$ \\
California &  0.285$_{(0.001)}$ & {\bfseries 0.213}$_{(9e-04)}$ \\
Communities & 0.072$_{(0.002)}$ & {\bfseries 0.065}$_{(0.002)}$ \\
Concrete &    3.067$_{(0.073)}$ & {\bfseries 1.849}$_{(0.098)}$ \\
Energy &      0.402$_{(0.023)}$ & {\bfseries 0.143}$_{(0.009)}$ \\
Kin8nm &      0.107$_{(8e-04)}$ & {\bfseries 0.051}$_{(6e-04)}$ \\
Naval &       1e-03$_{(1e-05)}$ & {\bfseries 2e-04}$_{(2e-06)}$ \\
Power &       2.225$_{(0.018)}$ & {\bfseries 1.542}$_{(0.020)}$ \\
STAR &        134$_{(1.493)}$   &  {\bfseries 130}$_{(1.214)}$  \\
Wine &        0.394$_{(0.005)}$ & {\bfseries 0.322}$_{(0.006)}$ \\
Yacht &       0.849$_{(0.039)}$ & {\bfseries 0.276}$_{(0.048)}$ \\
\midrule
IBUG W-T-L & 11-0-0 & - \\
% BART W-T-L & 0-0-11 & - \\
\bottomrule
\end{tabular}
\end{table}

\begin{table}[ht]
\caption{Point~(RMSE~$\downarrow$) performance comparison between IBUG and BART.}
\label{app_tab:bart_rmse}
\centering
\vspace{0.5em}
\begin{tabular}{lcc}
\toprule
Dataset & BART & IBUG \\
\midrule
Bike &        8.396$_{(0.273)}$ & {\bfseries 2.826}$_{(0.200)}$ \\
California &  0.547$_{(0.003)}$ & {\bfseries 0.432}$_{(0.001)}$ \\
Communities & 0.137$_{(0.004)}$ & {\bfseries 0.133}$_{(0.004)}$ \\
Concrete &    5.507$_{(0.161)}$ & {\bfseries 3.629}$_{(0.183)}$ \\
Energy &      0.685$_{(0.039)}$ & {\bfseries 0.264}$_{(0.023)}$ \\
Kin8nm &      0.186$_{(0.001)}$ & {\bfseries 0.086}$_{(8e-04)}$ \\
Naval &       0.002$_{(2e-05)}$ & {\bfseries 5e-04}$_{(5e-06)}$ \\
Power &       4.057$_{(0.049)}$ & {\bfseries 2.941}$_{(0.059)}$ \\
STAR &        234$_{(2.479)}$   &  {\bfseries 228}$_{(1.985)}$  \\
Wine &        0.708$_{(0.008)}$ & {\bfseries 0.596}$_{(0.012)}$ \\
Yacht &       1.624$_{(0.121)}$ & {\bfseries 0.668}$_{(0.125)}$ \\
\midrule
IBUG W-T-L & 11-0-0 & - \\
% BART W-T-L & 0-0-11 & - \\
\bottomrule
\end{tabular}
\end{table}

\newpage

\subsection{Different Tree-Sampling Strategies}
\label{app_sec:tree_sampling}

\begin{figure}[h]
\centering
\begin{subfigure}{\textwidth}
  \centering
  \includegraphics[width=1.0\linewidth]{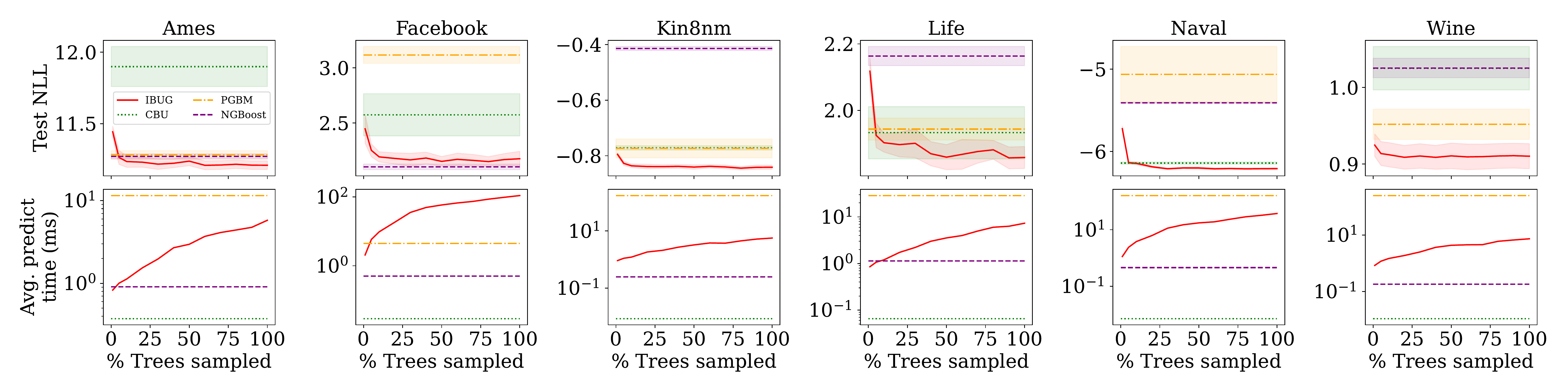}
  \caption{Sampling trees \emph{uniformly at random}.}
\end{subfigure}
\begin{subfigure}{\textwidth}
  \centering
  \includegraphics[width=1.0\linewidth]{figures/tree_sampling/ascending.pdf}
  \caption{Sampling trees \emph{first-to-last}.}
\end{subfigure}
\begin{subfigure}{\textwidth}
  \centering
  \includegraphics[width=1.0\linewidth]{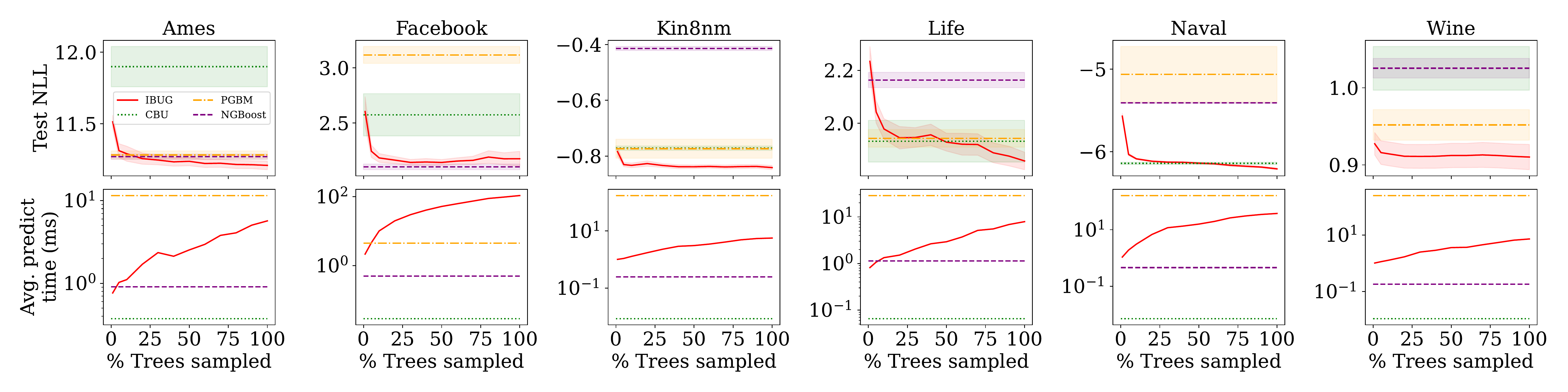}
  \caption{Sampling trees \emph{last-to-first}.}
\end{subfigure}
\caption{Probabilistic~(NLL) performance~(lower is better) and average prediction time~(in milliseconds) per test example~(lower is better) as a function of~$\tau$ for different sampling techniques.~\emph{Top}: sample trees uniformly at random,~\emph{middle}: sample trees first-to-last (in terms of boosting iteration),~\emph{bottom}: sample trees last-to-first. All methods result in similar prediction times; however, \emph{first-to-last} sampling typically provides the best NLL with the fewest number of trees sampled.}
\label{app_fig:tree_sampling}
\end{figure}

Figure~\ref{app_fig:tree_sampling} shows the probabilistic~(NLL) performance of IBUG as the number of trees sampled~($\tau$) increases using three different sampling strategies: \emph{uniformly at random}, \emph{first-to-last}, and \emph{last-to-first}.

We observe that sampling trees \emph{last-to-first} often requires sampling all trees in order to achieve the lowest NLL on the test set. When sampling \emph{uniformly at random}, NLL tends to plateau starting around 10\%. In contrast, sampling trees \emph{first-to-last} on the Kin8nm, Naval, and Wine datasets requires 5\% of the trees or less to result in the same or better NLL than when sampling all trees; these results provide some evidence that trees early in training contribute most, and suggest that sampling trees first-to-last may be most effective at obtaining the best probabilistic performance while sampling the fewest number of trees.

\newpage

\subsection{Leaf Density}
\label{app_sec:leaf_density}

\begin{figure}[h]
\centering
\begin{subfigure}{0.32\textwidth}
  \centering
  \includegraphics[width=1.0\linewidth]{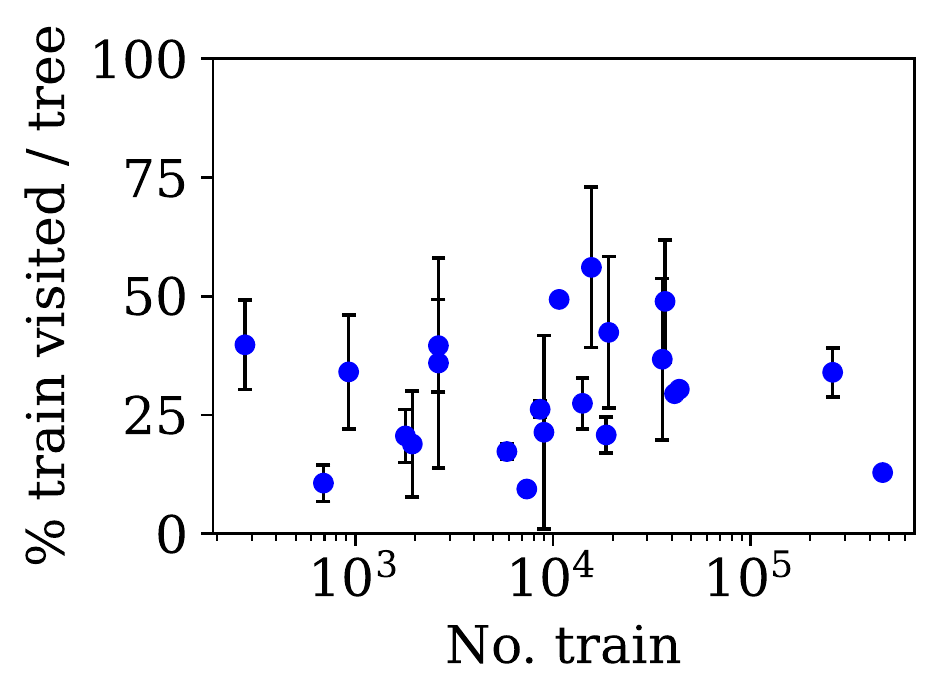}
  \caption{CatBoost}
\end{subfigure}
\begin{subfigure}{0.32\textwidth}
  \centering
  \includegraphics[width=1.0\linewidth]{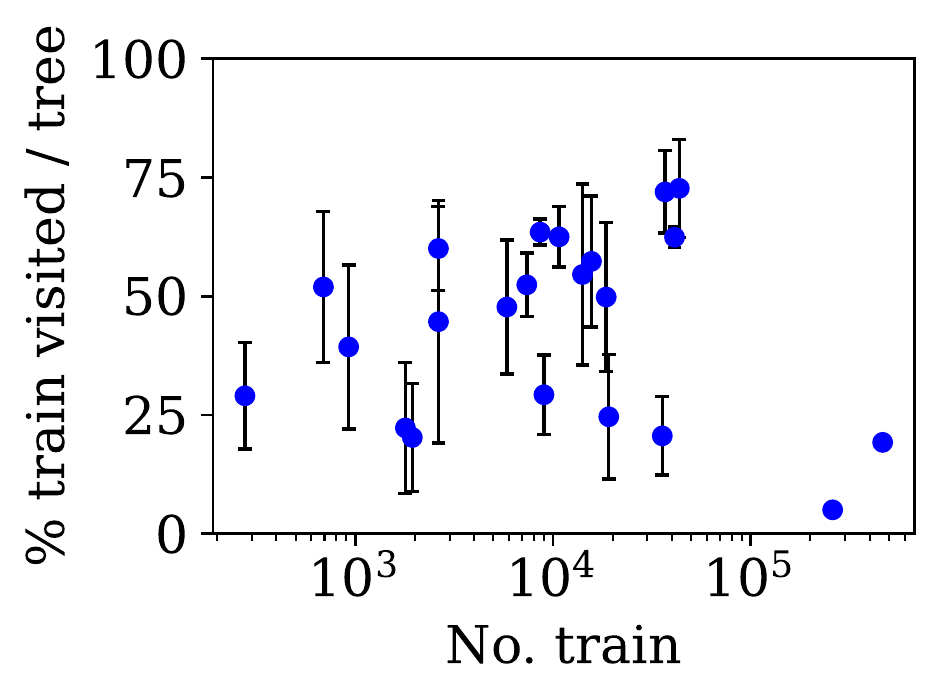}
  \caption{LightGBM}
\end{subfigure}
\begin{subfigure}{0.32\textwidth}
  \centering
  \includegraphics[width=1.0\linewidth]{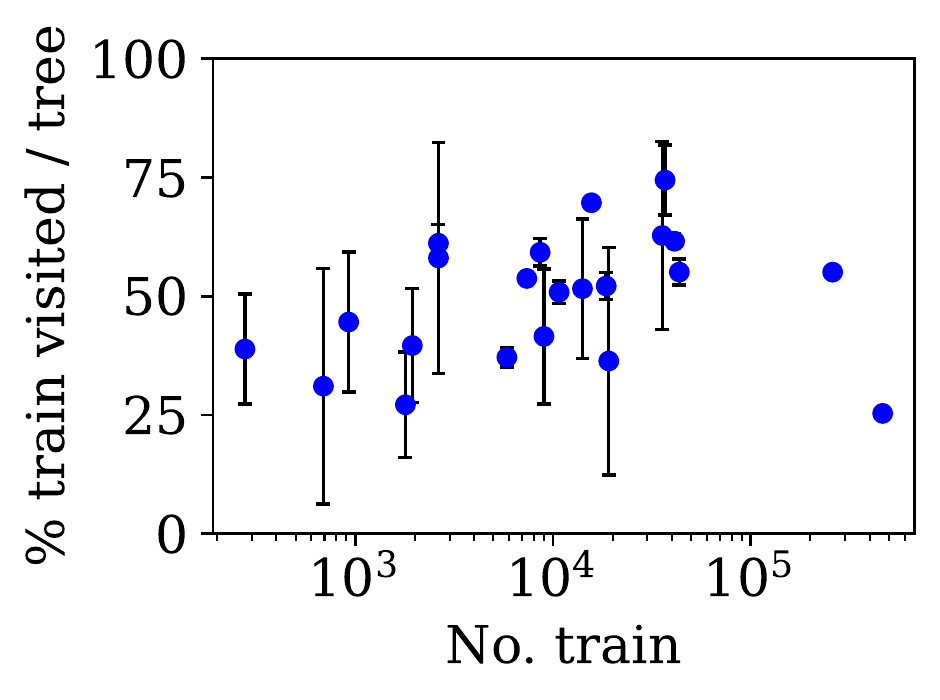}
  \caption{XGBoost}
\end{subfigure}
\caption{Average percentage of training instances visited per tree while computing affinity vectors on the test set test for each dataset. Results are averaged over all test instances, and error bars represent standard deviation; lower is better. In general, the number of training instances visited per tree is highly dependent on the dataset; and for some datasets, is also highly dependent on the test example~(points with large standard deviations).}
\label{fig:leaf_density}
\end{figure}

\begin{figure}[h]
\centering
\begin{subfigure}{\textwidth}
  \centering
  \includegraphics[width=1.0\linewidth]{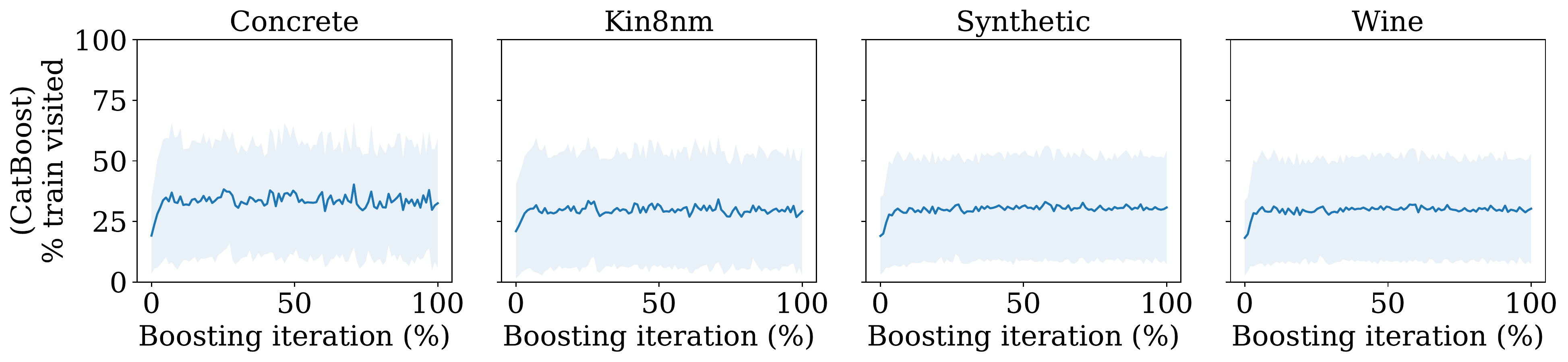}
\end{subfigure}
\begin{subfigure}{\textwidth}
  \centering
  \includegraphics[width=1.0\linewidth]{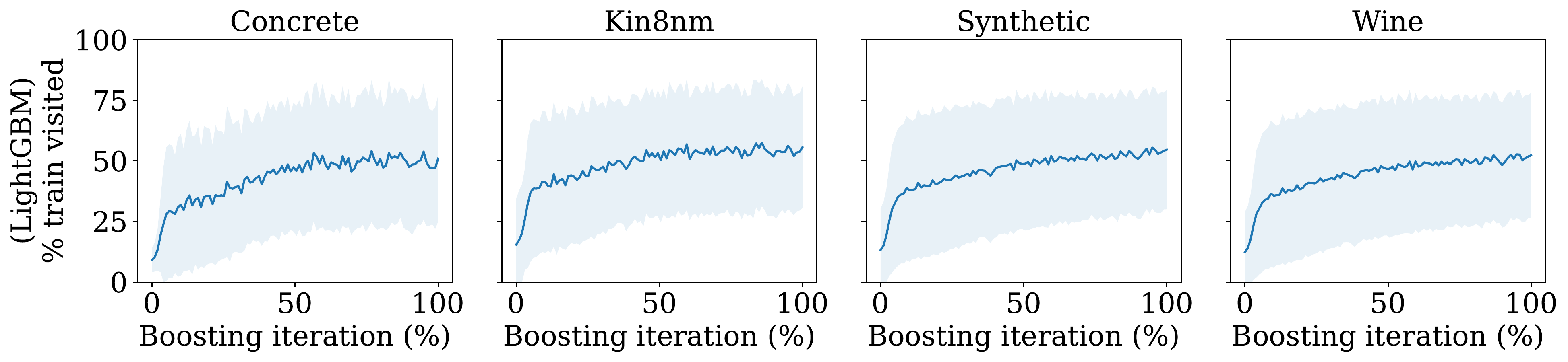}
\end{subfigure}
\begin{subfigure}{\textwidth}
  \centering
  \includegraphics[width=1.0\linewidth]{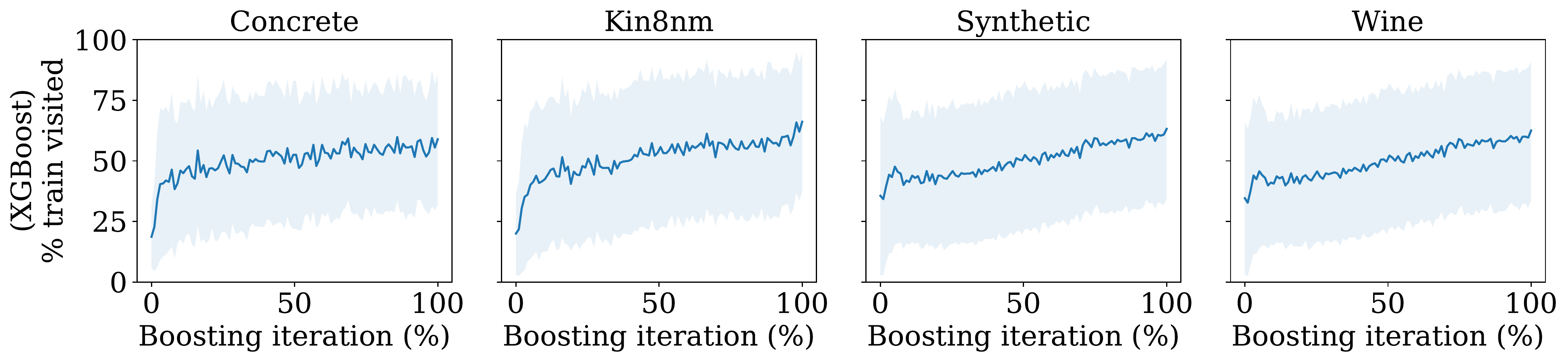}
\end{subfigure}
\caption{Average percentage of training instances visited at each iteration while computing affinity vectors on the test set for the Concrete, Kin8nm, Synthetic, and Wine datasets. Results are averaged over all test instances, and error bars represent standard deviation; lower is better. For LightGBM and XGBoost, weak learners later in training tend to pool a larger proportion of training instances into fewer leaves; in contrast, CatBoost has less dense leaves and training instances are more equally distributed among the leaves in each tree.}
\label{fig:leaf_density_tree}
\end{figure}

Figure~\ref{fig:leaf_density} shows the average percentage of train instances visited per tree as a function of the total number of training instances for each dataset. We note that for some datasets, CatBoost, LightGBM, and XGBoost induce regression trees with very dense leaves where over half the training instances belong to those leaves. Figure~\ref{fig:leaf_density_tree} shows average leaf density for each tree in the GBRT.

\end{document}